\pdfoutput=1

\documentclass[pdflatex,sn-nature]{sn-jnl}

\usepackage{graphicx}%
\usepackage{multirow}%
\usepackage{amsmath,amssymb,amsfonts}%
\usepackage{amsthm}%
\usepackage[title]{appendix}%
\usepackage{xcolor}%
\usepackage{textcomp}%
\usepackage{manyfoot}%
\usepackage{booktabs}%
\usepackage{array}%
\usepackage{tabularx}%
\usepackage{listings}%
\usepackage{placeins}%
\usepackage[protrusion=true,expansion=true]{microtype}%
\newcolumntype{L}[1]{>{\raggedright\arraybackslash}p{#1}}
\newcolumntype{C}[1]{>{\centering\arraybackslash}p{#1}}
\newcolumntype{Y}{>{\raggedright\arraybackslash}X}

\newcommand{\tabfont}{\fontsize{7.2bp}{9.5bp}\selectfont}
\newcommand{\tabnotefont}{\fontsize{6.8bp}{8.5bp}\selectfont}
\newcommand{\colh}[1]{{\boldmath\textbf{#1}}}
\newcommand{\tabnote}[1]{\par\vspace{3pt}{\tabnotefont\leftskip=0pt\rightskip=0pt\parfillskip=0pt plus 0.72\linewidth\emergencystretch=1.5em\noindent #1\par}}
\makeatletter
\def\abstractfont{\reset@font\fontsize{9bp}{11bp}\selectfont\leftskip=24pt\rightskip=24pt\parfillskip=0pt plus 0.68\textwidth\emergencystretch=1.5em}%
\long\def\@figurecaption#1#2{{\figurecaptionfont\parfillskip=0pt plus 0.72\linewidth\emergencystretch=1.5em{\bfseries#1}\hskip.7em#2\par}}%
\makeatother
\AddToHook{env/thebibliography/begin}{\setlength{\parfillskip}{0pt plus 1fil}}

\theoremstyle{thmstyleone}%
\newtheorem{theorem}{Theorem}%
\newtheorem{proposition}[theorem]{Proposition}%

\theoremstyle{thmstyletwo}%

\theoremstyle{thmstylethree}%

\let\table\tableorg
\let\endtable\endtableorg

\unnumbered

\begin{document}

\setlength{\parfillskip}{0pt plus 0.72\textwidth}
\setlength{\emergencystretch}{1.5em}
\renewcommand{\topfraction}{0.95}
\renewcommand{\bottomfraction}{0.92}
\renewcommand{\textfraction}{0.05}
\renewcommand{\floatpagefraction}{0.80}
\setcounter{topnumber}{3}
\setcounter{bottomnumber}{2}
\setcounter{totalnumber}{4}

\title[Causal Envelopes for Cardiopulmonary Risk Prediction]{Causally Constrained and Uncertainty-Aware Cardiopulmonary Risk Prediction}

\author[1]{\fnm{Yuxiang} \sur{Lai}}
\author[2]{\fnm{Jiashuo} \sur{Zhang}}
\author*[1]{\fnm{Liang} \sur{Zhao}}\email{liang.zhao@emory.edu}
\author*[1,3]{\fnm{Xiaofeng} \sur{Yang}}\email{xfyang@uchicago.edu}

\affil[1]{\orgdiv{Department of Computer Science}, \orgname{Emory University}, \orgaddress{\city{Atlanta}, \state{GA}, \country{USA}}}
\affil[2]{\orgdiv{Department of Computer Science}, \orgname{Johns Hopkins University}, \orgaddress{\city{Baltimore}, \state{MD}, \country{USA}}}
\affil[3]{\orgdiv{Department of Radiation and Cellular Oncology}, \orgname{The University of Chicago}, \orgaddress{\city{Chicago}, \state{IL}, \country{USA}}}

\abstract{Routine chest computed tomography (CT) captures pulmonary, cardiac, and vascular phenotypes that jointly stratify lung cancer and cardiovascular disease, the two leading causes of death in lung-screening populations. Yet clinical risk calculators evaluate each disease in isolation and omit CT findings, whereas unconstrained machine learning overfits small hospital cohorts, learns treatment- and illness-confounded risk directions, and conceals uncertainty when patient records are incomplete. Here we present the Causal Envelope Agent (CE-Agent), a two-stage framework that compiles peer-reviewed epidemiological evidence into explicit constraints on risk-factor weights before any outcome is seen, while estimating local baseline risk and newly added CT phenotypes from cohort data. At deployment, closed-form uncertainty propagation accounts for missing clinical or CT inputs within milliseconds per patient, running over two orders of magnitude faster than Monte Carlo sampling and returning calibrated risk intervals that direct chart review and selective imaging to borderline patients. Across 53,452 lung-screening participants from 33 US hospitals and 6,814 multi-ethnic adults, CE-Agent outperforms cardiovascular calculators in lung screening, exceeds the Pooled Cohort Equations on hard atherosclerotic events while correcting their 1.7-fold risk overestimation, and generalizes from a single hospital or CT scanner to external sites while lowering required sample sizes up to fivefold. Applied unchanged to lung cancer, CE-Agent matches established screening models with accurate calibration even when restricted to risk-factor directions alone, and pairing both predictions improves joint cardiopulmonary triage from a single screening visit.}

\maketitle
\setlength{\parfillskip}{0pt plus 0.72\textwidth}
\setlength{\emergencystretch}{1.5em}


\section{Introduction}\label{sec:intro}

Low-dose chest computed tomography (CT) is the standard of care for lung cancer screening in adults with a heavy smoking history~\cite{nlst2011reduced,dekoning2020nelson,uspstf2021lung}. The same adults carry a high burden of cardiovascular disease (CVD): smoking and aging raise the risk of both~\cite{lewington2002age,mons2015impact,tammemagi2013selection}, and in the National Lung Screening Trial (NLST) they were the two leading causes of death, with comparable shares of mortality~\cite{nlst2011reduced}. Every screening scan already images the heart and great vessels alongside the lungs, capturing pulmonary nodules, emphysema, coronary artery calcification (CAC), thoracic aortic calcification (TAC), pulmonary arterial remodeling, and cardiac chamber geometry~\cite{mcwilliams2013probability,chiles2015association,gondrie2010cardiovascular,jacobs2012coronary,wells2012pulmonary,hecht2017scct,miller2024predicting}. Society guidelines recommend reporting coronary calcification on every noncontrast chest CT~\cite{hecht2017scct}, yet screening eligibility is set by lung cancer risk alone~\cite{uspstf2021lung}. Established risk calculators each address one disease with fixed coefficients that omit chest CT phenotypes, and often miscalibrate when carried to new populations~\cite{vancalster2019calibration}. This holds for lung-screening models such as the Prostate, Lung, Colorectal, and Ovarian Cancer Screening Trial 2012 model (PLCOm2012)~\cite{tammemagi2013selection} and for cardiovascular equations, including Framingham general CVD~\cite{dagostino2008general}, the American College of Cardiology/American Heart Association (ACC/AHA) Pooled Cohort Equations (PCE)~\cite{goff20142013,grundy2019aha,arnett2019acc}, and AHA Predicting Risk of Cardiovascular Disease Events (PREVENT)~\cite{khan2024development}.

Data-driven risk models have advanced individualized prediction from imaging~\cite{miller2024predicting,niu2025medical} and molecular profiles~\cite{li2026proteomic} and have informed treatment prioritization~\cite{demircan2026data}. Extending them to joint cardiopulmonary assessment at individual screening sites raises three challenges. First, individual hospitals, demographic groups, and CT scanner platforms rarely accumulate enough outcome events to estimate multi-parameter risk models reliably~\cite{riley2020calculating,vancalster2019calibration}. Second, observational records mix baseline exposures with post-scan events, derived indices such as body mass index (BMI) or pulse pressure that duplicate their components, and measurements distorted by treatment or illness. Comorbidity lowers total cholesterol while raising mortality~\cite{jacobs1992report}, statin initiation in high-risk patients~\cite{grundy2019aha,uspstf2022statin} inverts lipid associations in treated cohorts~\cite{sperrin2018using}, and chronic obstructive pulmonary disease (COPD) enlarges the pulmonary artery~\cite{wells2012pulmonary}, confounding its cardiovascular signal. Unconstrained models can therefore learn risk directions opposite to clinical biology~\cite{pearl2009causality,peters2017elements}. Third, outpatient records often lack recent laboratory panels or a prior chest CT at the point of care, and substituting a single imputed value into a nonlinear risk equation hides the uncertainty of the individual risk estimate~\cite{kompa2021second,riley2025uncertainty} near preventive treatment thresholds~\cite{grundy2019aha,arnett2019acc,uspstf2022statin,pauker1980threshold}.

\begin{figure}[!t]
\centering
\includegraphics[width=\textwidth]{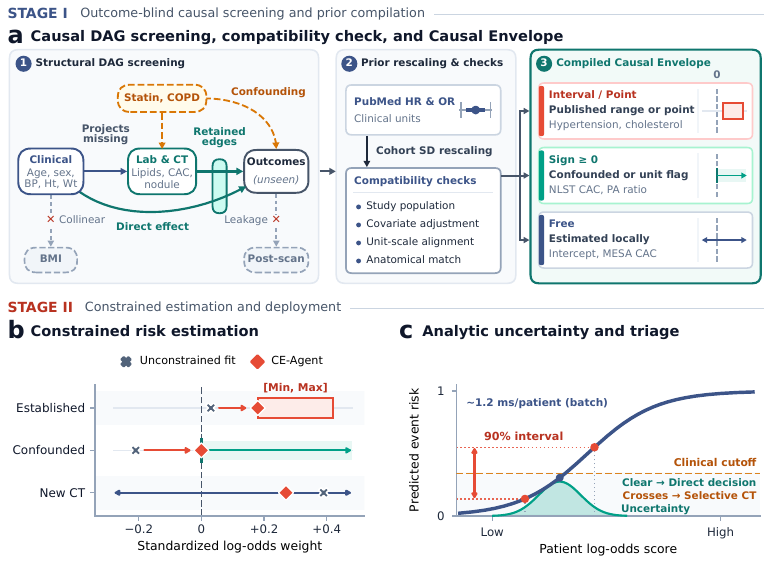}
\caption{\textbf{Overview of the Causal Envelope Agent (CE-Agent).}
\textbf{a}, Stage I, before any outcome is seen: a directed acyclic graph (DAG) screen removes composite indices and post-scan information; published hazard and odds ratios (HR, OR) per standard deviation (SD) assign each predictor a range, point value, direction, or free estimation.
\textbf{b}, Stage II fits weights within these constraints, zeroes those contradicting the published direction, and estimates new CT phenotypes locally.
\textbf{c}, Closed-form uncertainty propagation returns a risk, its 90\% interval, and a triage decision. BP, blood pressure; Ht, height; Wt, weight; BMI, body mass index; CAC, coronary artery calcification; COPD, chronic obstructive pulmonary disease; PA, pulmonary artery.}\label{fig:fig8}
\end{figure}

Here we introduce the Causal Envelope Agent (CE-Agent), a two-stage framework that grounds clinical and chest CT risk modeling in peer-reviewed epidemiological evidence (Fig.~\ref{fig:fig8}). Its central object, the Causal Envelope, states in advance how each predictor may relate to an outcome: post-scan events and composite indices that duplicate their components are excluded, established risk factors keep their published direction and range, and local baseline risk and newly added CT phenotypes are estimated from cohort data. Stage I compiles this envelope from published hazard and odds ratios before any outcome label is seen, and Stage II fits each predictor's weight inside it. At deployment, an analytic formula evaluated once per patient propagates missing-data uncertainty, directing chart review and selective imaging to patients whose unmeasured inputs could change a clinical decision~\cite{geifman2017selective,leibig2017leveraging,raghu2019direct,dvijotham2023enhancing}. The name reflects its role as the deterministic, fixed-path core of a broader cardiopulmonary agentic framework: CE-Agent executes five modules in a fixed, auditable sequence (clinical-record reading, CT-biomarker extraction, prior compilation, constrained estimation, and uncertainty propagation).

We evaluate it in two complementary US cohorts totaling $60{,}266$ adults. NLST includes $53{,}452$ adults with a heavy smoking history, randomized to low-dose CT (LDCT) or chest radiography at $33$ screening hospitals~\cite{nlst2011reduced}. The Multi-Ethnic Study of Atherosclerosis (MESA) includes $6{,}814$ adults from four racial/ethnic groups imaged on four CT scanner platforms~\cite{bild2002multi,detrano2008coronary}. We first establish accuracy, single-hospital learning, protection against confounding, and uncertainty handling on cardiovascular outcomes in both cohorts. We then apply the same framework, unchanged, to lung cancer in NLST and combine the two predictions into joint cardiopulmonary triage from a single screening encounter, which adds cardiovascular prevention to lung cancer screening.


\section{Results}\label{sec:results}

\subsection{Prognostic accuracy and calibration across lung screening and primary prevention}\label{sec:res_nlst}

\begin{figure}[!htbp]
\centering
\includegraphics[width=\textwidth]{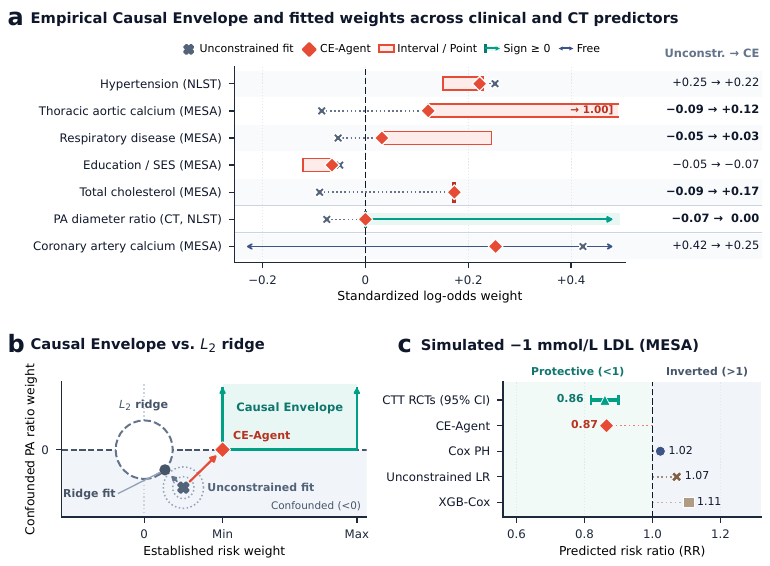}
\caption{\textbf{The Causal Envelope: constraints, geometry, and simulated interventions.}
\textbf{a}, Standardized log-odds weights by constraint type, from unconstrained fits (Unconstr.) and CE-Agent (CE).
\textbf{b}, Schematic: ridge ($L_2$) shrinkage pulls an established risk factor and the confounded pulmonary artery (PA) diameter ratio toward zero; the Causal Envelope holds the first in range and zeroes the second.
\textbf{c}, Risk ratios (RR) for MESA cardiovascular mortality under a simulated $1\text{ mmol/L}$ reduction in low-density lipoprotein (LDL) cholesterol, against Cholesterol Treatment Trialists' (CTT) randomized controlled trials (RCTs; 95\% confidence interval, CI)~\cite{ctt2010efficacy}. SES, socioeconomic status; Cox PH, Cox proportional hazards; LR, logistic regression; XGB-Cox, Cox-boosted trees.}\label{fig:fig_ce}
\end{figure}

\begin{figure}[!htbp]
\centering
\includegraphics[width=\textwidth]{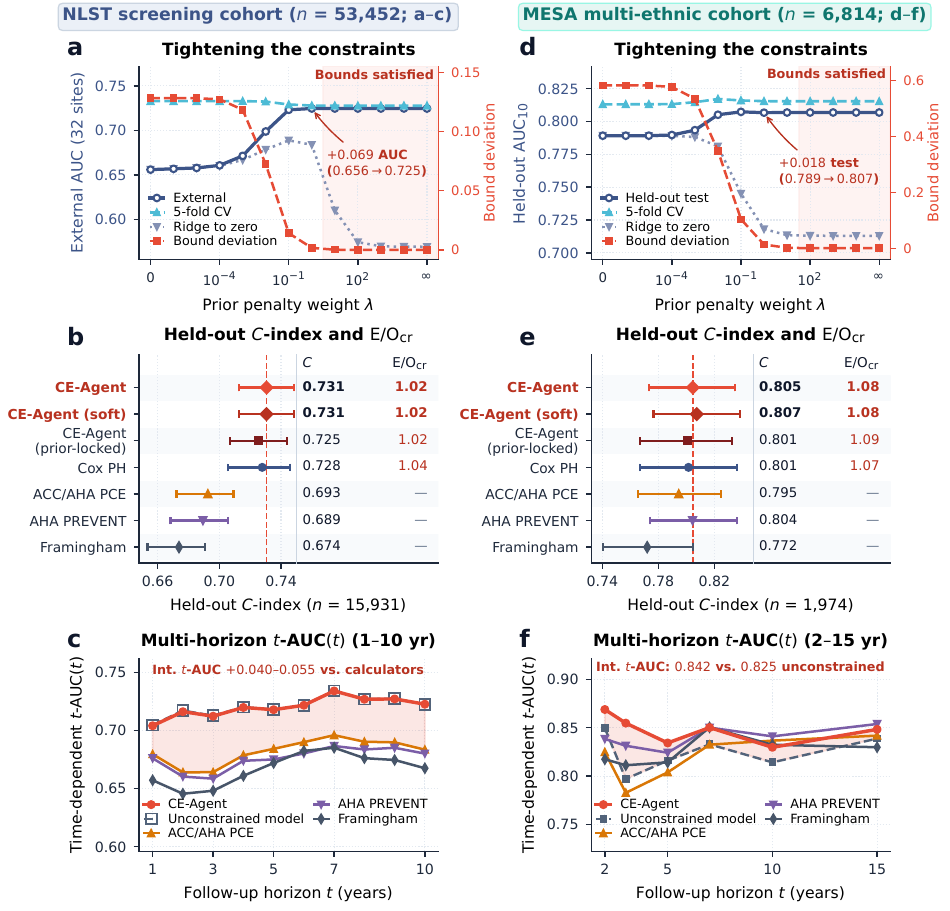}
\caption{\textbf{Agreement with published evidence and held-out performance.}
\textbf{a},\textbf{d}, Discrimination against penalty $\lambda$ on departures from published bounds ($\lambda = 0$, unconstrained; $\lambda = \infty$, CE-Agent) or, for ridge, from zero, in single-hospital NLST (\textbf{a}) and MESA (\textbf{d}); CV, cross-validation; right axis, mean bound deviation.
\textbf{b},\textbf{e}, Test-set concordance index ($C\text{-index}$; 95\% confidence interval, CI) and 10-year $\text{E/O}_{\text{cr}}$ for CE-Agent, variants with weights penalized toward published ranges (soft) or fixed at published values (prior-locked), and comparators; dashes, calculators target composite endpoints with nonfatal events (calibration in Tables~\ref{tab:mesa_main} and~\ref{tab:app_composite_endpoints}).
\textbf{c},\textbf{f}, Time-dependent $\text{AUC}$ ($t\text{-AUC}$) for cardiovascular mortality in NLST under competing mortality~\cite{blanche2013estimating} (\textbf{c}; shading, gain over PCE; inset $\Delta\text{Int.\ }t\text{-AUC}$ spans $1\text{--}13\text{ yr}$) and in MESA~\cite{uno2007evaluating} (\textbf{f}; gain over unconstrained). $\text{E/O}_{\text{cr}}$, competing-risk expected-to-observed ratio; $\text{AUC}_{10}$, 10-year $\text{AUC}$; PCE, ACC/AHA Pooled Cohort Equations; PREVENT, AHA risk equations; Int., integrated.}\label{fig:fig1}
\end{figure}

Published evidence improved cardiovascular risk models in both lung screening (NLST) and multi-ethnic primary prevention (MESA). Before any outcome was seen, Stage I retained $13$ predictors in NLST and $22$ in MESA and assigned each a published range or value, a direction only, or free estimation (Figs.~\ref{fig:fig8}a and~\ref{fig:fig_ce}a). In single-hospital NLST models, fitting within the Causal Envelope (Fig.~\ref{fig:fig8}b) raised the mean area under the receiver operating characteristic curve ($\text{AUC}$) for cardiovascular mortality at the other $32$ hospitals from $0.656$ without constraints to $0.725$ (Fig.~\ref{fig:fig1}a). In the MESA test set, it raised 10-year $\text{AUC}$ from $0.789$ to $0.807$, ahead of ridge shrinkage toward zero~\cite{hoerl1970ridge} and of the same model with weights fixed at published values (Fig.~\ref{fig:fig1}d,e and Table~\ref{tab:app_mesa_frontier}).

In the held-out NLST test cohort ($n = 15{,}931$), CE-Agent trained across all $33$ hospitals achieved a concordance index ($C\text{-index}$) of $0.731$ and accurate 10-year calibration (expected-to-observed event ratio $\text{E/O}_{\text{cr}} = 1.02$, accounting for competing non-cardiovascular deaths~\cite{blanche2013estimating}). It outperformed the Framingham, PCE, and PREVENT calculators while matching unconstrained survival and machine learning (ML) models pooled across all $33$ hospitals (Fig.~\ref{fig:fig1}b,c and Tables~\ref{tab:nlst_main} and~\ref{tab:app_frontier_benchmarks}). Because NLST did not measure blood pressure, lipids, or kidney function, those inputs were set to standard untreated US adult reference values for the three calculators (Methods).

In the held-out MESA cohort ($n = 1{,}974$), where blood pressure, fasting lipids, and kidney function were directly measured, the $22\text{-predictor}$ CE-Agent outperformed every survival and ML baseline built on the same predictors and matched both the clinical calculators and an expanded $58\text{-predictor}$ ($63\text{-column}$) Cox proportional hazards model, while staying calibrated for cardiovascular death ($\text{E/O}_{\text{cr}} = 1.08$; Fig.~\ref{fig:fig1}e,f and Tables~\ref{tab:mesa_main}, \ref{tab:app_mesa_frontier}, and~\ref{tab:app_mesa_22v53}). Evaluated directly on each calculator's composite target, CE-Agent exceeded PCE in integrated time-dependent $\text{AUC}$ ($t\text{-AUC}$)~\cite{uno2007evaluating} on hard atherosclerotic cardiovascular disease (ASCVD; $0.802$ versus $0.777$, $P = 0.010$; Methods) while correcting PCE's $1.7\text{-fold}$ risk overestimation ($\text{E/O}_{\text{cr}} = 1.04$ versus $1.70$; Tables~\ref{tab:mesa_main} and~\ref{tab:app_composite_endpoints}). On the broad cardiovascular composite matched to Framingham, CE-Agent likewise exceeded Framingham in integrated $t\text{-AUC}$ ($0.813$ versus $0.789$, $P = 0.012$) and maintained accurate calibration (expected-to-observed ratio $0.97$ versus $1.44$; Table~\ref{tab:app_composite_endpoints}).

\subsection{Generalization across hospitals, demographics, and scanners}\label{sec:res_q75}\label{sec:res_clinical_utility}

\begin{figure}[!t]
\centering
\includegraphics[width=\textwidth]{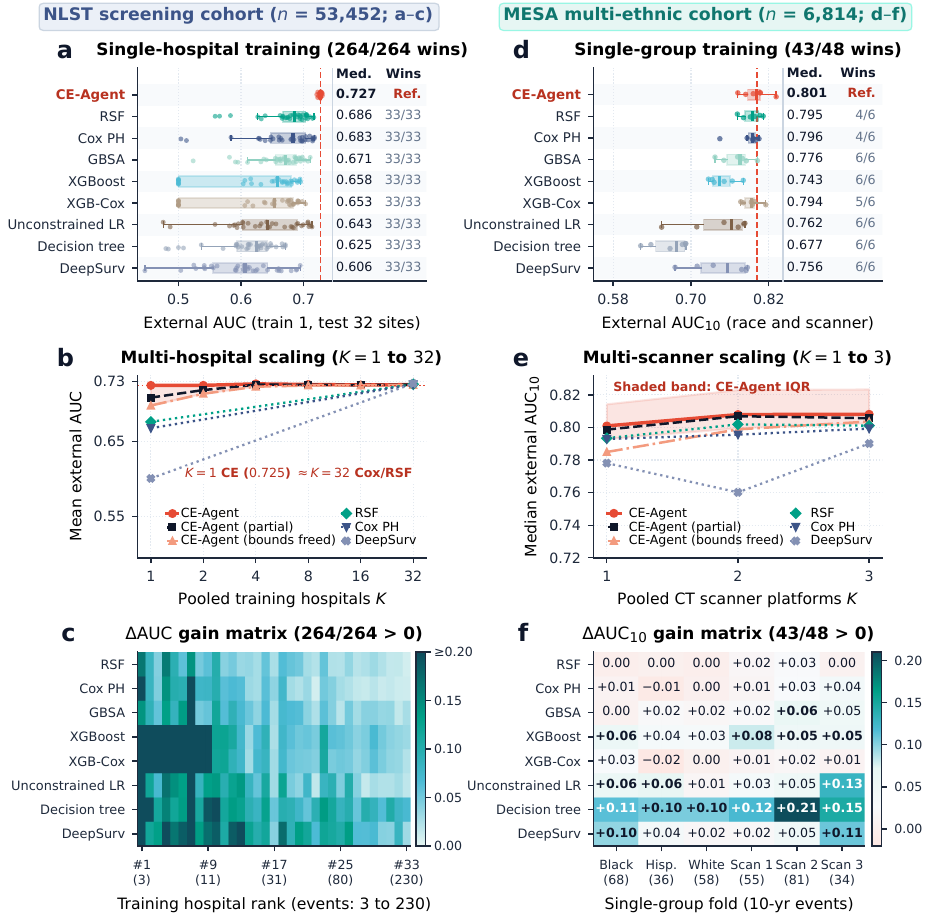}
\caption{\textbf{Generalization across environments and scaling with pooled centers.}
\textbf{a},\textbf{d}, External $\text{AUC}$ of models trained at one NLST hospital (\textbf{a}) or one of six MESA racial/ethnic or scanner groups with $\ge 20$ training events (\textbf{d}); wins, training environments favoring CE-Agent.
\textbf{b},\textbf{e}, External discrimination after pooling NLST hospitals (\textbf{b}) and MESA scanner platforms (\textbf{e}; band, interquartile range, IQR). Partial, some published ranges retained; bounds freed, range-constrained predictors unconstrained.
\textbf{c},\textbf{f}, External $\text{AUC}$ gain of CE-Agent over each baseline by NLST hospital (\textbf{c}) and MESA group (\textbf{f}). RSF, random survival forest; GBSA, gradient-boosted survival analysis; XGBoost, gradient-boosted trees; DeepSurv, neural-network survival model; Med., median; Ref., reference; Hisp., Hispanic.}\label{fig:fig2}
\end{figure}

Models developed at one hospital or on one CT scanner generalized to the others (Tables~\ref{tab:nlst_ood}--\ref{tab:mesa_ood}). Trained at any single NLST hospital, CE-Agent outperformed eight survival and ML baselines at the other hospitals in all $264$ hospital--baseline pairs (Fig.~\ref{fig:fig2}a,c). Trained on one of six MESA racial/ethnic or CT scanner groups with at least $20$ training events, it outperformed the baselines in $43$ of $48$ comparisons, including every transfer between scanners (Fig.~\ref{fig:fig2}d,f). A single-hospital or single-scanner CE-Agent matched the external discrimination of unconstrained models pooled across $32$ hospitals or three scanners (Fig.~\ref{fig:fig2}b,e) and outperformed methods designed for transfer to new sites~\cite{rothenhausler2021anchor,duchi2021learning,steyerberg2004validation} (Table~\ref{tab:app_lambda_rivals}).

The constraints lowered the minimum sample size required for reliable model development~\cite{riley2020calculating} by $5.4\text{-fold}$ in NLST and $2.5\text{-}$ to $3.7\text{-fold}$ in MESA (Fig.~\ref{fig:fig5}b,e). On their own data, $12$ of $33$ NLST hospitals met this minimum, against none without constraints. External calibration slopes ($1.0$ indicating ideal calibration) stayed within the target range ($0.80$ to $1.25$) for $28$ of $33$ single-hospital NLST models, against $8$ without constraints (Table~\ref{tab:app_ood_licensing}).

CE-Agent delivered consistent clinical net benefit across hospitals and cohorts. In decision curve analysis~\cite{vickers2006decision}, which weighs true against false positives across clinical risk thresholds, each of the $33$ single-hospital NLST models gave higher mean net benefit at the other $32$ hospitals than treat-all and treat-none strategies across thresholds of $5\%$ to $10\%$ (Fig.~\ref{fig:fig5}a). For MESA hard ASCVD, CE-Agent led the eight baselines in net benefit integrated across decision thresholds (Fig.~\ref{fig:fig5}d and Table~\ref{tab:mesa_ood}). NLST-trained weights applied directly in MESA, adjusted only for baseline prevalence, achieved the highest concordance of any model carried across cohorts (Table~\ref{tab:app_zeroshot_transport} and Fig.~\ref{fig:fig7}a).

\subsection{Integrating chest CT phenotypes without confounded weights}\label{sec:res_mechanics}\label{sec:res_mediators_imaging}

\begin{figure}[!htbp]
\centering
\includegraphics[width=\textwidth]{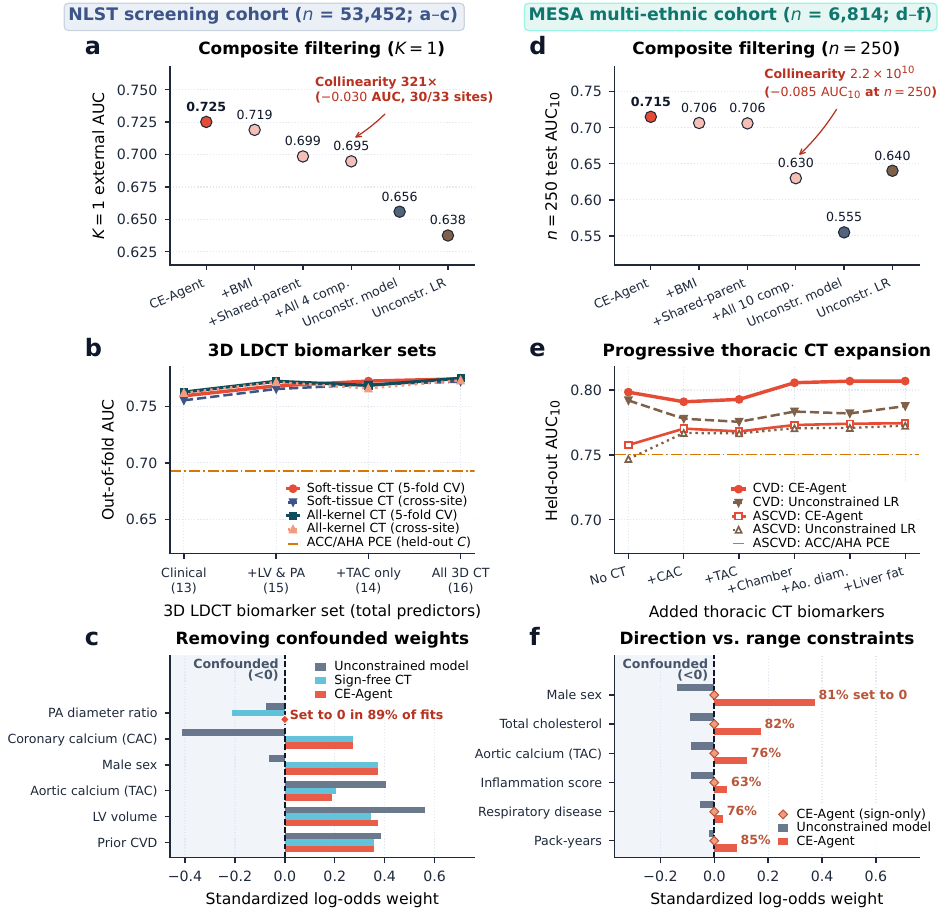}
\caption{\textbf{Composite-index screening, added CT phenotypes, and zeroed confounded weights.}
\textbf{a},\textbf{d}, Discrimination of CE-Agent, unconstrained baselines, and models retaining composite indices (+BMI; +Shared-parent, sharing a parent measurement; +All 4 or +All 10 comp., all composites) in single-hospital ($K = 1$ hospital) NLST fits (\textbf{a}) and $250\text{-participant}$ MESA subsamples (\textbf{d}); collinearity, condition number (\textbf{a}, ratio to CE-Agent).
\textbf{b},\textbf{e}, Discrimination with added CT phenotypes: NLST low-dose CT (LDCT), soft-tissue ($n = 5{,}269$) or any ($n = 6{,}467$) reconstruction kernel (\textbf{b}); five MESA measures, for cardiovascular disease (CVD) death and hard atherosclerotic CVD (ASCVD; \textbf{e}).
\textbf{c},\textbf{f}, Standardized log-odds weights from unconstrained fits, fits with CT directions free (sign-free CT), a direction-only envelope (sign-only), and CE-Agent; percentages (\textbf{f}), sign-only fits with the weight zeroed. LV, left ventricular; TAC, thoracic aortic calcification; 3D, three-dimensional; Ao.\ diam., ascending aortic diameter.}\label{fig:fig4}
\end{figure}

CE-Agent incorporates new three-dimensional (3D) chest CT phenotypes while keeping confounded associations out of the model. Adding automatically measured left ventricular volume, pulmonary artery (PA) diameter ratio, and TAC raised cross-validated $\text{AUC}$ from $0.759$ to $0.774$ among NLST participants with soft-tissue reconstructions ($n = 5{,}269$; Fig.~\ref{fig:fig4}b), and five electrocardiogram-gated CT phenotypes gave parallel gains in MESA (Fig.~\ref{fig:fig4}e). For the PA diameter ratio, which enlarges in COPD, the envelope specifies only the expected direction; Stage II zeroed this confounded weight in $89\%$ of NLST cross-validation fits (Figs.~\ref{fig:fig_ce}a,b and~\ref{fig:fig4}c). In MESA, where statin therapy and chronic illness invert crude lipid and blood-pressure slopes, published interval bounds kept six established risk factors---including total cholesterol---positive, whereas a direction-only envelope collapsed them to zero in $63\%$ to $85\%$ of fits (Fig.~\ref{fig:fig4}f and Tables~\ref{tab:app_prior_channel_mechanics}--\ref{tab:app_mesa_mechanics}).

Composite-index screening and published bounds each improved small-sample accuracy (Figs.~\ref{fig:fig4}a,d and~\ref{fig:fig3}a,b,e,f). Excluding redundant composite indices while retaining their underlying primary clinical measurements raised external $\text{AUC}$ by $0.030$ in single-hospital NLST fits and by $0.085$ in $250\text{-participant}$ MESA subsamples.

Under simulated reductions in smoking, body weight, cholesterol, and systolic blood pressure~\cite{mons2015impact,psc2009body,ctt2010efficacy,ettehad2016blood}, CE-Agent's predicted risk fell or held steady for every participant in both cohorts, whereas unconstrained ML baselines predicted paradoxical risk increases in up to half of patients in NLST and $46\%\text{--}100\%$ on lipid reduction in MESA (Figs.~\ref{fig:fig_ce}c and~\ref{fig:fig9}c,g). In MESA, lowering low-density lipoprotein (LDL) cholesterol by $1\text{ mmol/L}$ gave a risk ratio of $0.865$, matching the pooled randomized-trial estimate for vascular mortality ($0.86$~\cite{ctt2010efficacy}; Fig.~\ref{fig:fig_ce}c), and lowering systolic blood pressure by $10\text{ mmHg}$ gave $0.876$, consistent with trial estimates for all-cause mortality ($0.87$) and major cardiovascular events ($0.80$~\cite{ettehad2016blood}; Table~\ref{tab:app_interventions_triage}).

\subsection{Real-time uncertainty propagation, outpatient adaptation, and targeted CT triage}\label{sec:res_aim13}\label{sec:res_mesa}

\begin{figure}[!htbp]
\centering
\includegraphics[width=\textwidth]{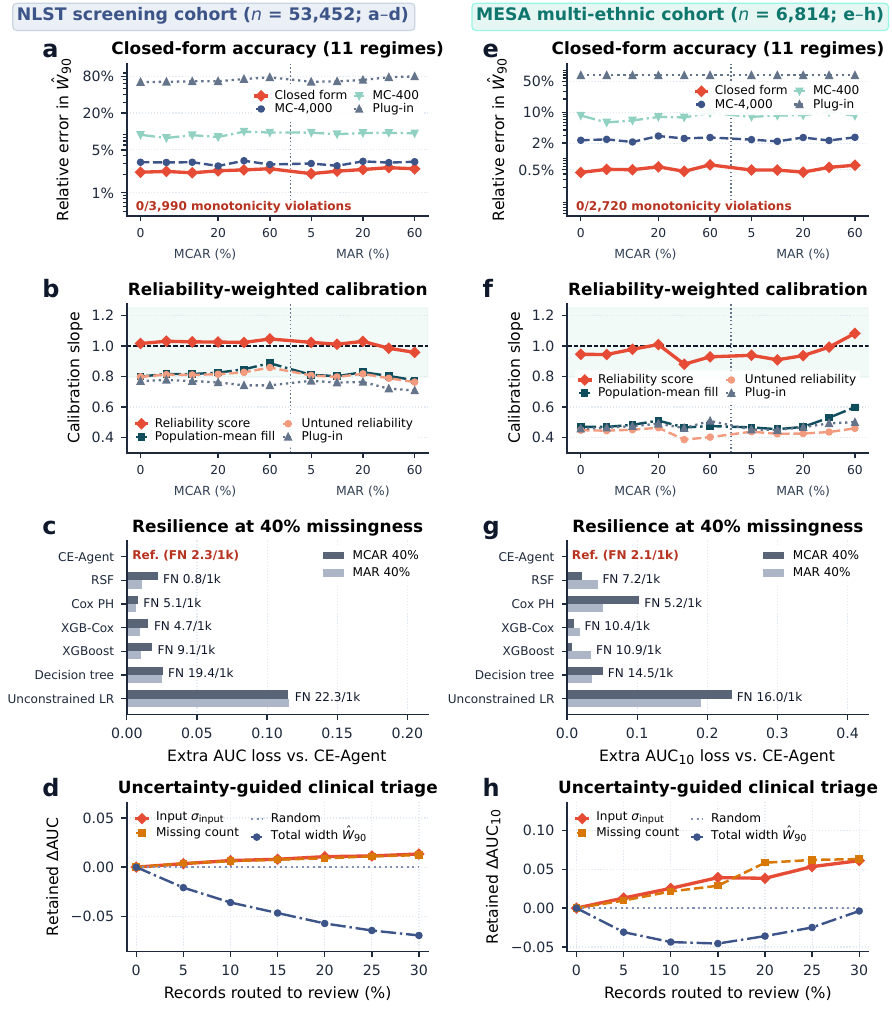}
\caption{\textbf{Uncertainty propagation, reliability scoring, and uncertainty-guided triage in NLST (a--d) and MESA (e--h).}
\textbf{a},\textbf{e}, Relative error of 90\% interval widths ($\hat{W}_{90}$) from closed-form propagation, $400\text{-}$ and $4{,}000\text{-draw}$ Monte Carlo (MC), and single-value (plug-in) imputation, against $100{,}000\text{-draw}$ MC; $0\%\text{--}60\%$ missing completely at random (MCAR) or at random given BMI (MAR); monotonicity violations, risk rising when a harmful predictor is zeroed.
\textbf{b},\textbf{f}, Calibration slopes for reliability scoring, its untuned version, population-mean fill, and plug-in imputation (band, $0.80\text{--}1.25$).
\textbf{c},\textbf{g}, Extra discrimination loss at $40\%$ missingness versus CE-Agent; labels, overconfident false negatives (FN; events predicted below half the training event rate) per $1{,}000$ (/1k): most uncertain fifth, $5\%\text{--}40\%$ missingness (\textbf{c}); all participants, MCAR (\textbf{g}).
\textbf{d},\textbf{h}, Discrimination gain among retained records with up to $30\%$ reviewed, ranked by missing-data uncertainty ($\sigma_{\text{input}}$), missing-predictor count, total interval width, or at random.}\label{fig:fig6}
\end{figure}

\begin{figure}[!htbp]
\centering
\includegraphics[width=\textwidth]{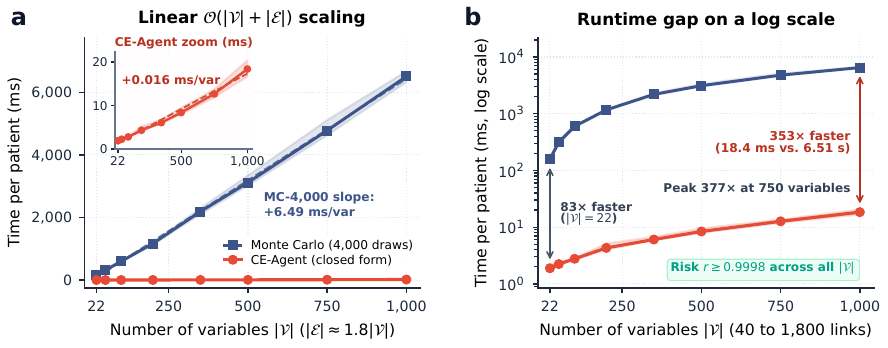}
\caption{\textbf{Per-patient runtime from 22 to 1,000 variables.}
\textbf{a}, Single-core runtime (median, IQR) for $4{,}000\text{-draw}$ MC and CE-Agent closed-form propagation on synthetic risk-factor graphs; inset, CE-Agent curve.
\textbf{b}, Logarithmic axis; closed-form propagation was $83\text{-}$ to $377\text{-fold}$ faster ($353\text{-fold}$ at $1{,}000$ variables), with Pearson $r \ge 0.9998$ between risk estimates. $|\mathcal{E}|$, links.}\label{fig:fig_scalability}
\end{figure}

Records are often incomplete at the point of care. With up to $60\%$ of predictors missing, CE-Agent's closed-form uncertainty propagation (Fig.~\ref{fig:fig8}c) reproduced Monte Carlo predictive intervals with mean relative error up to $2.3\%$, in $1$ to $3\text{ ms}$ per patient on a single core (Fig.~\ref{fig:fig6}a,e and Tables~\ref{tab:app_uncertainty_propagation}--\ref{tab:app_mesa_uncertainty}). On synthetic risk-factor graphs, runtime grew linearly with model size, reaching $18.4\text{ ms}$ at $1{,}000$ variables. Across sizes it was $83\text{-}$ to $377\text{-fold}$ faster than $4{,}000\text{-draw}$ Monte Carlo sampling ($353\text{-fold}$ at $1{,}000$ variables), with near-identical risk estimates (Fig.~\ref{fig:fig_scalability}a,b).

To prevent missing inputs from distorting predictions, CE-Agent weights each predictor by a reliability score that down-weights missing or high-variance imputations (Methods). This score kept calibration slopes near $1.0$, preserved discrimination at $40\%$ missingness, and reduced overconfident false negatives (events assigned low risk) at least twofold versus Cox, boosted-tree, logistic, and decision-tree baselines (Fig.~\ref{fig:fig6}b,c,f,g). When all five chest CT phenotypes were unmeasured at simulated MESA outpatient visits, projecting them from routine clinical predictors gave an expected-to-observed event ratio of $0.96$, against $0.81$ under population-mean filling (Fig.~\ref{fig:fig7}d and Tables~\ref{tab:app_mesa_mediators}--\ref{tab:app_mesa_missing_ct}).

The same uncertainty estimate directs chart review and selective imaging. Reviewing the $30\%$ of records with the largest missing-data uncertainty raised $\text{AUC}$ among the remaining records by $0.013$ in NLST and $0.061$ in MESA (Fig.~\ref{fig:fig6}d,h). In two-stage selective CT triage using published weights (Methods), CE-Agent requested a CAC read only when its $90\%$ pre-CT risk interval---computed with CT predictors treated as unmeasured---crossed the $5\%$ 10-year fatal-CVD threshold~\cite{conroy2003estimation,debacker2003european}. Among CVD-free LDCT-arm participants in the NLST training set ($n = 15{,}727$), this rule required $55\%$ fewer radiologist reads than CAC referral of everyone at intermediate risk by recalibrated PCE while raising net benefit by $1.14$ per $1{,}000$ ($95\%$ CI, $0.23$ to $2.13$; Fig.~\ref{fig:fig9}d). For guideline statin decisions~\cite{grundy2019aha} among MESA adults not on lipid-lowering therapy ($n = 1{,}465$), CE-Agent matched the net benefit of guideline CAC referral at the same scan budget ($+16.0$ versus $+16.1$ per $1{,}000$) and reached $+15.4$ without imaging, exceeding PCE alone ($+14.0$; Fig.~\ref{fig:fig9}h and Table~\ref{tab:app_interventions_triage}).

\subsection{Lung cancer prediction and joint cardiopulmonary triage}\label{sec:res_lung_dual}

\begin{figure}[!htbp]
\centering
\includegraphics[width=\textwidth]{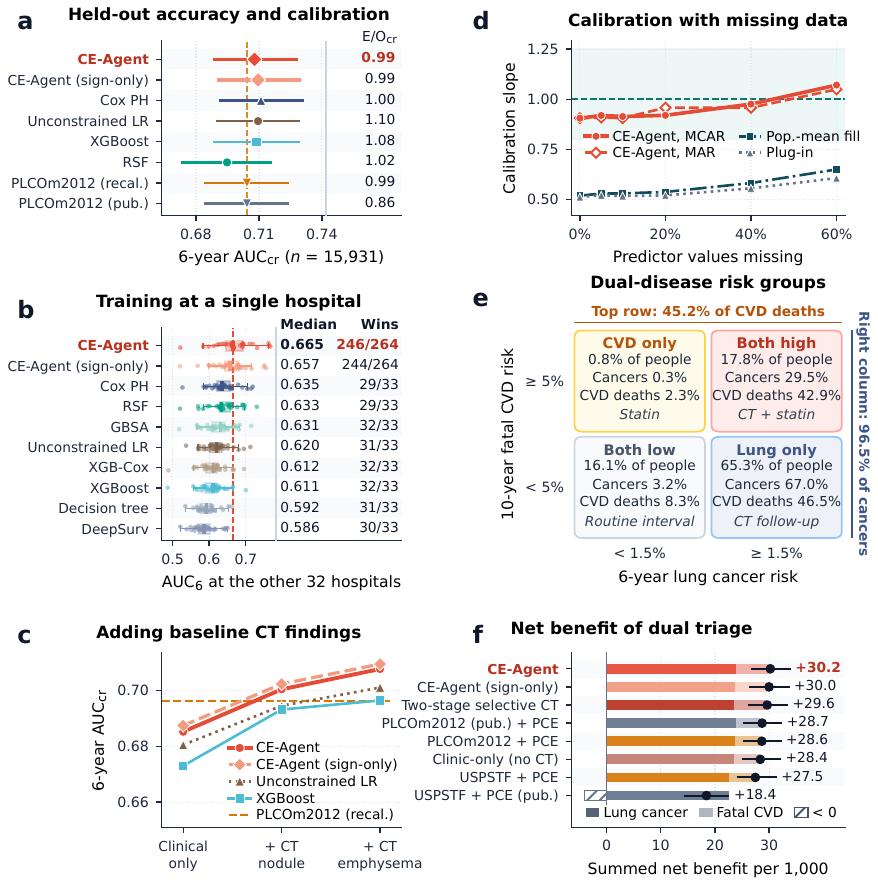}
\caption{\textbf{Lung cancer prediction and joint cardiopulmonary triage in NLST.}
\textbf{a}, Test-set 6-year competing-risk $\text{AUC}$ ($\text{AUC}_{\text{cr}}$; 95\% CI) and $\text{E/O}_{\text{cr}}$ for CE-Agent, baselines on the same $11$ predictors, and PLCOm2012~\cite{tammemagi2013selection} recalibrated to NLST (recal.; dashed line) or published (pub.).
\textbf{b}, Six-year $\text{AUC}$ ($\text{AUC}_6$) of single-hospital models evaluated across the remaining $32$ hospitals.
\textbf{c}, $\text{AUC}_{\text{cr}}$ in test-set LDCT participants ($n = 7{,}786$) with radiologist-reported nodules and emphysema added.
\textbf{d}, Training-cohort calibration slopes with $0\%\text{--}60\%$ of predictors missing (band, $0.80\text{--}1.25$; published weights).
\textbf{e}, Risk groups at $1.5\%$ lung cancer and $5\%$ fatal CVD cutoffs~\cite{tammemagi2014evaluation,conroy2003estimation,debacker2003european}.
\textbf{f}, Summed net benefit per $1{,}000$ (95\% CI). USPSTF, US Preventive Services Task Force~\cite{uspstf2021lung}; two-stage, CT read only when the 90\% pre-CT risk interval crosses a cutoff; clinic-only, lung cancer CE-Agent without CT, paired with PCE; PLCOm2012 and PCE recalibrated unless marked pub.}\label{fig:fig_lung_dual}
\end{figure}

Applied unchanged to NLST lung cancer ($n = 15{,}931$ held-out participants; Fig.~\ref{fig:fig_lung_dual} and Tables~\ref{tab:app_lung_main}--\ref{tab:app_dual_triage}), CE-Agent achieved accurate $6\text{-year}$ calibration ($\text{E/O}_{\text{cr}} = 0.99$, versus $0.86$ for published PLCOm2012) and matched recalibrated PLCOm2012 in competing-risk $\text{AUC}$ ($\text{AUC}_{\text{cr}} = 0.708$ versus $0.704$; difference $+0.004$, $95\%$ CI $-0.008$ to $+0.017$, meeting the prespecified $-0.02$ criterion; Fig.~\ref{fig:fig_lung_dual}a). It matched or outperformed survival and ML baselines built on the same $11$ predictors, achieved comparable accuracy when restricted to risk-factor directions alone ($\text{AUC}_{\text{cr}} = 0.710$), and maintained accurate calibration ($\text{E/O}_{\text{cr}} = 0.98$) and discrimination when refitted to $10\text{-year}$ lung cancer mortality (Table~\ref{tab:app_lung_main}).

The same properties held across single-hospital training, baseline CT expansion, and incomplete records. Trained at a single hospital, CE-Agent won $246$ of $264$ external comparisons with same-hospital ML baselines (Fig.~\ref{fig:fig_lung_dual}b). Among held-out LDCT participants ($n = 7{,}786$), adding radiologist-reported nodules and visual emphysema~\cite{mcwilliams2013probability} raised $\text{AUC}_{\text{cr}}$ from $0.685$ to $0.708$ ($P = 0.036$; Fig.~\ref{fig:fig_lung_dual}c). With up to $60\%$ of predictors missing in the training cohort ($n = 37{,}521$, using published weights), reliability scoring kept calibration slopes between $0.91$ and $1.07$ (Fig.~\ref{fig:fig_lung_dual}d).

Pairing the lung cancer and cardiovascular predictions from the same screening encounter identified a substantial dual-risk subgroup (Fig.~\ref{fig:fig_lung_dual}e,f and Table~\ref{tab:app_dual_triage}). At cutoffs of $1.5\%$ $6\text{-year}$ lung cancer risk and $5.0\%$ $10\text{-year}$ fatal-CVD risk~\cite{tammemagi2014evaluation,conroy2003estimation,debacker2003european}, participants above both thresholds comprised $17.8\%$ of the held-out cohort yet accounted for $42.9\%$ of cardiovascular deaths and $29.5\%$ of lung cancers. Whereas US Preventive Services Task Force (USPSTF) criteria~\cite{uspstf2021lung} qualified nearly all participants ($99.4\%$), CE-Agent focused lung follow-up on $83.1\%$ while capturing $96.5\%$ of $6\text{-year}$ lung cancers.

With CE-Agent, the combined net benefit of lung cancer and cardiovascular decisions reached $+30.2$ per $1{,}000$, outperforming USPSTF criteria paired with recalibrated PCE ($+27.5$ per $1{,}000$, $P < 0.001$; Fig.~\ref{fig:fig_lung_dual}f). Selective two-stage reading---scoring CT features only when a participant's pre-CT risk interval crossed either cutoff---required CT reads for $61.1\%$ of participants while preserving full-scan net benefit (difference $-0.58$ per $1{,}000$, $95\%$ CI $-1.26$ to $0.00$, within the prespecified $-2.0$ criterion). Within the LDCT arm ($n = 7{,}786$), where the screening scan itself supplies all CT inputs, CE-Agent again outperformed USPSTF criteria plus recalibrated PCE ($+33.4$ versus $+30.8$ per $1{,}000$, $P = 0.024$; Table~\ref{tab:app_dual_triage}).


\section{Discussion}\label{sec:discussion}

Risk stratification from routine clinical records and chest CT needs models that respect clinical timing, learn reliably from a single hospital's cohort, and quantify uncertainty when records are incomplete. CE-Agent compiles peer-reviewed epidemiological evidence into a Causal Envelope before any outcome fitting and pairs constrained estimation with closed-form uncertainty propagation. The result is calibrated, uncertainty-aware prediction of cardiovascular risk in a national lung-screening trial and a multi-ethnic primary-prevention cohort, and of lung cancer risk in the screening trial.

Published epidemiological bounds guide a model differently from generic statistical shrinkage (Fig.~\ref{fig:fig_ce}b). Generic shrinkage penalties~\cite{hoerl1970ridge,tibshirani1996regression} pull every coefficient toward zero regardless of medical evidence, leaving small-sample models exposed to variance and to confounding-driven sign reversals~\cite{pearl2009causality}. The Causal Envelope instead removes post-scan events and collinear composite indices before fitting, holds established risk factors within their published ranges, and lets baseline incidence and the effect sizes of newly added CT phenotypes adapt to the local cohort. In small cohorts, the constraints leave fewer free parameters and raise expected discrimination, so established sample-size criteria~\cite{riley2020calculating} are met with a $2.5\text{-}$ to $5.4\text{-fold}$ smaller development sample, enough for $12$ of $33$ NLST hospitals to develop a model on their own data. In large multi-center cohorts, the constrained model matches the discrimination of unconstrained pooled models while keeping treatment-confounded cholesterol positive and zeroing comorbidity-confounded weights such as COPD-related PA enlargement~\cite{wells2012pulmonary}.

At the point of care, combining constrained risk estimation with closed-form uncertainty propagation lets one model serve both imaging centers and routine outpatient clinics. When no prior scan is available, the model projects the unmeasured CT phenotypes from clinical predictors, which kept predicted risk close to observed risk in simulated outpatient visits. Measuring missing-data uncertainty directs chart review and selective CT~\cite{kompa2021second,riley2025uncertainty,geifman2017selective,leibig2017leveraging} toward borderline patients whose unmeasured features could change statin decisions at guideline thresholds~\cite{grundy2019aha,arnett2019acc,uspstf2022statin,pauker1980threshold}.

Assessing lung cancer and cardiovascular risk together on the baseline screening CT addresses a gap in preventive thoracic care~\cite{nlst2011reduced,chiles2015association,hecht2017scct}. One screening scan informs both. Here about one in six screened participants were at high risk for both diseases and accounted for more than two in five cardiovascular deaths. Where binary eligibility criteria qualified nearly every participant, individual lung cancer risk focused follow-up on five in six while capturing $96.5\%$ of $6\text{-year}$ lung cancers (Table~\ref{tab:app_dual_triage}). In the LDCT arm, both assessments drew on nodule, emphysema, and coronary calcium descriptors from the screening scan itself, with no additional radiation or scan time. Reading CT only for borderline cases---about three in five participants---preserved the net benefit of reading every scan.

CE-Agent's modular design opens several directions for clinical translation. First, risk is computed from a weighted sum of clinical and CT phenotypes, which clinicians can audit and which supports closed-form uncertainty propagation. This layer can sit directly on automated 3D thoracic and abdominal CT segmentation~\cite{wasserthal2023totalsegmentator,pickhardt2020opportunistic,miller2024predicting} and on imaging foundation models~\cite{niu2025medical} for screening, inpatient, and emergency scans. Second, building on transfer across MESA groups, automated prior compilation can incorporate international multi-ancestry registries, proteomic panels~\cite{li2026proteomic}, and genomic panels. Third, extending validation beyond the two US cohorts studied here, prospective studies that add clinic vital signs, laboratory panels, longitudinal health records, and serial chest CT to screening programs can measure real-world effects on nodule management, statin prescribing, and cardiopulmonary outcomes.

In summary, CE-Agent shows that peer-reviewed epidemiological knowledge can be compiled into explicit constraints and paired with real-time uncertainty propagation at deployment. Grounding estimation in accumulated medical evidence yields cardiovascular models that stay calibrated and generalize from a single hospital or CT scanner to the others, alongside calibrated lung cancer prediction matching established screening models. It keeps risk directions consistent with established biology, accommodates incomplete outpatient records, and turns a single chest CT into an opportunistic dual-disease screening exam.


\section{Methods}\label{sec:methods}

\subsection{Ethics oversight}\label{sec:meth_ethics}

This study analyzed de-identified secondary data from two National Institutes of Health (NIH) clinical cohorts conducted in accordance with the Declaration of Helsinki. NLST (ClinicalTrials.gov identifier NCT00047385) was approved by the institutional review board at each of the $33$ participating US screening centers, and all participants provided written informed consent~\cite{nlst2011reduced}. MESA (NCT00005487) was approved by the institutional review boards of all six participating field centers and the National Heart, Lung, and Blood Institute (NHLBI) coordinating center, with written informed consent obtained from all participants~\cite{bild2002multi}. Access to de-identified NLST and MESA datasets was granted under approved National Cancer Institute (NCI) Cancer Data Access System and NHLBI Biologic Specimen and Data Repository Information Coordinating Center (BioLINCC) data-use agreements.

\subsection{Problem formulation and two-stage architecture}\label{sec:meth_problem}\label{sec:meth_arch}

We consider a training cohort $\mathcal{D} = \{(x_i, y_i)\}_{i=1}^{n}$ of $n$ participants with baseline clinical and chest CT records $x_i$ across $q$ candidate variables and adjudicated binary cardiopulmonary outcomes $y_i \in \{0, 1\}$. NLST models are fitted to event status over follow-up and aligned to the $6\text{-}$ or $10\text{-year}$ competing-risk horizon, whereas MESA models are fitted directly to $10\text{-year}$ status, with deaths from other causes coded as non-events and participants censored earlier excluded (Appendix~\ref{app:protocols}). Within our cardiopulmonary agentic framework, CE-Agent forms the deterministic, fixed-path core executing specialized clinical-record-reading, CT-biomarker-extraction, prior-compilation, constrained-estimation, and uncertainty-propagation modules in two stages (Fig.~\ref{fig:fig8}). The clinical-record reader maps each cohort's baseline fields to model variables, enters recorded values as observed, and flags missing fields, which carry their population prior into the uncertainty propagation described below. In Stage I (Fig.~\ref{fig:fig8}a), structural screening selects $d$ admissible baseline predictors $\mathcal{V}_{\text{adm}} \subseteq \{1, \dots, q\}$ (re-indexed as $\{1, \dots, d\}$). Each observed entry $X_{ij}$ is standardized to $Z_{ij} = (X_{ij} - \hat{\mu}_j)/\hat{\sigma}_j$ using center $\hat{\mu}_j$ and positive scale $\hat{\sigma}_j > 0$ (empirical training moments in MESA and pre-specified cohort scales in NLST; Appendix~\ref{app:protocols}), forming $Z \in \mathbb{R}^{n \times d}$. Verified published effect ranges are then compiled into a convex parameter set, the Causal Envelope $\Omega_{\text{CE}} \subseteq \mathbb{R}^{1+d}$, before viewing outcome labels. In Stage II (Fig.~\ref{fig:fig8}b,c), CE-Agent estimates the unpenalized baseline intercept $\hat{c}_0 \in \mathbb{R}$ and standardized log-odds weights $\hat{c} \in \mathbb{R}^d$ within $\Omega_{\text{CE}}$. At deployment, analytic variance propagation outputs calibrated risk $\hat{p}_i \in (0, 1)$, 90\% predictive interval width $\hat{W}_{90,i}$, and missing-data log-odds uncertainty $\sigma_{\text{input},i} \ge 0$.

\subsection{Causal structural screening and missing-biomarker projection}\label{sec:meth_dag}

Candidate variables are organized on a domain directed acyclic graph (DAG) encoding temporal ordering and biological hierarchy (Appendix~\ref{app:math}). Structural screening excludes post-baseline events to prevent outcome leakage and separates derived linear composite scores from directly measured intermediate biomarkers. Any derived linear composite $C_\ell = \sum_{p \in \operatorname{pa}(C_\ell)} w_{p\ell} Z_p$ formed with fixed weights $w_{p\ell} \in \mathbb{R}$ from retained standardized parent columns $Z_p \in \mathbb{R}^n$ ($\operatorname{pa}(C_\ell) \subseteq \mathcal{V}_{\text{adm}}$) lies in the column space of $Z$. Excluding $C_\ell$ therefore leaves full-rank risk predictions unchanged while preventing small-sample collinearity and Fisher information singularity (Proposition~\ref{prop:mediator}, Appendix~\ref{app:math}). Ratio indices such as BMI, which are nearly collinear with their retained parents, are screened by the same rule in the NLST cardiovascular model. By contrast, directly measured intermediate biomarkers $M_m$ with standardized column $Z_m \in \mathbb{R}^n$---including blood pressure, lipids, glucose, high-sensitivity C-reactive protein (CRP), kidney function, chronic respiratory status, and thoracic CT phenotypes---capture subclinical biological variation beyond upstream clinical factors. Chronic systemic inflammation, represented by a pre-specified linear summary of smoking exposure and BMI (the upstream inflammation score), enters the cardiovascular models as a single column; BMI, which is not a direct predictor in the NLST cardiovascular model, contributes through this score. When a continuous intermediate biomarker $M_m$ is unmeasured at deployment, CE-Agent projects its conditional mean $\mu_{im}$ and variance $v_{im}$ recursively from its parents, using sign-constrained training weights $(\hat{b}_m, \hat{w}_{pm})$, parent moments $(\mu_{ip}, v_{ip})$ equal to $(Z_{ip}, 0)$ when observed, and the training residual variance $\hat{\sigma}_m^2$:
\begin{equation}\label{eq:projection}
\begin{aligned}
\mu_{im} &= \hat{b}_m + \sum_{p \in \operatorname{pa}(M_m)} \hat{w}_{pm} \mu_{ip}, \\
v_{im} &= \hat{\sigma}_m^2 + \sum_{p \in \operatorname{pa}(M_m)} \hat{w}_{pm}^2 v_{ip}, \qquad \hat{\sigma}_m^2 = \widehat{\operatorname{Var}}(Z_m - \mu_m).
\end{aligned}
\end{equation}
Here $\mu_m = (\mu_{1m}, \dots, \mu_{nm})^\top$, and the variance recursion treats parent projection errors as independent. Binary intermediate biomarkers $M_m \in \{0, 1\}$ analogously use sign-constrained logistic projection and delta-method variance (Appendix~\ref{app:math}). When all chest CT biomarkers are unmeasured at simulated outpatient visits, CE-Agent substitutes the unattenuated expectations $\mu_{im}$, which preserve the cohort risk scale; variants attenuate these projections by their variance or by the reliability score defined below (Fig.~\ref{fig:fig7}d and Table~\ref{tab:app_mesa_missing_ct}).

\subsection{Literature prior compilation and compatibility screening}\label{sec:meth_compiler}

Published epidemiological studies report hazard, odds, or risk ratios across clinical increments $\Delta x_j^{\text{pub}} > 0$, such as smoking intensity, hypertension status, alcohol intake, educational attainment, particulate air pollution, chronic respiratory disease, or thoracic and coronary calcification increments~\cite{tammemagi2013selection,chiles2015association,bos2015comparison,aune2021hypertension,ronksley2011association,khaing2017effects,gonzalez2025emphysema,heo2022effect,mattila2023mortality,li2025additive,navarese2026causal,tasdighi2025association}. Supporting publications for each predictor--outcome link were cataloged before model fitting, with PubMed abstracts and source tables archived with identifier, access date, and checksum. In the pooled NLST cardiovascular specification, regular-expression verification against archived verbatim PubMed abstracts extracts the published log-effect interval $[\beta_{j,\text{lo}}^{\text{raw}}, \beta_{j,\text{hi}}^{\text{raw}}]$ and point estimate $\beta_j^{\text{raw}}$ directly~\cite{aune2021hypertension,ronksley2011association,gonzalez2025emphysema}; lung-cancer and point-anchored specifications additionally compile verified intervals and point priors from archived full-text tables and meta-analyses~\cite{tammemagi2013selection,chiles2015association,bos2015comparison,khaing2017effects,heo2022effect,mattila2023mortality,li2025additive,navarese2026causal,tasdighi2025association}. When several archived estimates qualified for one predictor, the estimate with the longest follow-up, and then the narrowest interval, was selected, and alternatives were logged with the selection rationale. Hazard, odds, and risk ratios all serve as log-odds effects because the outcomes are rare (cumulative incidence below $10\%$)~\cite{zhang1998whats}.

CE-Agent rescales these effects to standardized local cohort units via $\kappa_j = \hat{\sigma}_j / \Delta x_j^{\text{pub}}$ and records the established biological direction $\delta_j = \operatorname{sign}(\kappa_j \beta_j^{\text{raw}}) \in \{-1, +1\}$. A four-criterion compatibility check (Appendix~\ref{app:math}) evaluates study population compatibility, covariate over-adjustment, unit-scale alignment, and anatomical match. Predictors with verified directional evidence form $\mathcal{C}_{\text{sign}} \subseteq \mathcal{V}_{\text{adm}}$, while predictors without it, such as sex and race/ethnicity in the lung cancer model and cardiac chamber area and aortic diameter in the MESA cardiovascular mortality model, remain unconstrained in $\mathcal{C}_{\text{free}} = \mathcal{V}_{\text{adm}} \setminus \mathcal{C}_{\text{sign}}$. A new CT phenotype with an expected direction but no compatible published effect size, such as the PA diameter ratio, carries the sign constraint alone, so its effect size is estimated from local data. Within $\mathcal{C}_{\text{sign}}$, predictors passing all four checks form the interval-bounded subset $\mathcal{C}_{\text{box}}$ with rescaled bounds $[\ell_j, u_j]$. Established clinical risk factors fixed at published point estimates form $\mathcal{C}_{\text{point}} \subseteq \mathcal{C}_{\text{sign}} \setminus \mathcal{C}_{\text{box}}$, and predictors flagged on any compatibility check retain directional sign constraints in $\mathcal{C}_{\text{sign}} \setminus (\mathcal{C}_{\text{box}} \cup \mathcal{C}_{\text{point}})$. A fixed weight is either $c_j^{\text{pub}} = \kappa_j \beta_j^{\text{raw}}$ or, in single-hospital NLST fits and in MESA, a weight obtained by numerically differentiating published risk equations~\cite{goff20142013,dagostino2008general,khan2024development}. For intercept $c_0 \in \mathbb{R}$ and weight vector $c = (c_1, \dots, c_d)^\top \in \mathbb{R}^d$, the compiled Causal Envelope is (Figs.~\ref{fig:fig8}a and~\ref{fig:fig_ce}a):
\begin{equation}\label{eq:jacobian}
\begin{aligned}
\bigl[\ell_j, u_j\bigr] &= \kappa_j \bigl[\beta_{j,\text{lo}}^{\text{raw}},\; \beta_{j,\text{hi}}^{\text{raw}}\bigr], \\
\Omega_{\text{CE}} &= \bigl\{(c_0, c) \in \mathbb{R}^{1+d} \;\big|\; c_j \in [\ell_j, u_j]\;\;\forall j \in \mathcal{C}_{\text{box}}, \\
&\qquad\qquad c_j = c_j^{\text{pub}}\;\;\forall j \in \mathcal{C}_{\text{point}},\;\; \delta_j c_j \ge 0\;\;\forall j \in \mathcal{C}_{\text{sign}}\bigr\}.
\end{aligned}
\end{equation}
In primary pooled multi-center NLST estimation of cardiovascular ($d = 13$, three interval bounds) and lung cancer ($d = 11$, five interval bounds) risk, event counts far exceed $d$, so predictors with compatible published intervals enter $\mathcal{C}_{\text{box}}$ and $\mathcal{C}_{\text{point}}$ is empty. A sign-only lung cancer variant keeps directional constraints on eight predictors without the PLCOm2012 numerical intervals (Appendix~\ref{app:protocols}). In single-hospital NLST fits and in MESA ($d = 22$), core clinical risk factors are fixed in $\mathcal{C}_{\text{point}}$, which reserves local degrees of freedom for pulmonary, laboratory, and CT biomarkers. The point-anchored NLST specification used for single-hospital results fixes $9$ of its $13$ weights (eight clinical and CAC factors and the upstream inflammation score) and fits $4$ behavioral and pulmonary predictors within $\mathcal{C}_{\text{box}}$. The prior-locked specification fixes every weight at its published value and sets only the intercept from training prevalence. Because the primary CAC weight was derived in NLST~\cite{chiles2015association}, single-hospital fits were repeated with a CAC weight from an independent screening trial~\cite{mascalchi2021moderate}; external discrimination changed by less than $0.001$ $\text{AUC}$ (Fig.~\ref{fig:fig3}c).

\subsection{Constrained risk estimation and sample-size efficiency}\label{sec:meth_convex}\label{sec:meth_execution}\label{sec:meth_stats}

We define the directional sign cone $\Omega_{\text{sign}} = \{(c_0, c) \in \mathbb{R}^{1+d} \mid \delta_j c_j \ge 0\;\forall j \in \mathcal{C}_{\text{sign}}\}$, linear log-odds $\eta_i = c_0 + \sum_{j=1}^d c_j Z_{ij}$, and empirical logistic negative log-likelihood $\mathcal{L}_{\text{emp}}(c_0, c; \mathcal{D}) = -\frac{1}{n} \sum_{i=1}^n [y_i \eta_i - \ln(1 + e^{\eta_i})]$. CE-Agent computes $(\hat{c}_0, \hat{c})$ by solving the convex optimization problem:
\begin{equation}\label{eq:objective}
\begin{aligned}
\min_{(c_0, c) \in \Omega_{\text{sign}}} \; \mathcal{J}_{\lambda,\lambda_{\text{free}}}(c_0, c; \mathcal{D}) &= \mathcal{L}_{\text{emp}}(c_0, c; \mathcal{D}) + \lambda \sum_{j \in \mathcal{C}_{\text{box}}} \phi_{\text{box}}(c_j; \ell_j, u_j) \\
&\quad + \lambda \sum_{j \in \mathcal{C}_{\text{point}}} (c_j - c_j^{\text{pub}})^2 + \frac{\lambda_{\text{free}}}{2} \sum_{j \notin \mathcal{C}_{\text{box}} \cup \mathcal{C}_{\text{point}}} c_j^2.
\end{aligned}
\end{equation}
Here $\phi_{\text{box}}(c_j; \ell_j, u_j) = \max(0, \ell_j - c_j, c_j - u_j)^2$ is a continuously differentiable squared-hinge penalty outside $[\ell_j, u_j]$, and $c_0$ remains unpenalized (in NLST competing-risk evaluations, $\hat{c}_0$ is subsequently shifted by an additive constant so mean predicted risk matches training-cohort Aalen--Johansen~\cite{aalen1978empirical} cumulative incidence at the $6\text{-}$ or $10\text{-year}$ horizon; Appendix~\ref{app:protocols}). The coefficient $\lambda_{\text{free}} \ge 0$ provides optional weak per-SD ridge stabilization solely on predictors without interval or point priors ($\lambda_{\text{free}} = 0$ in primary fits and $\lambda_{\text{free}}(n) = 1/n$ in small-sample MESA subcohorts; Appendices~\ref{app:math}--\ref{app:protocols}). Taking $\lambda \to \infty$ yields the primary hard-bound estimator with exact membership $(\hat{c}_0, \hat{c}) \in \Omega_{\text{CE}}$ and zero hyperparameter tuning, whereas finite $\lambda = \Lambda_0 / n$ keeps the total prior precision $\Lambda_0$ fixed, so its influence relative to the data declines as the cohort grows. The penalty path of Fig.~\ref{fig:fig1}a,d applies the box and point penalties without the sign cone and keeps only the upstream inflammation score fixed, so $\lambda = 0$ is the otherwise unconstrained maximum-likelihood fit. By Karush--Kuhn--Tucker (KKT) optimality conditions, directional sign constraints preserve concordant empirical associations and set $\hat{c}_j = 0$ on sign-only predictors $j \in \mathcal{C}_{\text{sign}} \setminus (\mathcal{C}_{\text{box}} \cup \mathcal{C}_{\text{point}})$ whenever the empirical gradient favors the sign opposite to $\delta_j$, as when confounding reverses the association (Fig.~\ref{fig:fig_ce}b and Proposition~\ref{prop:kkt}). Restricting parameters to $\Omega_{\text{CE}}$ lowers the minimum development sample size $N_{\min}$ of Riley et al.~\cite{riley2020calculating} through two routes: fewer effective degrees of freedom ($p_{\text{eff}} \le d$) and higher anticipated discrimination of the constrained model, with the anticipated Cox--Snell $R^2$ converted from each specification's median held-out area under the receiver operating characteristic curve (AUC)~\cite{riley2021note} (equation~(\ref{eq:app_riley})).

\subsection{Analytic uncertainty propagation and selective clinical triage}\label{sec:meth_uncertainty}

At deployment, each predictor $j \in \mathcal{V}_{\text{adm}}$ in patient record $x_i$ carries mean $\mu_{ij}$, missing-data variance $v_{ij}$, and missingness indicator $m_{ij} \in \{0, 1\}$, taking $(Z_{ij}, 0, 0)$ when observed and imputed or projected moments $(\mu_{ij}, v_{ij}, 1)$ when missing. Propagation runs over either the Stage II fitted weights or the Stage I graph, which carries published weights and an intercept set to training prevalence (Appendices~\ref{app:math}--\ref{app:protocols}). With $\operatorname{expit}(u) = (1 + e^{-u})^{-1}$ denoting the logistic sigmoid, a single reverse-topological pass with out-of-fold gate parameters $\alpha_g = (\alpha_0, \alpha_m, \alpha_v) \in \mathbb{R} \times \mathbb{R}_{\ge 0}^2$ computes each predictor's reliability gate $g_{ij}(\alpha_g) \in (0, 1)$, reported as the reliability score in Results. The same pass returns the total pathway sensitivities $T_{ij}$, the summed effect of predictor $j$ on the log-odds over all directed paths to the outcome. Each link weight entering $T_{ij}$ is multiplied by the gate of its source predictor (Appendix~\ref{app:math}), so missing or uncertain predictors contribute less to the propagated log-odds mean $M_i$ and missing-data variance $V_i$:
\begin{equation}\label{eq:analytic_uncertainty}
\begin{aligned}
g_{ij}(\alpha_g) &= \operatorname{expit}(\alpha_0 - \alpha_m m_{ij} - \alpha_v v_{ij}), \\
M_i &= \sum_{j \in \mathcal{V}_{\text{adm}}} T_{ij} \mu_{ij}, \qquad V_i = \sum_{j \in \mathcal{V}_{\text{adm}}} T_{ij}^2 v_{ij} + V_{\text{lat},i},
\end{aligned}
\end{equation}
where $V_{\text{lat},i} \ge 0$ captures latent-pathway residual variance when latent nodes are modeled (Appendix~\ref{app:math}). With training-prevalence intercept variance $\sigma_b^2 = (n\bar{y}(1 - \bar{y}))^{-1}$, fixed-size Gauss--Hermite quadrature over $\xi \sim \mathcal{N}(0, 1)$ and safeguarded Newton--bisection inversion of the log-odds mixture cumulative distribution $F_i(t)$ ($t \in \mathbb{R}$; full mixture quadrature in equation~(\ref{eq:mixture_cdf}), Appendix~\ref{app:math}) yield calibrated risk $\hat{p}_i$, missing-data log-odds uncertainty $\sigma_{\text{input},i}$, and 90\% predictive interval width $\hat{W}_{90,i}$:
\begin{equation}\label{eq:variance_channels}
\begin{aligned}
\hat{p}_i &= \mathbb{E}_\xi\Bigl[\operatorname{expit}\bigl(\hat{c}_0 + M_i + (\sigma_b^2 + V_i)^{1/2}\xi\bigr)\Bigr], \qquad \sigma_{\text{input},i} = V_i^{1/2}, \\
\hat{W}_{90,i} &= \operatorname{expit}\bigl(F_i^{-1}(0.95)\bigr) - \operatorname{expit}\bigl(F_i^{-1}(0.05)\bigr).
\end{aligned}
\end{equation}
By first-order delta expansion, $\hat{W}_{90,i} \approx 3.29\,\operatorname{SD}(\eta_i)\hat{p}_i(1 - \hat{p}_i)$, where $\operatorname{SD}(\eta_i)$ is the standard deviation of the predicted log-odds, combining baseline, missing-data, and parameter variance (equation~(\ref{eq:app_variance_channels})). This width is largest near $\hat{p}_i = 0.50$ whether or not a record is complete. Uncertainty-guided chart review therefore ranks records by $\sigma_{\text{input},i}$---or its cohort-measured missing-predictor component $\sigma_{\text{miss},i}$ (Appendix~\ref{app:math})---to isolate sensitivity to unmeasured features, and two-stage CT triage reads CT features only when the $90\%$ CT-uncertainty interval, $\operatorname{logit}(\hat{p}_i) \pm 1.645\,\sigma_{\text{input},i}$ on the log-odds scale with all CT-derived predictors treated as unmeasured, contains a decision threshold (Appendix~\ref{app:protocols}).

Per-patient runtime scales linearly with the number of DAG variables $|\mathcal{V}|$ and directed links $|\mathcal{E}|$. Monte Carlo propagation with $S = 4{,}000$ draws redraws every input node and re-evaluates every link for each draw and then sorts the outcomes, in $\Theta\bigl(S(|\mathcal{V}| + |\mathcal{E}|) + S \log S\bigr)$ time. CE-Agent instead accumulates pathway sensitivities in one reverse-topological pass over a fixed parameter grid, and the integration and quantile root-finding that follow operate on a fixed-size scalar mixture whose cost does not depend on $|\mathcal{V}|$ or $|\mathcal{E}|$, giving $\mathcal{O}(|\mathcal{V}| + |\mathcal{E}|)$ time per patient. In single-record scoring, each added variable cost $0.016\text{ ms}$, against $6.49\text{ ms}$ for $4{,}000\text{-draw}$ sampling (approximately $400\text{-fold}$ lower), and total runtime was $83\text{-}$ to $377\text{-fold}$ lower (Fig.~\ref{fig:fig_scalability}a and Appendix~\ref{app:math}).

\subsection{Study cohorts, benchmarks, and statistical analysis}\label{sec:meth_cohorts}\label{sec:meth_availability}

NLST enrolled $53{,}454$ heavy smokers aged $55\text{--}74\text{ years}$ across $33$ US screening hospitals~\cite{nlst2011reduced}; we analyzed the $53{,}452$ participants in the released trial dataset. Cardiovascular death was a death whose recorded cause fell in the circulatory chapter (I00--I99) of the International Classification of Diseases, 10th revision. Causes of death came from the trial's extended follow-up, which includes National Death Index linkage through 2015; deaths from other causes were competing events. Lung cancer incidence was a confirmed diagnosis, and lung cancer death followed the trial's endpoint verification process where available and the death certificate otherwise. Cardiovascular and lung cancer analyses shared a deterministic 70\%/30\% split derived from participant identifiers into $37{,}521$ training and $15{,}931$ held-out test participants. With up to $13.4\text{ years}$ of mortality follow-up, the training set had $1{,}789$ cardiovascular and $1{,}663$ lung cancer deaths, and the test set had $750$ and $747$. Over the trial's cancer-incidence follow-up (median $6.6\text{ years}$), $1{,}440$ training and $618$ test participants developed lung cancer. In the test set, $14{,}971$ participants had uncensored 10-year vital status, including $525$ cardiovascular and $1{,}684$ competing deaths (Table~\ref{tab:nlst_main}).

Within the baseline LDCT arm, $18{,}519$ training and $7{,}786$ held-out participants experienced $732$ and $338$ full-follow-up incident lung cancers, respectively. Lung cancer evaluations assessed $6\text{-year}$ competing-risk incidence ($1{,}350$ training and $575$ held-out cases within $6\text{ years}$; LDCT arm $694$ and $317$ cases, accounting for pre-diagnosis mortality) and $10\text{-year}$ competing-risk mortality ($1{,}242$ training and $535$ held-out deaths within $10\text{ years}$) across $11$ screened baseline predictors. Joint cardiopulmonary triage paired $6\text{-year}$ lung cancer thresholds ($1.5\%$ primary and $2.0\%$ sensitivity)~\cite{tammemagi2014evaluation,tammemagi2021ontario} with the $5.0\%$ $10\text{-year}$ fatal-CVD threshold~\cite{conroy2003estimation,debacker2003european}. In the NLST 3D LDCT subcohorts ($5{,}269$ training participants with soft-tissue reconstructions and $6{,}467$ with any reconstruction kernel), thoracic aortic calcium volume, left ventricular volume, proximal main-to-sinotubular-junction PA diameter ratio, and quantitative emphysema were measured automatically on baseline chest CT from TotalSegmentator~\cite{wasserthal2023totalsegmentator} and lungmask~\cite{hofmanninger2020automatic} segmentations in a mean of $45.6\text{ s}$ per scan on one graphics processing unit (GPU; Table~\ref{tab:app_ct_timing}).

MESA~\cite{bild2002multi,detrano2008coronary} Exam 1 enrolled $6{,}814$ White, Chinese American, Black, and Hispanic adults aged $45\text{--}84\text{ years}$ free of clinical CVD at baseline, with a median follow-up of $18.9\text{ years}$ (maximum $20.9\text{ years}$) in $6{,}795$ participants. Electrocardiogram-gated cardiac CT provided five phenotypes: CAC, TAC, cardiac chamber area, ascending aortic diameter, and liver attenuation. Cardiovascular death was an adjudicated death attributed to coronary heart disease, stroke, or other atherosclerotic or cardiovascular disease; deaths from other or unknown causes were competing events. Using the same deterministic 70\%/30\% split, cardiovascular mortality analyses comprised $4{,}821$ training and $1{,}974$ held-out test participants ($4{,}731$ and $1{,}932$ with uncensored 10-year status; $49$ test cardiovascular deaths within 10 years and $147$ overall, with $168$ competing deaths within 10 years). Adjudicated composite hard ASCVD (myocardial infarction, stroke, or death from coronary heart disease or stroke) analyses comprised $4{,}812$ training and $1{,}971$ test participants ($4{,}301$ and $1{,}769$ with uncensored 10-year status; $137$ test events within 10 years and $283$ overall, with $95$ competing deaths within 10 years; Tables~\ref{tab:mesa_main} and~\ref{tab:app_composite_endpoints}). Two-stage cardiovascular CT triage was evaluated in both cohorts (Table~\ref{tab:app_interventions_triage}). The NLST analysis used baseline CVD-free LDCT participants in the training cohort ($n = 15{,}727$), each scored with published weights and an intercept estimated at the other $32$ hospitals; $1{,}256$ participants with diabetes were routed directly to statins and $14{,}471$ candidates were triaged. The MESA analysis used test adults aged $45\text{--}75\text{ years}$ not on lipid-lowering therapy at baseline ($n = 1{,}465$), with the no-CT risk recalibrated in the large to the training event rate; $178$ participants with diabetes or $\text{LDL} \ge 190\text{ mg/dL}$ ($4.9\text{ mmol/L}$) were routed directly to statins and $1{,}287$ candidates were triaged.

Clinical comparators comprised PLCOm2012~\cite{tammemagi2013selection} and 2021 USPSTF criteria~\cite{uspstf2021lung} for lung cancer, alongside Framingham general CVD~\cite{dagostino2008general}, ACC/\allowbreak AHA PCE~\cite{goff20142013}, and the AHA PREVENT base $10\text{-year}$ ASCVD equation~\cite{khan2024development} for CVD, evaluated with their published coefficients and after training-prevalence logit recalibration. The cardiovascular calculators were built for composite endpoints that include nonfatal events (general CVD for Framingham; ASCVD for PCE and PREVENT). Calibration with published coefficients was therefore assessed on MESA's adjudicated composite endpoints matched to each calculator: hard ASCVD for PCE and PREVENT, and for Framingham the broad cardiovascular composite, the closest counterpart of general CVD (Tables~\ref{tab:mesa_main} and~\ref{tab:app_composite_endpoints}). For cardiovascular death, the calculators were compared on discrimination, and their calibration is reported after recalibration. Because NLST recorded no blood pressure, lipid, or renal measurements, NLST evaluations supplied the three cardiovascular calculators with each participant's recorded age, sex, current smoking status, and diabetes history (plus race/ethnicity in PCE) alongside standard US untreated adult reference values: $197\text{ mg/dL}$ ($5.1\text{ mmol/L}$) total cholesterol, $50\text{ mg/dL}$ ($1.3\text{ mmol/L}$) high-density lipoprotein (HDL) cholesterol, $130\text{ mmHg}$ systolic blood pressure, and $90\text{ mL/min/1.73 }\text{m}^2$ estimated glomerular filtration rate (eGFR; PREVENT only). The HDL cholesterol, blood pressure, and eGFR values are the centering values of the PREVENT equations~\cite{khan2024development}. In MESA Exam 1, all three calculators used directly measured seated blood pressure, fasting lipids, and the 2009 Chronic Kidney Disease Epidemiology Collaboration (CKD-EPI) eGFR supplied with the MESA dataset~\cite{levey2009new}. In both cohorts, antihypertensive and statin indicators were set to untreated, so that each calculator returns pre-treatment risk.

Machine learning comparators comprised Cox proportional hazards (Cox PH)~\cite{cox1972regression}, random survival forests (RSF)~\cite{ishwaran2008random}, gradient-boosted survival analysis (GBSA)~\cite{friedman2001greedy}, XGBoost and XGBoost-Cox~\cite{chen2016xgboost}, DeepSurv neural networks~\cite{katzman2018deepsurv}, decision trees, random forests~\cite{breiman2001random}, and logistic regression (LR) on the CE-Agent predictor set and on expanded feature sets; single-hospital NLST and single-group MESA comparisons used the eight baselines listed in Tables~\ref{tab:nlst_ood} and~\ref{tab:mesa_ood}. Domain-generalization and prior-informed comparators included Anchor Regression~\cite{rothenhausler2021anchor}, $\chi^2$ and conditional value-at-risk (CVaR) distributionally robust optimization (DRO)~\cite{duchi2021learning}, prior-offset logistic model updating~\cite{steyerberg2004validation} and its boosted-tree extension, and width-matched Gaussian maximum a posteriori (MAP) priors (Appendix~\ref{app:protocols}). Performance across $1\text{--}15\text{ years}$ was evaluated using Harrell's $C\text{-index}$~\cite{harrell1996multivariable} for the cause-specific event (deaths from other causes censored), inverse-probability-of-censoring weighted time-dependent AUC ($t\text{-AUC}$)~\cite{uno2007evaluating}, and competing-risk $\text{AUC}_{\text{cr}}$~\cite{blanche2013estimating}. Calibration, overall accuracy, and clinical utility were summarized by competing-risk $\text{Brier}_{\text{cr}}$ and Aalen--Johansen $\text{E/O}_{\text{cr}}$, binary Brier score~\cite{brier1950verification}, logistic calibration slope~\cite{vancalster2019calibration}, and decision curve analysis (DCA) net benefit~\cite{vickers2006decision} (equation~(\ref{eq:dca})). Tests were two-sided, and 95\% confidence intervals came from $1{,}000\text{-replicate}$ participant-level bootstrap resampling unless stated otherwise. For lung cancer discrimination against recalibrated PLCOm2012, preregistered noninferiority required the lower limit of the two-sided 95\% bootstrap interval of the $\text{AUC}_{\text{cr}}$ difference to exceed $-0.02$, and for $6\text{-year}$ incidence also a positive point estimate. For two-stage against full CT reading, it required the lower limit of the net-benefit difference to exceed $-2.0$ per $1{,}000$ participants, with fewer than $70\%$ of scans read and a positive point difference from equal-budget random reading. In NLST, primary competing-risk cardiovascular evaluation, equivalence testing between compiled and point-anchored literature bounds (Table~\ref{tab:nlst_main}), and the lung cancer and joint cardiopulmonary triage analyses were preregistered in time-stamped, version-controlled analysis plans before the held-out test split was scored once on each endpoint. External MESA replications and mechanistic or pathway analyses outside those plans, including single-hospital lung cancer fits, added radiologist CT findings, cardiovascular CAC triage, LDCT-arm triage, and missing-data propagation benchmarks, are secondary, exploratory, or post hoc analyses.

\bmhead{Reporting summary}
Further information on research design, cohort eligibility, and statistical evaluation protocols is provided in Appendices~\ref{app:results}--\ref{app:protocols} and the Nature Portfolio Reporting Summary linked to this article.

\bmhead{Data availability}
Individual-level clinical and chest CT data from NLST are available to approved researchers upon application through the NCI Cancer Data Access System (\url{https://cdas.cancer.gov/nlst/}). MESA Exam 1 clinical, laboratory, CT, and adjudicated cardiovascular outcome data are available to qualified investigators through the NHLBI BioLINCC repository (\url{https://biolincc.nhlbi.nih.gov/studies/mesa/}) and the MESA Coordinating Center (\url{https://www.mesa-nhlbi.org/}). All PubMed abstracts and epidemiological effect estimates compiled in Stage I are publicly accessible and archived in the project repository.

\bmhead{Code availability}
Source code for CE-Agent, automated PubMed prior compilation, constrained convex estimation, deterministic Gauss--Hermite uncertainty propagation, figure generation, and cross-cohort benchmark reproduction is publicly available at \url{https://github.com/Yuxiang-Lai117/CP_Agent_Paper}. Baseline models and evaluation ran in Python 3.13 with NumPy 2.5.1, SciPy 1.18.0, scikit-learn 1.9.0, scikit-survival 0.28.0, lifelines 0.30.3, XGBoost 3.3.0, and PyTorch 2.13.0; CT phenotypes were extracted with TotalSegmentator 2.17.0 and lungmask 0.2.20 (Appendix~\ref{app:protocols}).

\bmhead{Acknowledgements}
We thank the participants, investigators, and study staff of the National Lung Screening Trial and the Multi-Ethnic Study of Atherosclerosis, as well as the National Cancer Institute (NCI), the National Heart, Lung, and Blood Institute (NHLBI), and the National Institutes of Health (NIH) for providing access to trial and cohort resources. The manuscript reflects the views of the authors and does not necessarily represent the official positions of the NCI, NHLBI, or NIH. This research is supported in part by the National Institutes of Health under Award Numbers 1R01HL183223, R01DE033512 and R01CA272991.

\bmhead{Author contributions}
Y.L., L.Z., and X.Y. conceived the study and designed the Causal Envelope methodology. Y.L. and J.Z. implemented the automated literature compiler, constrained estimation solver, analytic uncertainty propagation module, and multi-cohort evaluation pipelines. Y.L. and J.Z. performed the statistical, cross-institution, and clinical triage analyses across NLST and MESA. Y.L. drafted the manuscript and designed the visualizations with critical revision from J.Z., L.Z., and X.Y. L.Z. and X.Y. supervised the project. All authors approved the final manuscript.

\bmhead{Competing interests}
The authors declare no competing interests.

\bibliography{sn-bibliography}%

\clearpage
\begin{appendices}
\numbered
\renewcommand{\topfraction}{0.95}
\renewcommand{\bottomfraction}{0.95}
\renewcommand{\textfraction}{0.04}
\renewcommand{\floatpagefraction}{0.82}
\setlength{\intextsep}{8pt plus 2pt minus 2pt}
\setlength{\textfloatsep}{10pt plus 2pt minus 2pt}

\section{Extended results}\label{app:results}

{\footnotesize\parfillskip=0pt plus 0.65\textwidth\emergencystretch=1.5em Abbreviations in Tables~\ref{tab:nlst_main}--\ref{tab:app_dual_triage}: AUC, area under the receiver operating characteristic curve; $t\text{-AUC}$, time-dependent AUC; $C$, Harrell's concordance index; E/O, expected-to-observed ratio; cr, competing-risk; CAC, coronary artery calcification; LDCT, low-dose CT; PCE, Pooled Cohort Equations; LR, logistic regression; RSF, random survival forest; XGB, XGBoost; OOD, out-of-distribution; $\hat{W}_{90}$, $90\%$ interval width; $\sigma_{\text{input}}$, missing-input log-odds uncertainty; USPSTF, US Preventive Services Task Force.\par}

\subsection{Held-out accuracy, composite endpoints, and transfer}

Tables~\ref{tab:nlst_main}--\ref{tab:app_zeroshot_transport} extend Fig.~\ref{fig:fig1}. They compare CE-Agent with clinical calculators, its own point-anchored and prior-locked variants, and seven survival and ML model families, on held-out cardiovascular mortality and adjudicated composite endpoints; the last table applies NLST-trained models to MESA.

\begin{table}[!htbp]
\centering
\caption{\textbf{NLST test-cohort cardiovascular mortality: CE-Agent versus ML baselines and clinical calculators ($n = 15{,}931$).} Ten-year Aalen--Johansen competing-risk $\text{AUC}_{\text{cr}}$ ($525$ cardiovascular deaths and $1{,}684$ competing deaths), full-follow-up $C\text{-index}$ ($750$ cardiovascular deaths over $13.4\text{ yr}$), Brier score, and $10\text{-yr}$ $\text{E/O}_{\text{cr}}$ across CE-Agent specifications, pooled $13\text{-predictor}$ survival and ML baselines, and clinical calculators (extended metrics in Table~\ref{tab:app_frontier_benchmarks}).}\label{tab:nlst_main}
{\tabfont
\setlength{\tabcolsep}{0pt}
\renewcommand{\arraystretch}{1.04}
\begin{tabular*}{\textwidth}{@{\extracolsep{\fill}}lcccc@{}}
\toprule
\colh{Model / Specification} & \colh{$\text{AUC}_{\text{cr}}$ (95\% CI)} & \colh{$C$-index (95\% CI)} & \colh{Brier} & \colh{$\text{E/O}_{\text{cr}}$} \\
\midrule
CE-Agent (compiled hard bounds, primary) & 0.723 (0.700--0.744) & 0.731 (0.713--0.749) & \textbf{0.0433} & \textbf{1.02} \\
CE-Agent (compiled soft bounds, CV $\lambda$) & 0.723 (0.700--0.744) & 0.731 (0.713--0.749) & \textbf{0.0433} & \textbf{1.02} \\
CE-Agent (point-anchored bounds)$^*$ & 0.721 (0.700--0.742) & 0.729 (0.711--0.747) & 0.0436 & \textbf{1.02} \\
CE-Agent (prior-locked) & 0.717 (0.696--0.738) & 0.725 (0.707--0.744) & 0.0435 & \textbf{1.02} \\
\addlinespace[2pt]
Logistic Regression & \textbf{0.725} (0.702--0.747) & 0.734 (0.713--0.749) & \textbf{0.0433} & 1.06 \\
XGBoost Classifier & 0.724 (0.702--0.745) & 0.735 (0.717--0.751) & 0.0434 & 1.04 \\
Random Forest / RSF & \textbf{0.725} (0.703--0.746) & \textbf{0.736} (0.716--0.752) & ---$^\dagger$ & 1.05 \\
Cox Proportional Hazards & 0.719 (0.697--0.740) & 0.728 (0.706--0.746) & ---$^\dagger$ & 1.04 \\
\addlinespace[2pt]
ACC/AHA ASCVD-PCE & 0.683 (0.659--0.705) & 0.693 (0.672--0.709) & ---$^\ddagger$ & ---$^\ddagger$ \\
AHA PREVENT ($10\text{-yr}$) & 0.680 (0.655--0.703) & 0.689 (0.668--0.706) & ---$^\ddagger$ & ---$^\ddagger$ \\
Framingham General CVD & 0.667 (0.644--0.690) & 0.674 (0.653--0.691) & ---$^\ddagger$ & ---$^\ddagger$ \\
\botrule
\end{tabular*}%
}
\tabnote{Note: Bold marks the best value in each column ($|\text{E/O}_{\text{cr}} - 1|$ closest to zero for calibration). $95\%$ CIs are percentile bootstrap intervals over held-out participants ($C\text{-index}$: $20{,}000$ replicates for CE-Agent rows and $200$ for baselines and calculators, in place of the default $1{,}000$). Brier, full-follow-up binary Brier score; RSF, random survival forest; ACC/AHA ASCVD-PCE, American College of Cardiology/American Heart Association Pooled Cohort Equations; PREVENT, Predicting Risk of Cardiovascular Disease Events; CV $\lambda$, soft-bound penalty selected by five-fold inner cross-validation; point-anchored bounds, established clinical factors anchored at rescaled literature point estimates, with imaging and pulmonary predictors bounded in $[\ell_j, u_j]$; prior-locked, weights fixed at published point estimates without outcome fitting. $^*$Preregistered equivalence of compiled and point-anchored bounds (compiled minus point-anchored; paired bootstrap $95\%$ CI): $\Delta C = +0.0016$ ($-0.0039$ to $+0.0072$; bounds $\pm 0.01$) and $\Delta\text{Brier} = -3.2 \times 10^{-4}$ ($-5.16 \times 10^{-4}$ to $-1.30 \times 10^{-4}$; bounds $\pm 0.001$); both intervals lie within their bounds. $^\dagger$The $10\text{-yr}$ competing-risk Brier score ($\text{Brier}_{\text{cr}}$) is $0.0320$ for compiled CE-Agent, logistic regression, and random forest, $0.0321$ for XGBoost, and $0.0322$ for Cox PH. For logistic regression, XGBoost, and the forest row, $C\text{-index}$ and Brier come from models trained on full-follow-up labels, and $\text{AUC}_{\text{cr}}$ and $\text{E/O}_{\text{cr}}$ from classifiers trained on $10\text{-yr}$ status; the forest row pairs a random survival forest with a random forest classifier. CE-Agent competing-risk predictions apply a logit shift aligned to the training-cohort Aalen--Johansen $10\text{-yr}$ incidence; Cox PH reports competing-risk cumulative incidence. Clinical calculators use published coefficients, with US untreated adult reference values for unmeasured blood pressure, lipids, and eGFR. $^\ddagger$They target composite endpoints that include nonfatal events, so Brier score and $\text{E/O}_{\text{cr}}$ are not scored against cardiovascular death (matched-endpoint calibration in Tables~\ref{tab:mesa_main} and~\ref{tab:app_composite_endpoints}); after training-prevalence logit recalibration all three reach $\text{E/O}_{\text{cr}} = 1.02$ (Methods).}
\end{table}

\begin{table}[!htbp]
\centering
\caption{\textbf{MESA test-cohort cardiovascular mortality and hard ASCVD: CE-Agent versus unconstrained models and clinical calculators.} Ten-year discrimination ($\text{AUC}_{10}$) and Brier score on uncensored $10\text{-yr}$ status ($n = 1{,}932$ for mortality; $1{,}769$ for hard ASCVD), full-follow-up $C\text{-index}$, and $10\text{-yr}$ competing-risk $\text{E/O}_{\text{cr}}$ ($168$ and $95$ competing $10\text{-yr}$ deaths; extended benchmarks in Fig.~\ref{fig:fig1}f and Tables~\ref{tab:app_mesa_frontier}--\ref{tab:app_composite_endpoints}).}\label{tab:mesa_main}
{\tabfont
\setlength{\tabcolsep}{0pt}
\renewcommand{\arraystretch}{1.14}
\begin{tabular*}{\textwidth}{@{\extracolsep{\fill}}lcccc@{}}
\toprule
\colh{Model / Specification} & \colh{$\text{AUC}_{10}$ (95\% CI)} & \colh{$C$-index (95\% CI)} & \colh{Brier} & \colh{$\text{E/O}_{\text{cr}}$} \\
\midrule
\multicolumn{5}{@{}l}{\textit{Cardiovascular mortality ($n = 1{,}974$; $49$ $10\text{-yr}$ and $147$ total deaths)}} \\
CE-Agent (hard, $\ln(\text{CAC}+1)$, primary) & 0.807 (0.758--0.855) & 0.805 (0.773--0.835) & 0.0239 & 1.08 \\
CE-Agent (soft, $\ln(\text{CAC}+1)$) & 0.805 (0.752--0.854) & \textbf{0.807} (0.776--0.838) & 0.0240 & 1.08 \\
CE-Agent (hard, binary CAC) & 0.811 (0.763--0.858) & 0.804 (0.772--0.833) & \textbf{0.0237} & ---$^\dagger$ \\
CE-Agent (prior-locked) & 0.794 (0.738--0.847) & 0.801 (0.767--0.832) & 0.0255 & 1.09 \\
Unconstrained $22\text{-Predictor}$ Model & 0.789 (0.729--0.842) & 0.795 (0.761--0.829) & 0.0245 & 1.08 \\
Cox Proportional Hazards & 0.797 (0.737--0.852) & 0.801 (0.767--0.836) & \textbf{0.0237} & \textbf{1.07} \\
\addlinespace[2pt]
ACC/AHA ASCVD-PCE & 0.813 (0.767--0.857) & 0.795 (0.765--0.825) & ---$^\ddagger$ & ---$^\ddagger$ \\
AHA PREVENT ($10\text{-yr}$) & \textbf{0.817} (0.769--0.865) & 0.804 (0.774--0.836) & ---$^\ddagger$ & ---$^\ddagger$ \\
Framingham General CVD & 0.815 (0.764--0.864) & 0.772 (0.740--0.804) & ---$^\ddagger$ & ---$^\ddagger$ \\
\midrule
\multicolumn{5}{@{}l}{\textit{Composite hard ASCVD ($n = 1{,}971$; $137$ $10\text{-yr}$ and $283$ total events)}} \\
CE-Agent (hard bounds, primary) & 0.774 (0.738--0.811) & 0.753 (0.727--0.780) & 0.0682 & 1.04 \\
CE-Agent (soft bounds) & \textbf{0.776} (0.742--0.815) & \textbf{0.756} (0.729--0.783) & \textbf{0.0663} & \textbf{1.02} \\
Unconstrained $22\text{-Predictor}$ Model & 0.773 (0.739--0.813) & 0.752 (0.725--0.779) & 0.0665 & \textbf{1.02} \\
Cox Proportional Hazards & 0.772 (0.735--0.810) & 0.752 (0.725--0.778) & 0.0665 & 0.96 \\
ACC/AHA ASCVD-PCE & 0.750 (0.711--0.785) & 0.730 (0.701--0.758) & 0.0724 & 1.70 \\
\botrule
\end{tabular*}%
}
\tabnote{Note: Bold marks the best value in each column per endpoint ($|\text{E/O}_{\text{cr}} - 1|$ closest to zero for calibration). CAC, coronary artery calcification; ASCVD, atherosclerotic cardiovascular disease; ACC/AHA ASCVD-PCE, American College of Cardiology/American Heart Association ASCVD Pooled Cohort Equations; PREVENT, Predicting Risk of Cardiovascular Disease Events. $^\dagger$Binary CAC coding has complete-case $10\text{-yr}$ $\text{E/O} = 1.07$ (Table~\ref{tab:app_mesa_frontier}); competing-risk $\text{E/O}_{\text{cr}}$ was evaluated on the primary continuous $\ln(\text{CAC}+1)$ specification with training-cohort Aalen--Johansen logit anchoring. On mortality, the $10\text{-yr}$ discrimination of CE-Agent is comparable to that of the published clinical equations ($P > 0.40$). $^\ddagger$The published equations target composite cardiovascular endpoints that include nonfatal events, so their Brier score and $\text{E/O}_{\text{cr}}$ are reported on the matched endpoint (PCE on hard ASCVD, lower block, where published PREVENT has $\text{E/O}_{\text{cr}} = 0.87$; Framingham in Table~\ref{tab:app_composite_endpoints}) and not against cardiovascular death; after training-prevalence logit recalibration the equations reach $\text{E/O}_{\text{cr}} = 1.05\text{--}1.06$ on mortality and $0.94$ for PCE on hard ASCVD. Cox PH reports competing-risk cumulative incidence. $22\text{-predictor}$ vs.\ $58\text{-predictor}$ ($63\text{-column}$) ML benchmarks are in Table~\ref{tab:app_mesa_22v53}.}
\end{table}

\begin{table}[!htbp]
\centering
\caption{\textbf{NLST test-cohort cardiovascular mortality: time-dependent discrimination, calibration, and net benefit ($n = 15{,}931$).} Bold marks overall column-best values ($|{\text{slope}} - 1|$ and $|\text{E/O}_{\text{cr}} - 1|$ closest to zero; $1\text{--}10\text{ yr}$ $t\text{-AUC}(t)$ curves in Fig.~\ref{fig:fig1}c).}\label{tab:app_frontier_benchmarks}
{\tabfont
\setlength{\tabcolsep}{0pt}
\renewcommand{\arraystretch}{1.10}
\begin{tabular*}{\textwidth}{@{\extracolsep{\fill}}lcccc@{}}
\toprule
\colh{Model / Specification} & \colh{$\text{AUC}_{\text{cr}}$ (95\% CI)} & \colh{$t\text{-AUC}_{\text{cr}}$ / CS} & \colh{Slope ($\text{E/O}_{\text{cr}}$)} & \colh{$\text{NB}_{5\%}$ ($10^{-3}$)} \\
\midrule
CE-Agent (hard, primary) & 0.7225 (0.700--0.744) & 0.7212 / 0.7386 & 0.951 (\textbf{1.02}) & 6.8 \\
CE-Agent (soft, CV $\lambda$) & 0.7225 (0.700--0.744) & 0.7212 / 0.7386 & 0.951 (\textbf{1.02}) & 6.8 \\
CE-Agent (sign-only) & 0.7225 (0.700--0.744) & 0.7213 / 0.7387 & 0.950 (\textbf{1.02}) & 6.8 \\
CE-Agent (point-anchored) & 0.7208 (0.700--0.742) & 0.7215 / 0.7379 & 0.805 (\textbf{1.02}) & 6.4 \\
CE-Agent (prior-locked) & 0.7166 (0.696--0.738) & 0.7179 / 0.7350 & 0.773 (\textbf{1.02}) & 6.6 \\
\addlinespace[2pt]
Random Forest (pooled) & \textbf{0.7251} (0.703--0.746) & \textbf{0.7236} / \textbf{0.7410} & 1.143 (1.05) & 6.4 \\
Logistic Regression (pooled) & 0.7248 (0.702--0.747) & 0.7223 / 0.7397 & 0.934 (1.06) & 6.5 \\
XGBoost Classifier (pooled) & 0.7240 (0.702--0.745) & 0.7234 / 0.7404 & 0.865 (1.04) & \textbf{6.9} \\
Cox PH (CIF, pooled) & 0.7188 (0.697--0.740) & 0.7186 / 0.7355 & 1.958 (1.04) & 5.5 \\
\addlinespace[2pt]
ACC/AHA ASCVD-PCE & 0.6834 (0.659--0.705) & 0.6816 / 0.6973 & 1.060 (---)$^*$ & 4.9$^*$ \\
AHA PREVENT ($10\text{-yr}$) & 0.6801 (0.655--0.703) & 0.6778 / 0.6943 & 1.546 (---)$^*$ & 4.4$^*$ \\
Framingham General CVD & 0.6675 (0.644--0.690) & 0.6659 / 0.6790 & \textbf{0.992} (---)$^*$ & 3.8$^*$ \\
\botrule
\end{tabular*}%
}
\tabnote{Note: Int.\ $t\text{-AUC}_{\text{cr}}$ / CS, integrated competing-risk and cause-specific time-dependent $\text{AUC}$ across $1\text{--}13\text{ yr}$; $\text{E/O}_{\text{cr}}$, $10\text{-yr}$ Aalen--Johansen competing-risk expected-to-observed ratio ($\text{E/O} = 0.99$ on $n = 14{,}971$ uncensored $10\text{-yr}$ status across all CE-Agent models); CIF, competing-risk cumulative incidence function (unadjusted cause-specific Cox PH: $\text{AUC}_{\text{cr}} = 0.7192$, slope $1.797$, $\text{E/O}_{\text{cr}} = 1.11$, $\text{NB}_{5\%} = 6.2$); $\text{NB}_{5\%}$, net benefit ($\times 10^{-3}$) at risk threshold $p_t = 5\%$ ($95\%$ CI $4.7\text{--}8.8$ for hard/soft CE-Agent); ICI, integrated calibration index ($0.0019$ for CE-Agent, $0.0024\text{--}0.0109$ for pooled ML, and $0.0010\text{--}0.0066$ for recalibrated calculators). Discrimination of CE-Agent is comparable to that of the $33\text{-center}$ pooled ML models ($P \ge 0.30$ across paired replicates; full-follow-up $C\text{-index}$ $0.7306$ for CE-Agent, $0.7327\text{--}0.7337$ for $10\text{-yr}$-fitted pooled LR/XGBoost/RF, and $0.7278$ for Cox PH). $^*$The published calculators target composite cardiovascular endpoints that include nonfatal events, so $\text{E/O}_{\text{cr}}$ against cardiovascular death is not reported for them ($1.02$ after training-prevalence recalibration; Table~\ref{tab:nlst_main}), and their $\text{NB}_{5\%}$ and ICI are given after recalibration.}
\end{table}

\begin{table}[!htbp]
\centering
\caption{\textbf{MESA test-cohort cardiovascular mortality: CE-Agent specifications versus clinical calculators ($n = 1{,}974$).} Bold values denote overall column-best values across non-ablation models ($|{\text{slope}} - 1|$ and $|\text{E/O} - 1|$ closest to zero); survival-increment-weighted integrated $t\text{-AUC}$ spans $1\text{--}15\text{ yr}$ (horizon curves across $2\text{--}15\text{ yr}$ in Fig.~\ref{fig:fig1}f). Rightmost column reports calibration slope alongside complete-case $10\text{-yr}$ $\text{E/O}$ (competing-risk $\text{E/O}_{\text{cr}}$ in Table~\ref{tab:mesa_main}).}\label{tab:app_mesa_frontier}
{\tabfont
\setlength{\tabcolsep}{0pt}
\renewcommand{\arraystretch}{1.20}
\begin{tabular*}{\textwidth}{@{\extracolsep{\fill}}lcccc@{}}
\toprule
\colh{Model / Specification} & \colh{$\text{AUC}_{10}$ (95\% CI)} & \colh{$C_{10}$ / $C_{\text{full}}$} & \colh{Int.\ $t\text{-AUC}$} & \colh{Slope (E/O)} \\
\midrule
CE-Agent (hard, $\ln(\text{CAC}+1)$)$^\dagger$ & 0.8068 (0.758--0.855) & 0.8157 / 0.8046 & 0.8423 & 0.777 (1.07) \\
CE-Agent (soft, $\ln(\text{CAC}+1)$) & 0.8051 (0.752--0.854) & 0.8145 / \textbf{0.8074} & 0.8411 & 0.833 (1.07) \\
CE-Agent (hard, binary CAC) & 0.8107 (0.763--0.858) & 0.8194 / 0.8035 & 0.8418$^\S$ & 0.826 (1.07) \\
CE-Agent (hard, ordinal CAC) & 0.8072 (0.759--0.855) & 0.8162 / 0.8032 & --- & 0.788 (\textbf{1.06}) \\
CE-Agent (prior-locked) & 0.7940 (0.738--0.847) & 0.8032 / 0.8005 & 0.8416 & 0.466 (1.08) \\
Unconstrained $22\text{-Predictor}$ Model & 0.7886 (0.729--0.842) & 0.7979 / 0.7952 & 0.8246 & 0.768 (1.07) \\
Zero-Locked Ridge ($8$ free) & 0.7130 (0.642--0.779) & 0.7211 / 0.7282 & --- & 0.726 (1.04)$^*$ \\
\addlinespace[2pt]
AHA PREVENT ($10\text{-yr}$) & \textbf{0.8174} (0.769--0.865) & \textbf{0.8273} / 0.8044 & \textbf{0.8428} & 1.742 (---)$^\ddagger$ \\
ACC/AHA ASCVD-PCE & 0.8131 (0.767--0.857) & 0.8209 / 0.7945 & 0.8306 & \textbf{1.096} (---)$^\ddagger$ \\
Framingham General CVD & 0.8149 (0.764--0.864) & 0.8204 / 0.7720 & 0.8306 & 1.149 (---)$^\ddagger$ \\
\botrule
\end{tabular*}%
}
\tabnote{Note: $C_{10}$, $10\text{-yr}$ Harrell's $C\text{-index}$; $C_{\text{full}}$, full-follow-up Harrell's $C\text{-index}$; $^\dagger$primary specification; $^\S$evaluated under the locked binary CAC prior ($C_{\text{full}} = 0.805$; Fig.~\ref{fig:fig3}g). $^*$Zero-locked ridge is a control ablation that shrinks the $13$ anchored and interval-bounded clinical and literature predictors toward zero (inflammation locked at its published point weight) while fitting the $8$ unconstrained predictors ($p_{\text{eff}} = 8.0$). Kaplan--Meier~\cite{kaplan1958nonparametric} net $\text{E/O}$ is $1.03$ for CE-Agent and unconstrained fits, $1.04$ prior-locked, and $1.01\text{--}1.02$ for the calculators after prevalence recalibration. $^\ddagger$The published calculators target composite cardiovascular endpoints that include nonfatal events, so their $\text{E/O}$ against cardiovascular death is not reported; calibration on matched endpoints is in Tables~\ref{tab:mesa_main} and~\ref{tab:app_composite_endpoints}. Brier score is $0.02389$ for ordinal CAC and $0.02459$ for zero-locked ridge (other models in Table~\ref{tab:mesa_main}). Integrated cause-specific $t\text{-AUC}$ spans $1\text{--}15\text{ yr}$, weighted by Kaplan--Meier survival increments (competing-risk integrated $t\text{-AUC}_{\text{cr}} = 0.8186$ vs.\ $0.7986$ unconstrained); hard-bound CE-Agent reaches $0.842$, PREVENT $0.843$, and unconstrained estimation $0.825$ (difference $+0.018$; $95\%$ CI $-0.004$ to $+0.040$). Dashes, integrated $t\text{-AUC}$ not computed.}
\end{table}

\begin{table}[!htbp]
\centering
\caption{\textbf{MESA cardiovascular mortality: seven model families on the $22\text{-predictor}$ CE-Agent set versus an expanded $58\text{-predictor}$ ($63\text{-column}$) Exam 1 set ($n = 1{,}974$).} Columns compare $22\text{-predictor}$ / $58\text{-predictor}$ test performance ($14$ model-by-feature configurations) alongside $22\text{-predictor}$ integrated $t\text{-AUC}$ ($1\text{--}15\text{ yr}$) and Brier score.}\label{tab:app_mesa_22v53}
{\tabfont
\setlength{\tabcolsep}{0pt}
\renewcommand{\arraystretch}{1.20}
\begin{tabular*}{\textwidth}{@{\extracolsep{\fill}}lccccc@{}}
\toprule
\colh{ML Architecture} & \colh{$\text{AUC}_{10}$ (22/58)} & \colh{$C_{10}$ (22/58)} & \colh{Int.\ $t\text{-AUC}$} & \colh{Brier} & \colh{Slope (22/58)} \\
\midrule
Cox Proportional Hazards & \textbf{0.7972} / \textbf{0.8108} & \textbf{0.8059} / \textbf{0.8216} & \textbf{0.8373} & \textbf{0.02367} & 1.401 / \textbf{0.937} \\
Random Survival Forest & 0.7832 / 0.7855 & 0.7970 / 0.8080 & 0.8236 & 0.02391 & \textbf{1.036} / 1.086 \\
Logistic Regression & 0.7876 / 0.7510 & 0.7962 / 0.7614 & 0.8209 & 0.02481 & 0.732 / 0.520 \\
Cox-Boosted Trees & 0.7793 / 0.8009 & 0.7907 / 0.8121 & 0.8192 & 0.02507 & 0.641 / 0.613 \\
DeepSurv Neural Net & 0.7760 / 0.7601 & 0.7857 / 0.7708 & 0.8067 & 0.02504 & 0.617 / 0.329 \\
XGBoost Classifier & 0.7658 / 0.7411 & 0.7753 / 0.7521 & 0.8017 & 0.02454 & 0.646 / 0.538 \\
Decision Tree & 0.7133 / 0.6803 & 0.7204 / 0.6881 & 0.7362 & 0.02539 & 0.594 / 0.253 \\
\botrule
\end{tabular*}%
}
\tabnote{Note: Bold values denote column-best baseline performance ($|\text{slope} - 1|$ closest to zero for calibration). Abbreviations: $C_{10}$, $10\text{-yr}$ Harrell's $C\text{-index}$; Int.\ $t\text{-AUC}$, integrated cause-specific time-dependent $\text{AUC}$ ($1\text{--}15\text{ yr}$, weighted by Kaplan--Meier survival increments as in Tables~\ref{tab:app_mesa_frontier} and~\ref{tab:app_composite_endpoints}) across the $22$ predictors.}
\end{table}

\begin{table}[!htbp]
\centering
\caption{\textbf{MESA adjudicated composite endpoints: CE-Agent versus PCE and Framingham ($n = 1{,}971$).} Hard ASCVD is defined in Methods; hard CVD is the adjudicated hard cardiovascular endpoint released with MESA; broad CVD comprises myocardial infarction, stroke, congestive heart failure, or cardiovascular death. Group headers list $10\text{-yr}$ / full-follow-up event counts and competing $10\text{-yr}$ non-cardiovascular deaths (uncensored $10\text{-yr}$ status: $n = 1{,}769$ for hard ASCVD and hard CVD; $n = 1{,}778$ for broad CVD).}\label{tab:app_composite_endpoints}
{\tabfont
\setlength{\tabcolsep}{0pt}
\renewcommand{\arraystretch}{1.20}
\begin{tabular*}{\textwidth}{@{\extracolsep{\fill}}lccccc@{}}
\toprule
\colh{Endpoint \& Specification} & \colh{$\text{AUC}_{10}$ (95\% CI)} & \colh{$C_{\text{full}}$} & \colh{Int.\ $t\text{-AUC}$} & \colh{Brier} & \colh{Slope (E/O)} \\
\midrule
\multicolumn{6}{@{}l}{\textit{Hard ASCVD ($137 / 283$ events; $95$ competing $10\text{-yr}$ non-CVD deaths)}} \\
CE-Agent (hard bounds) & 0.7740 (0.738--0.811) & 0.7531 & \textbf{0.8020} & 0.06820 & 0.706 (\textbf{0.93}) \\
CE-Agent (soft bounds) & \textbf{0.7765} (0.742--0.815) & \textbf{0.7561} & 0.8019 & \textbf{0.06630} & \textbf{0.904} (0.92) \\
ACC/AHA ASCVD-PCE & 0.7503 (0.711--0.785) & 0.7302 & 0.7774 & 0.07241 & 0.799 (1.55) \\
\midrule
\multicolumn{6}{@{}l}{\textit{Hard CVD ($138 / 287$ events; $94$ competing $10\text{-yr}$ non-CVD deaths)}} \\
CE-Agent (hard bounds) & \textbf{0.7740} (0.738--0.811) & \textbf{0.7520} & \textbf{0.8012} & \textbf{0.06878} & 0.701 (\textbf{0.95}) \\
ACC/AHA ASCVD-PCE & 0.7475 (0.709--0.782) & 0.7271 & 0.7725 & 0.07291 & \textbf{0.790} (1.54) \\
\midrule
\multicolumn{6}{@{}l}{\textit{Broad CVD ($181 / 371$ events; $85$ competing $10\text{-yr}$ non-CVD deaths)}} \\
CE-Agent (hard bounds) & \textbf{0.7870} (0.755--0.817) & \textbf{0.7625} & \textbf{0.8132} & \textbf{0.08474} & 0.721 (\textbf{0.97}) \\
ACC/AHA ASCVD-PCE & 0.7512 (0.718--0.783) & 0.7363 & 0.7762 & 0.08746 & 0.806 (1.18) \\
Framingham General CVD & 0.7635 (0.731--0.794) & 0.7362 & 0.7885 & 0.08655 & \textbf{1.012} (1.44) \\
\botrule
\end{tabular*}%
}
\tabnote{Note: Bold marks the best value in each column per endpoint ($|\text{slope} - 1|$ and $|\text{E/O} - 1|$ closest to zero for calibration). On hard ASCVD, the integrated $t\text{-AUC}$ of CE-Agent (hard bounds) exceeds that of PCE by $0.025$ ($P = 0.010$). $10\text{-yr}$ concordance $C_{10}$ equals $0.7671$ (hard) and $0.7692$ (soft) vs.\ $0.7459$ (PCE) on Hard ASCVD, $0.7665$ vs.\ $0.7426$ on Hard CVD, and $0.7776$ vs.\ $0.7442$ (PCE) and $0.7567$ (Framingham) on Broad CVD; the rightmost column reports complete-case $10\text{-yr}$ $\text{E/O}$ (competing-risk $\text{E/O}_{\text{cr}}$ on hard ASCVD in Table~\ref{tab:mesa_main}). PCE targets hard ASCVD; Framingham targets general CVD, for which broad CVD is the closest adjudicated composite, and on which the integrated $t\text{-AUC}$ of CE-Agent exceeds that of Framingham by $0.025$ ($P = 0.012$).}
\end{table}

\begin{table}[!htbp]
\centering
\caption{\textbf{NLST-trained models applied to MESA cardiovascular mortality on $12$ shared predictors ($n = 1{,}974$).} Models are trained exclusively on NLST ($n = 37{,}521$) and evaluated on the MESA test cohort without refitting weights, recalibrating only the intercept to MESA prevalence.}\label{tab:app_zeroshot_transport}
{\tabfont
\setlength{\tabcolsep}{0pt}
\renewcommand{\arraystretch}{1.20}
\begin{tabular*}{\textwidth}{@{\extracolsep{\fill}}lcccc@{}}
\toprule
\colh{Transported Architecture} & \colh{$\text{AUC}_{10}$ (95\% CI)} & \colh{$C_{\text{full}}$} & \colh{Brier} & \colh{Slope (E/O)} \\
\midrule
CE-Agent (compiled hard bounds) & \textbf{0.8029} (0.750--0.851) & \textbf{0.8066} & 0.02372 & 1.126 (1.06) \\
CE-Agent (point-anchored soft bounds) & 0.7961 (0.741--0.850) & 0.8052 & 0.02376 & 0.954 (1.07) \\
CE-Agent (point-anchored hard bounds) & 0.7940 (0.739--0.847) & 0.8043 & 0.02379 & 0.913 (1.07) \\
\addlinespace[2pt]
Unconstrained Logistic Reg. & \textbf{0.8029} (0.750--0.852) & 0.8023 & \textbf{0.02371} & \textbf{1.011} (1.06) \\
DeepSurv Neural Network & 0.8022 (0.748--0.852) & 0.8030 & \textbf{0.02371} & 1.026 (1.06) \\
Cox Proportional Hazards & 0.7921 (0.735--0.845) & 0.7977 & 0.02403 & 2.107 (\textbf{1.05}) \\
Cox-Boosted Trees (XGB-Cox) & 0.7761 (0.718--0.832) & 0.7866 & 0.02408 & 0.871 (\textbf{1.05}) \\
Random Survival Forest (RSF) & 0.7713 (0.710--0.827) & 0.7817 & --- & --- \\
XGBoost Classifier & 0.7699 (0.715--0.823) & 0.7807 & 0.02425 & 0.927 (\textbf{1.05}) \\
Classification Decision Tree & 0.7674 (0.720--0.814) & 0.7554 & 0.02420 & 1.286 (\textbf{1.05}) \\
\botrule
\end{tabular*}%
}
\tabnote{Note: Bold values denote overall column-best performance ($|\text{slope} - 1|$ and $|\text{E/O} - 1|$ closest to zero); $C_{\text{full}}$ denotes full-follow-up Harrell's $C\text{-index}$ and E/O denotes complete-case $10\text{-yr}$ expected-to-observed ratio (--- on RSF denotes nonparametric hazard ranking without a parametric logit intercept for prevalence recalibration).}
\end{table}

\FloatBarrier
\subsection{Generalization, sample size, and clinical utility}

\begin{figure}[!htbp]
\centering
\includegraphics[width=\textwidth]{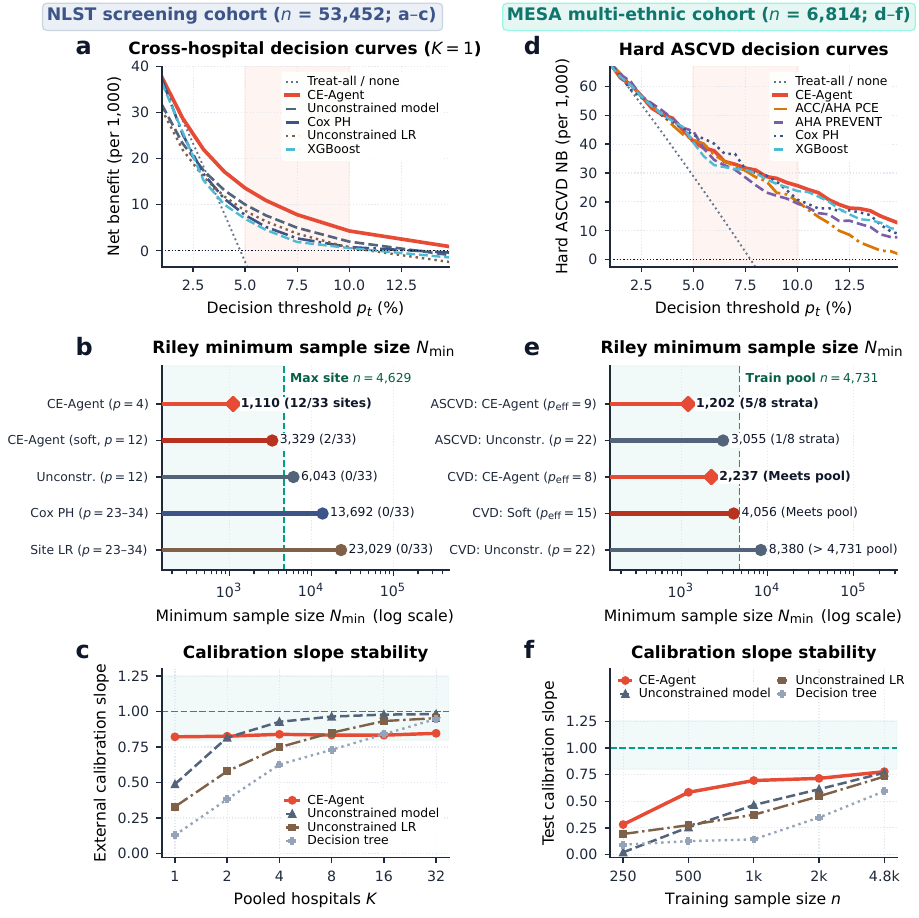}
\caption{\textbf{Clinical utility, minimum sample size, and calibration stability in NLST (a--c) and MESA (d--f).}
\textbf{a},\textbf{d}, Decision curves for CE-Agent, unconstrained models, Cox PH, XGBoost, clinical calculators (\textbf{d}), and treat-all or treat-none strategies in single-hospital ($K = 1$ hospital) NLST external evaluation (\textbf{a}) and held-out MESA hard ASCVD (\textbf{d}; NB, net benefit; $p_t$, decision threshold; shading, $p_t$ of $5\%$ to $10\%$).
\textbf{b},\textbf{e}, Minimum sample size $N_{\min}$ under the criteria of Riley et al.~\cite{riley2020calculating} as constraints are added, in NLST (\textbf{b}; $p$, free slope parameters; soft, $\lambda \ge 1$; Site LR, site-level logistic regression; Max site, largest single hospital) and for MESA hard ASCVD and cardiovascular mortality (\textbf{e}; $p_{\text{eff}}$, effective number of free parameters; Unconstr., unconstrained model; Train pool, training participants with known 10-year status; strata, the four racial/ethnic groups and four scanner cohorts).
\textbf{c},\textbf{f}, Calibration slopes as training data grow from $1\text{ to }32$ pooled NLST hospitals (\textbf{c}) and from $250\text{ to }4{,}821$ MESA participants (\textbf{f}); band, $0.80$ to $1.25$. Unconstrained model, CE-Agent's predictors fitted without published bounds; Unconstrained LR, logistic regression baseline.}\label{fig:fig5}
\end{figure}

Figure~\ref{fig:fig5} and Tables~\ref{tab:nlst_ood}--\ref{tab:app_mesa_ood_licensing} extend Fig.~\ref{fig:fig2}. They report out-of-distribution (OOD) performance across hospitals, racial/ethnic groups, and scanners, scaling with training size, minimum sample size~\cite{riley2020calculating}, and net benefit~\cite{vickers2006decision}.

\begin{table}[!htbp]
\centering
\caption{\textbf{NLST models trained at one hospital and tested at the other $32$: discrimination, minimum sample size, and net benefit ($n = 37{,}521$).} Each architecture has $1{,}056$ single-hospital external evaluations; multi-hospital scaling is in Fig.~\ref{fig:fig2}b and Table~\ref{tab:app_ood_licensing}.}\label{tab:nlst_ood}
{\tabfont
\setlength{\tabcolsep}{0pt}
\renewcommand{\arraystretch}{1.12}
\begin{tabular*}{\textwidth}{@{\extracolsep{\fill}}lcccc@{}}
\toprule
\colh{Model Architecture} & \colh{$\text{AUC}$ (IQR)} & \colh{CE-Agent Wins} & \colh{$N_{\min}$} & \colh{Excess NB ($10^{-3}$)} \\
\midrule
CE-Agent (point-anchored bounds) & \textbf{0.727} (0.726--0.727) & \textbf{264/264} & \textbf{1,110} & \textbf{0.83} \\
Random Survival Forest & 0.686 (0.667--0.703) & $33/33^{***}$ & 13,275 & --- \\
Cox Proportional Hazards & 0.683 (0.648--0.703) & $33/33^{***}$ & 13,692 & 0.22 \\
Gradient-Boosted Survival & 0.671 (0.655--0.688) & $33/33^{***}$ & 15,777 & 0.09 \\
XGBoost Classifier & 0.658 (0.500--0.679) & $33/33^{***}$ & 18,790 & 0.29 \\
Cox-Boosted Trees & 0.653 (0.500--0.683) & $33/33^{***}$ & 19,943 & --- \\
Logistic Regression & 0.643 (0.602--0.689) & $33/33^{***}$ & 23,029 & 0.37 \\
Decision Tree & 0.625 (0.594--0.641) & $33/33^{***}$ & 30,332 & 0.23 \\
DeepSurv Neural Network & 0.606 (0.556--0.642) & $33/33^{***}$ & 42,300 & 0.18 \\
\botrule
\end{tabular*}%
}
\tabnote{Note: Bold marks the best value in each column. IQR, interquartile range across the $33$ single-hospital mean external $\text{AUC}$ values (median outside parentheses); CE-Agent Wins, single-hospital comparisons won by CE-Agent against each row's baseline ($264/264$ across the eight baselines; $^{***}P < 0.001$, two-sided paired Wilcoxon signed-rank test); $N_{\min}$, minimum required sample size under Riley et al.~\cite{riley2020calculating}; Excess NB ($10^{-3}$), integrated excess net benefit ($\times 10^{-3}$) over $\max(\text{treat-all}, 0)$ across $330$ institution-threshold cells. Point-anchored CE-Agent ($p = 4$ free slope parameters out of $12$ active slopes; the thirteenth weight, on the upstream inflammation score, is fixed in every specification) requires $N_{\min} = 1{,}110$, met by $12/33$ hospitals on their own data; an unconstrained $12\text{-parameter}$ clinical/CT model requires $6{,}043$, and the baselines, with $23\text{--}34$ encoded predictor columns, a median of $13{,}275\text{--}42{,}300$ ($0/33$ hospitals each). Net benefit is positive in $289/330$ cells for CE-Agent ($33/33$ hospitals across $5\%\text{--}10\%$) and in $123\text{--}213/330$ cells for baselines. Random survival forest and Cox-boosted trees give relative hazard ranks without $10\text{-yr}$ absolute risks (---).}
\end{table}

\begin{table}[!htbp]
\centering
\caption{\textbf{MESA models trained on one racial/ethnic or scanner group and tested on the others: discrimination, minimum sample size, and net benefit ($n = 6{,}795$).} Training groups are the six racial/ethnic and CT scanner subgroups with $\ge 20$ training events ($34\text{--}81$ per fold); finite-sample scaling is in Fig.~\ref{fig:fig2}e and Table~\ref{tab:app_mesa_ood_licensing}.}\label{tab:mesa_ood}
{\tabfont
\setlength{\tabcolsep}{0pt}
\renewcommand{\arraystretch}{1.12}
\begin{tabular*}{\textwidth}{@{\extracolsep{\fill}}lcccc@{}}
\toprule
\colh{Model Architecture} & \colh{$\text{AUC}_{10}$ (IQR)} & \colh{CE-Agent Wins} & \colh{$N_{\min}$} & \colh{Excess NB$^\dagger$ ($10^{-3}$)} \\
\midrule
CE-Agent (hard bounds) & \textbf{0.801} (0.787--0.807) & \textbf{43/48} & \textbf{2,237} & \textbf{2.36} \\
Random Survival Forest & 0.795 (0.784--0.804) & 4/6 & Undef. & 2.24 \\
Cox Proportional Hazards & 0.796 (0.789--0.799) & 4/6 & 8,380 & 2.25 \\
Gradient-Boosted Survival & 0.776 (0.756--0.784) & 6/6 & Undef. & 1.43 \\
XGBoost Classifier & 0.743 (0.734--0.761) & 6/6 & Undef. & 2.17 \\
Cox-Boosted Trees & 0.794 (0.784--0.797) & 5/6 & Undef. & 2.04 \\
Logistic Regression & 0.762 (0.719--0.779) & 6/6 & 8,380 & 2.30 \\
Decision Tree & 0.677 (0.646--0.690) & 6/6 & Undef. & 1.33 \\
DeepSurv Neural Network & 0.756 (0.716--0.783) & 6/6 & Undef. & 1.81 \\
\botrule
\end{tabular*}%
}
\tabnote{Note: Bold marks the best value in each column. IQR, interquartile range (median OOD $\text{AUC}_{10}$ across the six folds outside parentheses); CE-Agent Wins, subgroup OOD folds won by CE-Agent against each row's baseline ($43/48$, $89.6\%$, across the eight baselines: $24/24$ scanner shifts, $P < 0.001$; $19/24$ demographic shifts, $P < 0.001$; two-sided paired Wilcoxon signed-rank tests); $N_{\min}$, minimum required sample size on cardiovascular mortality ($p_{\text{eff}} = 8.0$ vs.\ $22.0$ unconstrained; hard ASCVD $N_{\min} = 1{,}202$ vs.\ $3{,}055$, met by $3/4$ racial/ethnic and $2/4$ scanner subcohorts); Undef., tree and neural models without a fixed number of parameters. $^\dagger$Integrated excess net benefit ($\times 10^{-3}$) on hard ASCVD across decision thresholds of $1\%$ to $15\%$ (Fig.~\ref{fig:fig5}d). Mortality excess net benefit and all $31$ baseline configurations ($153/186$ mortality and $193/217$ hard ASCVD wins) are in Table~\ref{tab:app_mesa_ood_licensing}.}
\end{table}

\begin{table}[!htbp]
\centering
\caption{\textbf{NLST models pooled across $1$ to $32$ training hospitals: external discrimination, calibration, minimum sample size, and net benefit ($n = 37{,}521$).}}\label{tab:app_ood_licensing}
{\tabfont
\setlength{\tabcolsep}{0pt}
\renewcommand{\arraystretch}{1.14}
\begin{tabular*}{\textwidth}{@{\extracolsep{\fill}}lccccccc@{}}
\toprule
\colh{NLST Regime} & \colh{CE-Agent} & \colh{Unboxed$^*$/Unc.$^\dagger$} & \colh{Cox PH} & \colh{RSF} & \colh{GBSA} & \colh{XGB} & \colh{DSurv} \\
\midrule
\multicolumn{8}{@{}l}{\textit{External OOD discrimination (mean $\text{AUC}$ across $33 - K$ centers)}} \\
$K = 1$ Single Hospital & \textbf{0.7246} & 0.6977 & 0.6676 & 0.6761 & 0.6620 & 0.6178 & 0.6007 \\
$K = 2$ Pooled Hospitals & \textbf{0.7246} & 0.7134 & --- & --- & --- & --- & --- \\
$K = 4$ Pooled Hospitals & \textbf{0.7266} & 0.7228 & --- & --- & --- & --- & --- \\
$K = 8$ Pooled Hospitals & \textbf{0.7258} & 0.7244 & --- & --- & --- & --- & --- \\
$K = 16$ Pooled Hospitals & \textbf{0.7251} & 0.7245 & --- & --- & --- & --- & --- \\
$K = 32$ LOCO ($32 \to 1$) & 0.7256 & 0.7251 & 0.7256 & 0.7260 & --- & \textbf{0.7280} & 0.7271 \\
\midrule
\multicolumn{8}{@{}l}{\textit{Calibration slope error ($|\text{median slope} - 1|$), sample-size criteria, and DCA}} \\
$K = 1$ $|\text{Med.\ slope} - 1|$ & \textbf{0.178} & 0.512$^\dagger$ & --- & --- & --- & 0.400 & --- \\
$K = 2$ $|\text{Med.\ slope} - 1|$ & \textbf{0.173} & 0.183$^\dagger$ & --- & --- & --- & 0.468 & --- \\
$K = 4$ $|\text{Med.\ slope} - 1|$ & 0.160 & \textbf{0.072}$^\dagger$ & --- & --- & --- & 0.388 & --- \\
$K = 8$ $|\text{Med.\ slope} - 1|$ & 0.166 & \textbf{0.035}$^\dagger$ & --- & --- & --- & 0.260 & --- \\
$K = 16$ $|\text{Med.\ slope} - 1|$ & 0.166 & \textbf{0.021}$^\dagger$ & --- & --- & --- & 0.163 & --- \\
$K = 32$ $|\text{Med.\ slope} - 1|$ & 0.153 & \textbf{0.017}$^\dagger$ & --- & --- & --- & 0.068 & --- \\
Riley Free Params / Cols.\ $p$ & \textbf{4.0} & 12.0$^\dagger$ & 23--34 & 23--34 & 23--34 & 23--34 & 23--34 \\
Riley Required $N_{\min}$ & \textbf{1,110} & 6,043$^\dagger$ & 13,692 & 13,275 & 15,777 & 18,790 & 42,300 \\
DCA Int.\ Excess NB ($\times 10^{-3}$) & \textbf{0.83} & 0.50$^\dagger$ & 0.22 & --- & 0.09 & 0.29 & 0.18 \\
DCA Mean Excess NB ($\times 10^{-3}$) & \textbf{4.36} & 2.42$^\dagger$ & 1.03 & --- & 0.97 & 4.16 & 2.17 \\
DCA Positive Cells ($/ 330$) & \textbf{289/330} & 196/330$^\dagger$ & 213/330 & --- & 125/330 & 132/330 & 131/330 \\
\botrule
\end{tabular*}%
}
\tabnote{Note: Bold marks the best value in each row. PH, proportional hazards; RSF, random survival forest; GBSA, gradient-boosted survival analysis; XGB, XGBoost; DSurv, DeepSurv neural network; LOCO, leave-one-center-out; DCA, decision curve analysis; NB, net benefit; Mean Excess NB, median across hospitals of average excess net benefit over positive-benefit thresholds. $^*$Unboxed (four bounds freed; Fig.~\ref{fig:fig3}a) leaves the four interval-bounded predictors unconstrained while keeping published point anchors and sign constraints on the remaining predictors ($0.6977$ mean external $\text{AUC}$ at $K = 1$, vs.\ $0.6559$ for fully unconstrained $12\text{-parameter}$ logistic estimation); $^\dagger$fully unconstrained $12\text{-parameter}$ model (Fig.~\ref{fig:fig5}a--c). Survival and ML baselines report median $N_{\min}$ across the $33$ hospitals using $23\text{--}34$ encoded predictor columns (Table~\ref{tab:nlst_ood}). Slope rows hold the single-hospital ($K = 1$) point-anchored specification fixed across $K$; the sample-size-adaptive $1/n$ relaxation converges to unit slope ($1.00$) as $K$ grows (Table~\ref{tab:app_lambda_rivals}). At $K = 1$, $28/33$ centers ($84.8\%$; $29/33$ at $K = 2$) have calibration slopes within $[0.80, 1.25]$, vs.\ $8/33$ ($24.2\%$) unconstrained, and $12/33$ ($36.4\%$) meet $N_{\min}$ on their own data, vs.\ $0/33$ unconstrained. XGB-Cox (not tabulated) reaches $0.6211$ at $K = 1$ and $0.7298$ at $K = 32$.}
\end{table}

\begin{table}[!htbp]
\centering
\caption{\textbf{MESA models by training size ($n = 250$ to $4{,}821$) and by racial/ethnic and scanner group: CE-Agent versus $31$ baseline configurations ($n = 6{,}795$).}}\label{tab:app_mesa_ood_licensing}
{\tabfont
\setlength{\tabcolsep}{0pt}
\renewcommand{\arraystretch}{1.16}
\begin{tabular*}{\textwidth}{@{\extracolsep{\fill}}lccccccc@{}}
\toprule
\colh{MESA Regime} & \colh{CE-Agent} & \colh{Unc.} & \colh{Cox PH} & \colh{RSF} & \colh{GBSA} & \colh{XGB-Cox} & \colh{XGB} \\
\midrule
\multicolumn{8}{@{}l}{\textit{Finite-sample mortality discrimination (mean $\text{AUC}_{10}$, $20$ draws)}} \\
$n = 250$ ($6$ deaths) & 0.7173 & 0.5549 & \textbf{0.7526} & 0.7487 & 0.7106 & 0.7055 & 0.5000 \\
$n = 500$ ($13$ deaths) & 0.7740 & 0.6789 & \textbf{0.7800} & 0.7667 & 0.7319 & 0.7445 & 0.7063 \\
$n = 1{,}000$ ($26$ deaths) & \textbf{0.7950} & 0.7429 & 0.7886 & 0.7672 & 0.7433 & 0.7605 & 0.7299 \\
$n = 2{,}000$ ($52$ deaths) & \textbf{0.8000} & 0.7659 & 0.7951 & 0.7674 & 0.7465 & 0.7663 & 0.7250 \\
Full Train ($n = 4{,}821$) & \textbf{0.8068} & 0.7886 & 0.7972 & 0.7832 & --- & 0.7793 & 0.7658 \\
\midrule
\multicolumn{8}{@{}l}{\textit{Cross-group OOD $\text{AUC}_{10}$, slope error ($|\text{slope}_{10} - 1|$), sample-size criteria, and DCA}} \\
Hispanic OOD ($E = 36$) & 0.7776 & 0.7184 & 0.7835 & 0.7717 & 0.7535 & \textbf{0.7919} & 0.7326 \\
Scanner OOD ($E = 34$) & \textbf{0.7848} & 0.7238 & 0.7079 & 0.7823 & 0.7352 & 0.7769 & 0.7381 \\
LOGO Mean (CVD Death) & 0.8051 & 0.8046 & 0.8054 & 0.7965 & 0.7946 & \textbf{0.8215} & 0.7827 \\
LOGO Mean (Hard ASCVD) & 0.7765 & \textbf{0.7846} & 0.7773 & 0.7734 & 0.7619 & 0.7773 & 0.7672 \\
$|\text{slope}_{10} - 1|$ ($n = 250$) & 0.720 & 0.973 & \textbf{0.160} & --- & --- & --- & --- \\
$|\text{slope}_{10} - 1|$ ($n = 500$) & 0.418 & 0.775 & \textbf{0.125} & --- & --- & --- & 0.378 \\
$|\text{slope}_{10} - 1|$ ($n = 1{,}000$) & 0.344 & 0.516 & \textbf{0.272} & --- & --- & --- & 0.448 \\
$|\text{slope}_{10} - 1|$ (Hispanic) & 0.330 & 0.513 & 0.219 & \textbf{0.051} & 0.727 & 0.391 & 0.459 \\
$|\text{slope}_{10} - 1|$ (Scanner) & 0.400 & 0.677 & 0.620 & 0.397 & \textbf{0.090} & 0.497 & 0.546 \\
Riley $N_{\min}$ (CVD Death) & \textbf{2,237} & 8,380 & 8,380 & Undef. & Undef. & Undef. & Undef. \\
Riley $N_{\min}$ (Hard ASCVD) & \textbf{1,202} & 3,055 & 3,055 & Undef. & Undef. & Undef. & Undef. \\
Excess NB ($\times 10^{-4}$, CVD) & 4.67 & 3.68 & \textbf{5.37} & 4.43 & 2.67 & 3.81 & 2.62 \\
Excess NB ($\times 10^{-3}$, ASCVD) & \textbf{2.36} & 2.35 & 2.25 & 2.24 & 1.43 & 2.04 & 2.17 \\
\botrule
\end{tabular*}%
}
\tabnote{Note: Bold marks the best value in each row. Unc., unconstrained $22\text{-predictor}$ model; GBSA, gradient-boosted survival analysis; XGB-Cox, Cox-boosted trees; XGB, XGBoost; Undef., no fixed number of parameters; $E$, training events (approximate mean training deaths in parentheses for $n = 250 \dots 2{,}000$); LOGO, leave-one-group-out ($4$ racial/ethnic groups and $4$ scanner cohorts); Excess NB, decision-curve integrated excess net benefit. With $50$ rather than $20$ subsampling draws, CE-Agent mean $\text{AUC}_{10}$ is $0.7171, 0.7794, 0.7949, 0.8009$ at $n = 250 \dots 2{,}000$ (DeepSurv $20\text{-draw}$ mean $0.6888, 0.7319, 0.7460, 0.7658$). Across $31$ baseline configurations, CE-Agent won $58.5\%\text{--}84.2\%$ of pairwise subsample comparisons on CVD mortality ($363/620$ at $n = 250$, $464/620$ at $n = 500$, $510/620$ at $n = 1{,}000$, and $522/620$ at $n = 2{,}000$), $74.7\%\text{--}89.7\%$ on hard ASCVD ($463/620$, $556/620$, and $550/620$ at $n = 250$, $500$, and $1{,}000$), and $66.9\%\text{--}91.4\%$ of cross-group OOD and LOGO comparisons. With $p_{\text{eff}} = 8.0$ (CVD) and $9.0$ (ASCVD) vs.\ $22$ unconstrained, CE-Agent meets $N_{\min}$ in $3/4$ racial/ethnic and $2/4$ scanner cohorts on hard ASCVD. Excess net benefit: CVD death, $4.67 \times 10^{-4}$ vs.\ $0.53 \times 10^{-4}$ for ASCVD-PCE ($P < 0.001$); hard ASCVD, $2.36 \times 10^{-3}$ vs.\ $1.95 \times 10^{-3}$ for PREVENT ($P = 0.038$) and $1.79 \times 10^{-3}$ for ASCVD-PCE ($P = 0.068$).}
\end{table}

\FloatBarrier
\subsection{Intermediate biomarkers and missing CT in MESA}

\begin{figure}[!htbp]
\centering
\includegraphics[width=\textwidth]{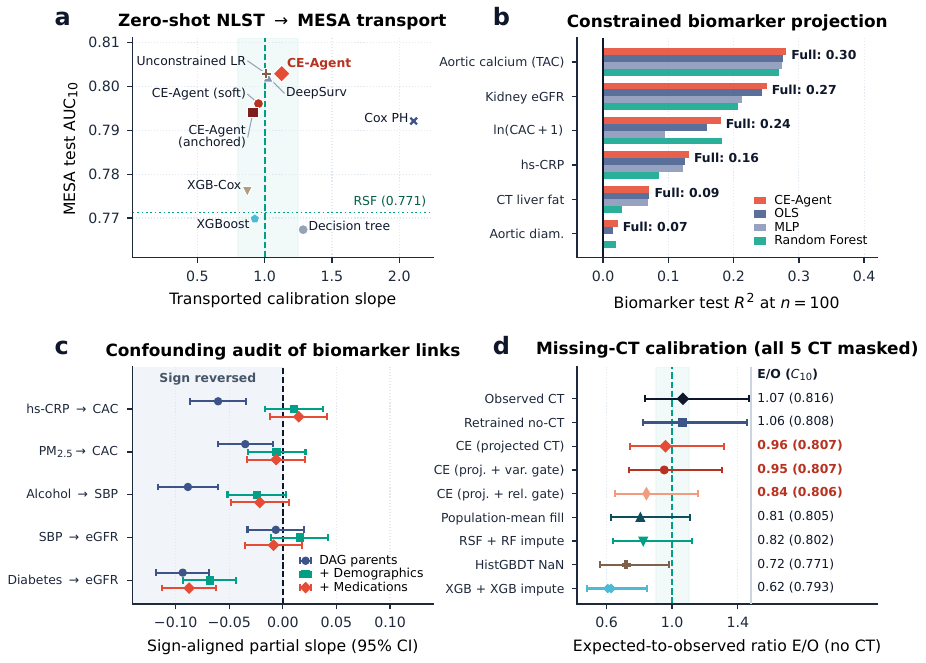}
\caption{\textbf{Transport from NLST to MESA and projection of intermediate biomarkers in MESA.}
\textbf{a}, NLST-trained weights applied to the held-out MESA test cohort on $12$ shared predictors with prevalence recalibration only: $\text{AUC}_{10}$ and calibration slope (band, $0.80$ to $1.25$) for CE-Agent with hard, soft, or point-anchored (anchored) constraints and for seven survival and ML baselines.
\textbf{b}, Test coefficient of determination ($R^2$) for predicting MESA's measured intermediate biomarkers from their parent variables with $n = 100$ training participants (bars) and with the full training cohort ($n = 4{,}821$; Full), for direction-constrained fitting (CE-Agent), ordinary least squares (OLS), multilayer perceptrons (MLP), and random forests. TAC, thoracic aortic calcification; eGFR, estimated glomerular filtration rate; CAC, coronary artery calcification; hs-CRP, high-sensitivity C-reactive protein.
\textbf{c}, Partial slopes aligned to the expected sign (negative, reversed; $95\%$ confidence intervals) of parent-to-biomarker links whose observed direction was reversed, adjusted for DAG parents only and after adding demographics and then medications ($\text{PM}_{2.5}$, fine particulate matter; SBP, systolic blood pressure).
\textbf{d}, Expected-to-observed ratio ($\text{E/O}$; band, $0.90$ to $1.10$) and 10-year concordance ($C_{10}$) with all five chest CT phenotypes masked. CE-Agent (CE) with CT projected from clinical predictors (proj.), alone or down-weighted by projection variance (var.\ gate) or by the reliability score (rel.\ gate), is compared with observed CT, a model retrained without CT, population-mean fill, histogram-based gradient-boosted decision trees that route missing values natively (HistGBDT NaN), and ML imputers (RF, random forest).}\label{fig:fig7}
\end{figure}

Figure~\ref{fig:fig7} and Tables~\ref{tab:app_mesa_mediators}--\ref{tab:app_mesa_missing_ct} evaluate zero-shot cross-cohort transport from NLST to MESA (Table~\ref{tab:app_zeroshot_transport}), sign-constrained estimation across MESA's $10$ directly measured intermediate clinical, laboratory, and CT biomarkers (mediators) compared against unconstrained regression and seven regularized, tree-ensemble, and neural baselines, and outpatient deployment when all five thoracic CT biomarkers are unmeasured.

\begin{table}[!htbp]
\centering
\caption{\textbf{Direction-constrained prediction of $10$ intermediate biomarkers from their parent variables in MESA, with a confounding audit ($n = 4{,}821$ training, $1{,}974$ test).} Constrained fitting gives no parent sign reversals, versus mean reversal rates of $10\%\text{--}27\%$ across unconstrained regression and seven ML baselines at $n = 100$.}\label{tab:app_mesa_mediators}
{\tabfont
\setlength{\tabcolsep}{0pt}
\renewcommand{\arraystretch}{1.04}
\begin{tabular*}{\textwidth}{@{\extracolsep{\fill}}lcccc@{}}
\toprule
\colh{Mediator (Metric)} & \colh{Test (C / U)} & \colh{$\Delta_{100}$} & \colh{$n = 100$ Wins} & \colh{Confounding Audit} \\
\midrule
1.\ $\ln(\text{TAC}+1)$ ($R^2$) & 0.2990 / 0.2990 & $+0.0047$ & $43.5\%$ / $56\text{--}100\%$ & $2/2$ concordant \\
2.\ CKD-EPI eGFR ($R^2$) & 0.2710 / \textbf{0.2771} & $+0.0070$ & $48.0\%$ / $63\text{--}100\%$ & DM: $+0.093 \to +0.087$ \\
3.\ $\ln(\text{CAC}+1)$ ($R^2$) & 0.2429 / \textbf{0.2467} & $+0.0204$ & $75.5\%$ / $44\text{--}99\%$ & CRP: $-0.060 \to \mathbf{+0.015}$ \\
3b.\ $\text{CAC} > 0$ ($\text{AUC}$) & 0.7663 / \textbf{0.7677} & $+0.0127$ & $77.5\%$ / $52\text{--}100\%$ & $\text{PM}_{2.5}$: $-0.035 \to \mathbf{-0.006}$ \\
4.\ $\ln(\text{CRP}+1)$ ($R^2$) & 0.1591 / 0.1591 & $+0.0065$ & $27.5\%$ ($100\%$ $\ge \text{U}$) & $2/2$ concordant \\
5.\ CT Liver Fat ($R^2$) & 0.0883 / 0.0883 & $+0.0000$ & $100\%$ concordant & $1/1$ concordant \\
6.\ Ao.\ Diameter ($R^2$) & 0.0680 / 0.0680 & $+0.0070$ & $27.5\%$ ($100\%$ $\ge \text{U}$) & $2/2$ concordant \\
7.\ Chamber Area ($R^2$) & 0.0296 / 0.0296 & $+0.0008$ & $4.0\%$ ($100\%$ $\ge \text{U}$) & $1/1$ concordant \\
8.\ Systolic BP ($R^2$) & 0.0171 / \textbf{0.0250} & $+0.0009$ & $23.5\%$ / $38\text{--}100\%$ & Alc.: $-0.088 \to \mathbf{-0.021}$ \\
9.\ Diabetes ($\text{AUC}$) & 0.6331 / 0.6331 & $+0.0067$ & $5.0\%$ ($100\%$ $\ge \text{U}$) & $1/1$ concordant \\
10.\ Respiratory Disease ($\text{AUC}$) & 0.5543 / 0.5543 & $+0.0193$ & $35.5\%$ ($100\%$ $\ge \text{U}$) & $1/1$ concordant \\
\botrule
\end{tabular*}%
}
\tabnote{Note: Bold in Test (C / U) marks the higher held-out score when C and U differ to four decimal places; bold adjusted slopes mark $>75\%$ attenuation of confounding or a sign reversal toward the causal direction. $R^2$, coefficient of determination; C, sign-constrained; U, unconstrained; $\Delta_{100}$, mean test gain ($\text{C} - \text{U}$) across $200$ subsamples of size $n = 100$ (non-negative across all $10$ mediators); $n = 100$ Wins, strict win rate vs.\ U / range of win rates across seven ML baselines (ridge, lasso, elastic net, random forest, gradient-boosted trees, XGBoost, and multilayer perceptron; $100\%$ $\ge \text{U}$, C matches or exceeds U on all $200$ draws); Confounding Audit, standardized parent slope on reversed U edges adjusted for DAG parents only $\to$ after adding demographics and medications (raw slopes; expected sign negative for DM- and SBP-to-eGFR, positive otherwise; SBP-to-eGFR $+0.007 \to +0.009$, Fig.~\ref{fig:fig7}c; $\text{PM}_{2.5}$ slope on continuous $\ln(\text{CAC}+1)$; $76\%$ of alcohol-to-SBP and $82\%$ of $\text{PM}_{2.5}$-to-CAC confounding attenuated); CKD-EPI, Chronic Kidney Disease Epidemiology Collaboration; eGFR, estimated glomerular filtration rate; CRP, high-sensitivity C-reactive protein; BP, blood pressure; $\text{PM}_{2.5}$, fine particulate matter ($\le 2.5\;\mu\text{m}$); DM, diabetes mellitus; Alc., alcohol; Ao., ascending aortic; Respiratory Disease, self-reported emphysema or asthma. Observed parents in the domain DAG (Methods; $22$ links, with binary $\text{CAC} > 0$ reusing the seven CAC parents): TAC (age, lipids); eGFR (age, DM, SBP); CAC (age, smoking, lipids, DM, liver fat, CRP, $\text{PM}_{2.5}$); CRP (smoking, BMI); liver fat (BMI); Ao.\ diameter (age, SBP); chamber area (SBP); SBP (alcohol, BMI); diabetes (BMI); respiratory disease (smoking).}
\end{table}

\begin{table}[!t]
\centering
\caption{\textbf{MESA cardiovascular mortality when all five CT phenotypes are unavailable at prediction ($n = 1{,}974$).} Row 1 provides the observed-CT reference; bold values denote best CE-Agent or retrained single-model missing-CT performance.}\label{tab:app_mesa_missing_ct}
{\tabfont
\setlength{\tabcolsep}{0pt}
\renewcommand{\arraystretch}{1.18}
\begin{tabular*}{\textwidth}{@{\extracolsep{\fill}}lccccc@{}}
\toprule
\colh{Inference Regime} & \colh{Full $C$} & \colh{$\text{AUC}_{10}$} & \colh{Brier} & \colh{E/O (95\% CI)} & \colh{Slope} \\
\midrule
1.\ Observed CT & 0.8046 & 0.8068 & 0.02390 & 1.07 (0.84--1.47) & 0.777 \\
\midrule
2.\ Population-Mean Fill ($Z_{\text{CT}} = 0$) & 0.7872 & 0.7954 & 0.02363 & 0.81 (0.63--1.11) & 0.907 \\
3.\ Projected CT (rel.-gated) & 0.7884 & 0.7961 & 0.02360 & 0.84 (0.65--1.16) & 0.884 \\
4.\ Projected CT (var.-gated) & 0.7903 & 0.7970 & 0.02359 & 0.95 (0.74--1.31) & 0.826 \\
5.\ Projected CT (unattenuated)$^*$ & \textbf{0.7904} & 0.7971 & 0.02360 & \textbf{0.96} (0.74--1.32) & 0.823 \\
6.\ Retrained No-CT ($18$ features) & 0.7903 & \textbf{0.7983} & \textbf{0.02356} & 1.06 (0.82--1.45) & \textbf{0.926} \\
7.\ External ML ($84$ configs) & 0.673--0.800 & 0.676--0.806 & 0.0235--0.0258 & 0.59--1.19 & 0.43--3.09$^\dagger$ \\
\botrule
\end{tabular*}%
}
\tabnote{Note: $^*$Unattenuated conditional projection uses $\mu_{im} = \hat{b}_m + \sum_p \hat{w}_{pm} \mu_{ip}$ in equation~(\ref{eq:projection}) with observed-entry gate weight $\operatorname{expit}(\alpha_0)$; variance-gated (var.-gated) scales $\mu_{im}$ by $\operatorname{expit}(\alpha_0 - \alpha_v v_{im}) / \operatorname{expit}(\alpha_0)$, and reliability-gated (rel.-gated) scales $\mu_{im}$ by $\operatorname{expit}(\alpha_0 - \alpha_m - \alpha_v v_{im}) / \operatorname{expit}(\alpha_0)$. Row 6 omits the $4$ direct CT columns from the $22\text{-predictor}$ outcome model (the fifth masked CT phenotype, liver attenuation, enters as an upstream parent of CAC). Relative to Population-Mean Fill (row 2), unattenuated projection (row 5) gives paired bootstrap changes of $\Delta C = +0.0031$ and $\Delta\text{E/O} = +0.153$ ($95\%$ CI $+0.118\text{ to }+0.209$) and moves calibration-in-the-large (CITL) from $+0.233$ to $+0.046$ ($\Delta\text{CITL} = -0.187$, $95\%$ CI $-0.200\text{ to }-0.174$); $10\text{-year}$ $C_{10}$ is $0.8052$ and $0.8074$ for rows 2 and 5, $0.8157$ with observed CT, and $0.8083$ for the retrained no-CT model. $^\dagger$Row 7 spans $12$ external ML architectures across $6$ imputers plus retrained no-CT ($84$ configurations; $C_{10}$ range $0.686\text{--}0.816$; Brier $0.02348\text{--}0.02579$). Unattenuated CE-Agent has lower $|\text{E/O} - 1|$ and higher $C$ than $68/84$ ($81.0\%$) configurations; reliability-gated and retrained no-CT CE-Agent have lower $|\text{slope} - 1|$ than $60/84$ and $71/84$ configurations, respectively; reliability gating gives the lowest expected calibration error ($\text{ECE} = 0.0081$). Histogram-based gradient-boosted decision trees with native missing-value handling reach $C = 0.7747$ ($C_{10} = 0.7706$), $\text{E/O} = 0.72$, and slope $1.149$.}
\end{table}

\FloatBarrier
\subsection{Ablations of the Causal Envelope}

\begin{figure}[!htbp]
\centering
\includegraphics[width=\textwidth]{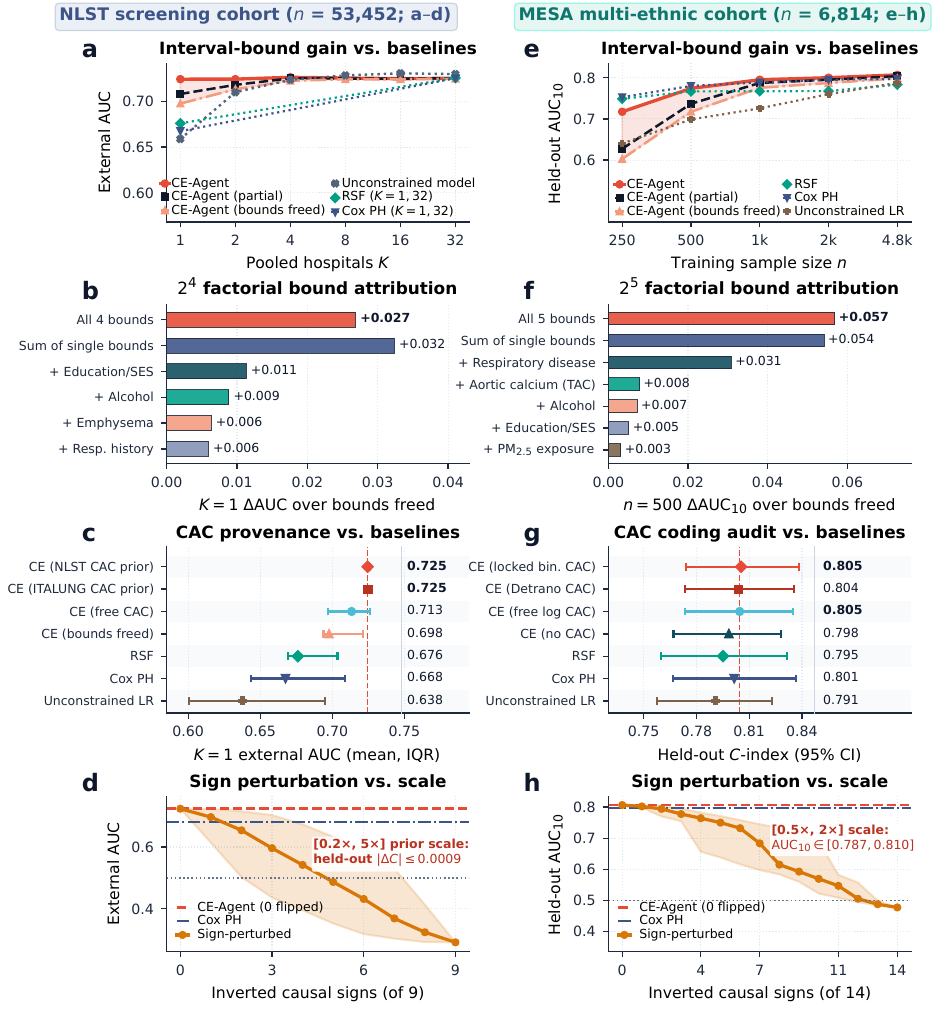}
\caption{\textbf{Contribution of published ranges, the CAC prior source, and predictor directions in NLST (a--d) and MESA (e--h).}
\textbf{a},\textbf{e}, External discrimination across sample size for CE-Agent with all, some (partial), or none (bounds freed) of the published ranges enforced, across $1\text{ to }32$ pooled NLST hospitals (\textbf{a}) and MESA subsamples of $250\text{ to }4{,}821$ participants (\textbf{e}); shading, all versus none. Comparators are RSF, Cox PH, and an unconstrained fit (the CE-Agent model without priors in \textbf{a}; unconstrained logistic regression in \textbf{e}).
\textbf{b},\textbf{f}, Gain in discrimination from all published ranges together, from the sum of single-predictor ranges, and from each range alone, in single-hospital NLST fits (\textbf{b}) and $500\text{-participant}$ MESA subsamples (\textbf{f}; $\text{PM}_{2.5}$, fine particulate matter; Resp., respiratory).
\textbf{c},\textbf{g}, CE-Agent under alternative sources for the CAC weight (the NLST-derived~\cite{chiles2015association} and ITALUNG~\cite{mascalchi2021moderate} estimates, a free CAC weight, and all four interval bounds freed in \textbf{c}) and alternative CAC codings (including the Detrano et al.~\cite{detrano2008coronary} log-CAC bound in \textbf{g}), compared with RSF, Cox PH, and unconstrained logistic regression.
\textbf{d},\textbf{h}, Discrimination when the directions of subsets of predictors are inverted (median; band, range in \textbf{d} and interquartile range in \textbf{h}), against intact CE-Agent and Cox PH; rescaling all priors changed discrimination little ($|\Delta C\text{-index}| \le 0.0009$ across a $25\times$ span in NLST, \textbf{d}; $\text{AUC}_{10}$ $0.787\text{--}0.810$ across $0.5\times\text{ to }2\times$ in MESA, \textbf{h}). CE, CE-Agent.}\label{fig:fig3}
\end{figure}

Tables~\ref{tab:app_prior_channel_mechanics}--\ref{tab:app_mesa_mechanics} and Fig.~\ref{fig:fig3} report mechanistic ablations across NLST and MESA, complementing Fig.~\ref{fig:fig4}. Each ablation removes or alters one part of the envelope: the published ranges, the CAC source~\cite{chiles2015association,detrano2008coronary}, the predictor directions, composite-index screening (Proposition~\ref{prop:mediator}), added CT phenotypes, or the zeroing of reversed weights (Proposition~\ref{prop:kkt}).

\begin{table}[!htbp]
\centering
\caption{\textbf{Mechanistic ablations of the Causal Envelope in NLST ($n = 37{,}521$ training; rows 5--6, $n = 15{,}931$ held-out test).}}\label{tab:app_prior_channel_mechanics}
{\fontsize{6.8bp}{8.2bp}\selectfont
\setlength{\tabcolsep}{3.1pt}
\renewcommand{\arraystretch}{1.04}
\begin{tabularx}{\textwidth}{@{}L{0.175\textwidth}YYYY@{}}
\toprule
\colh{Experiment} & \colh{Protocol} & \colh{CE-Agent} & \colh{ML Baseline} & \colh{Effect Size} \\
\midrule
1.\ Interval-Bound Sample Scaling & $K \in \{1, \allowbreak 2, \allowbreak 4, \allowbreak 8, \allowbreak 16, \allowbreak 32\}$ hospitals ($4$ active bounds) & $K=1$: $\mathbf{0.7246} \to$ $K=32$: $\mathbf{0.7256}$ & Four bounds freed: $0.6977 \to \allowbreak 0.7251$; Cox: $0.6676 \to \allowbreak 0.7256$ & $+0.0268$ ($K=1$, $P<0.001$) $\to$ $+0.0005$ ($K=32$) \\
2.\ Factorial Bound Attribution ($2^4$) & All $16$ bound coalitions across $33$ centers ($K=1$) & Joint $4\text{-bound}$ $\Delta\text{AUC} = \mathbf{+0.0268}$ & Sum of singletons $\sum\Delta_j = +0.0323$; Four bounds freed $0.6977$ & $83.0\%$ main-effect share; interaction $\max|I_S| = 0.0036$ \\
3.\ CAC Provenance Audit & Replace the NLST CAC estimate~\cite{chiles2015association} with the ITALUNG trial estimate~\cite{mascalchi2021moderate} ($K=1$) & NLST estimate $\mathbf{0.7246}$; ITALUNG estimate $\mathbf{0.7248}$ & Free CAC $0.7134$; mean RSF $0.6761$; Cox $0.6676$; LR $0.6376$ & $\Delta\text{AUC} = +0.0002$; free-CAC model above $8/8$ baselines \\
4.\ Sign-Inversion Perturbation & All $2^9 = 512$ sign coalitions ($0 \to 9$ inverted) & $0$ flipped: med.\ $\text{AUC} = \mathbf{0.7246}$ & $3$ flipped: $0.596$; $6$ flipped: $0.431$; $9$ flipped: $0.290$ & $\Delta\text{AUC} = -0.434$ ($0 \to 9$ inverted) \\
5.\ Positive Scale Resilience & Global prior multiplier $s \in [0.2\times, 5.0\times]$ ($25\times$ span) & Held-out $C \in [\mathbf{0.7297}, \mathbf{0.7306}]$ & Sign inversion $\Delta\text{AUC} = -0.434$ ($480\times$ larger) & $|\Delta C| \le 0.0009$ across $25\times$ span \\
6.\ Algebraic Equivalence & Include derived vascular-aging composite ($n=15{,}931$) & Screened: $C = \mathbf{0.7306}$ & Included: $r = 0.9999999999$, $\max|\Delta\eta| = 1.78\times 10^{-15}$ & $\Delta\text{rank}=0$ (Proposition~\ref{prop:mediator}) \\
7.\ Composite Relaxation & Screened vs.\ progressive relaxation at $K=1$ ($33$ centers) & Screened, published weights: $\mathbf{0.7251}$ ($\kappa\text{-ratio } 1.0\times$) & Relax BMI: $0.7190$; Shared-parent: $0.6985$; All $4$ free: $0.6947$ ($\kappa\text{-ratio } 321\times$) & $\Delta\text{AUC} = -0.0304$ ($30/33$ centers) \\
8.\ 3D LDCT Expansion & Training LDCT subcohorts ($n = 5{,}269$; $6{,}467$; $1{,}243$), 5-fold CV and cross-site: $+3$ LDCT & $\text{AUC}$: $\mathbf{0.748\text{--}0.763} \to$ $\mathbf{0.763\text{--}0.775}$ (soft-tissue 5-fold CV: $0.759 \to 0.774$) & Clinical-only CE-Agent ($13$ predictors): $0.748\text{--}0.763$; PCE held-out $C\text{-index}$: $0.6926$ & $\Delta\text{AUC} = +0.011$ to $+0.017$ ($6/6$ comparisons) \\
9.\ Zeroing of Reversed Weights & COPD-confounded PA diameter ratio ($1{,}040$ CV fits) & $\hat{c}_{\text{PA}} = \mathbf{0.000}$ in $\mathbf{928/1{,}040}$ ($\mathbf{89.2\%}$) & Direction constraint lifted: $\hat{c}_{\text{unc}} \in [-0.25, -0.12]$ & $89.2\%$ ($928/1{,}040$) zeroed$^*$ \\
\botrule
\end{tabularx}%
}
\tabnote{Note: Bold values highlight primary CE-Agent metrics. Abbreviations: $\Delta_j$, marginal single-bound $\text{AUC}$ gain; $I_S$, interaction for bound set $S$ from the inclusion--exclusion (Möbius) decomposition of all $2^4$ bound combinations; $r$, Pearson correlation; $\Delta\eta$, log-odds difference; $\kappa\text{-ratio}$, design-matrix condition-number ratio relative to screened matrix; $\hat{c}_{\text{PA}}$ and $\hat{c}_{\text{unc}}$, standardized pulmonary artery (PA) diameter ratio weights with and without the direction constraint (other predictors under hard bounds); CV, cross-validation. $^*$Across the $1{,}040$ CV fits, proximal main PA-to-sinotubular junction ratio is zeroed in $668/780$ fits and direct ascending-aorta-normalized ratio sensitivity fits ($n = 1{,}243$) are zeroed in $260/260$ ($100\%$).}
\end{table}

\begin{table}[!htbp]
\centering
\caption{\textbf{Ablations of the Causal Envelope in MESA ($n = 6{,}795$).}}\label{tab:app_mesa_mechanics}
{\tabfont
\setlength{\tabcolsep}{3.0pt}
\renewcommand{\arraystretch}{1.07}
\begin{tabularx}{\textwidth}{@{}L{0.175\textwidth}YYYY@{}}
\toprule
\colh{Experiment} & \colh{Protocol} & \colh{CE-Agent} & \colh{ML Baseline} & \colh{Effect Size} \\
\midrule
1.\ Interval-Bound Sample Scaling & $n \in \{250, \allowbreak 500, \allowbreak 1{,}000, \allowbreak 2{,}000, \allowbreak 4{,}821\}$ ($5$ active bounds) & $n=250$: $\mathbf{0.7171} \to$ Full: $\mathbf{0.8068}$ & Five bounds freed: $0.6025 \to \allowbreak 0.7983$; Unc.: $0.5855 \to \allowbreak 0.7886$ & $+0.1146 \to \allowbreak +0.0569 \to \allowbreak +0.0195 \to \allowbreak +0.0129 \to \allowbreak +0.0086$ \\
2.\ Factorial Bound Attribution ($2^5$) & All $32$ coalitions at $n=500$ ($50$ draws) \& full split & $n=500$: $\Delta_{\text{all}} = \mathbf{+0.0569}$; Full $\mathbf{+0.0086}$ & $n=500$: $\sum\Delta_j = +0.0542$ ($95.3\%$); Full $\sum\Delta_j = +0.0087$ ($101\%$) & $95.3\%\text{--}101\%$ additive; interaction $|I_{jk}| \le 0.0044$ \\
3.\ CAC Provenance Audit & Binary free/locked vs.\ Detrano interval~\cite{detrano2008coronary} vs.\ Drop CAC & CVD bin.\ free $\mathbf{0.8107}$ (locked $0.7983$); $\ln(\text{CAC}+1)$ $\mathbf{0.8068}$ & Unc.\ $0.7886$; RSF $0.7832$; Drop CAC: ASCVD $\Delta C = -0.0106$ ($95\%$ CI $-0.017$ to $-0.005$) & $+0.0043$ on ASCVD at $n=500$ ($40/50$ draws)$^*$ \\
4.\ Sign-Inversion Perturbation & $0 \to 14$ inverted signs (CVD death \& Hard ASCVD) & $0$ flipped: $\mathbf{0.8068}$ (CVD), $\mathbf{0.7740}$ (ASCVD) & $14$ flipped: $0.4774$ (CVD), $0.5602$ (ASCVD); Unc.\ $6/14$ flipped & $\Delta\text{AUC}_{10} = -0.329$ (CVD), $-0.214$ (ASCVD) \\
5.\ Positive Scale Resilience & Global prior multiplier $s \in [0.5\times, 2.0\times]$ ($0.2\times\text{--}5\times$) & $\text{AUC}_{10}$ $0.787\text{--}0.810$ across $0.5\times\text{--}2\times$ & Zero-locked ridge ($8$ free): $\text{AUC}_{10} = 0.7130$ ($-0.094$) & $|\Delta\text{AUC}_{10}| \le 0.020$ ($16\times$ smaller vs.\ sign flip) \\
6.\ Algebraic Equivalence & Include derived composites ($n=1{,}974$ test) & Screened: $\text{AUC}_{10} = \mathbf{0.8068}$ & Included: $\max|\Delta\eta| = 6.08\times 10^{-7}$, $\Delta\text{AUC}_{10} = 0.0000$ & $\Delta\text{AUC}_{10} = 0.0000$ (Proposition~\ref{prop:mediator}) \\
7.\ Composite Relaxation & Screened vs.\ progressive relaxation at $n=250$ \& $n=500$ ($200$ event-stratified draws) & Screened: $\mathbf{0.7150}$ ($n=250$), $\mathbf{0.7662}$ ($n=500$) & All $10$ free: $0.6299$ ($n=250$, $\kappa = 2.2\times 10^{10}$), $0.7277$ ($n=500$) & $\Delta\text{AUC}_{10} = -0.0851$ ($n=250$) \\
8.\ 5-Phenotype CT Progression & No CT $\to$ +CAC $\to$ +TAC $\to$ +Chamber $\to$ +Ao.\ diam.\ $\to$ +Liver fat & ASCVD $\mathbf{0.7576} \to \mathbf{0.7744}$; CVD $\mathbf{0.7983} \to \mathbf{0.8069}$ & Unc.\ LR: $0.7468 \to \allowbreak 0.7725$ (ASCVD), $0.7920 \to \allowbreak 0.7876$ (CVD) & ASCVD $\Delta\text{AUC}_{10} = +0.0168$ ($95\%$ CI $+0.006$ to $+0.027$) \\
9.\ Zeroing of Reversed Weights & $213$ fits ($200$ bootstrap, $5$ cross-validation, and $8$ single-group) & $\max|\nu_j\hat{c}_j| \le \mathbf{4.4\times 10^{-9}}$; $\Delta\text{AUC}_{10}=\mathbf{+0.0123}$ & Unconstrained flips $6/14$ ($42.9\%$) signs ($\text{AUC}_{10} = 0.7886$) & $63.4\%\text{--}84.5\%$ zeroed across $6$ predictors \\
\botrule
\end{tabularx}%
}
\tabnote{Note: Bold values highlight primary CE-Agent metrics. Abbreviations: Unc., unconstrained model; $\Delta_j$, marginal single-bound $\text{AUC}_{10}$ gain; $\Delta_{\text{all}}$, gain with all bounds applied; $I_{jk}$, pairwise interaction from the inclusion--exclusion (Möbius) decomposition of bound combinations; bin., binary CAC coding; $\Delta\eta$, log-odds difference; $\kappa$, design-matrix condition number; Ao.\ diam., ascending aortic diameter; $\nu_j$, KKT dual multiplier satisfying complementary slackness $\nu_j \hat{c}_j = 0$ (equation~(\ref{eq:kkt})). $^*$In Row 3, the Detrano et al.~\cite{detrano2008coronary} coronary interval bound improves small-sample Hard ASCVD discrimination at $n = 500$ ($+0.0043$, winning $40/50$ draws). In full-sample Hard ASCVD ($n = 4{,}812$, $137$ test events), Stage I compatibility screening assigns CAC a directional sign constraint (or $1/n$ soft bound) because Detrano et al.\ report a coronary-only endpoint, which differs from the coronary-plus-stroke composite (soft-bound $\text{AUC}_{10} = 0.7765$, Table~\ref{tab:app_composite_endpoints}). All cell discrimination values denote 10-year $\text{AUC}_{10}$; small-sample experiments ($n = 250, 500$) average $20\text{--}50$ independent stratified subsample draws ($200$ in row 7, with $6$ and $13$ events per draw).}
\end{table}

\FloatBarrier
\subsection{Uncertainty propagation and missing data}

Tables~\ref{tab:app_uncertainty_propagation}--\ref{tab:app_mesa_uncertainty} report missing-data benchmarks in NLST and MESA across $11$ missingness settings ($0\%\text{--}60\%$; missing completely at random [MCAR] or at random given BMI [MAR]), complementing Fig.~\ref{fig:fig6} and Table~\ref{tab:app_lung_ood_mech}.

\begin{table}[!htbp]
\centering
\caption{\textbf{Uncertainty propagation with missing data in NLST cardiovascular mortality (Stage I graph; five-fold out-of-fold, $n = 37{,}521$).}}\label{tab:app_uncertainty_propagation}
{\fontsize{7.2bp}{8.4bp}\selectfont
\setlength{\tabcolsep}{2.7pt}
\renewcommand{\arraystretch}{0.91}
\begin{tabularx}{\textwidth}{@{}L{0.18\textwidth}YYYY@{}}
\toprule
\colh{Dimension} & \colh{Metric / Setup} & \colh{CE-Agent} & \colh{Baselines} & \colh{Summary} \\
\midrule
1.\ $\hat{W}_{90}$ Interval Accuracy & Rel.\ MAE vs.\ $10^5\text{-draw}$ MC ($r$) & \textbf{2.30\%} [2.05--2.57\%] Cond.\ imp.\ ($r=0.9997$) & MC-4000: $3.06\%$ [$2.74\text{--}3.33\%$]; MC-400: $9.07\%$; single Gaussian: $8.03\%$; $v=0$: $64.3\text{--}79.3\%$ & $11/11$ cells; lung graph: $0.64\text{--}0.79\%$ \\
\addlinespace[0.6pt]
2.\ Predictive SD $\hat{\sigma}_p$ & Rel.\ MAE vs.\ $10^5\text{-draw}$ MC & \textbf{2.00\%} [1.83--2.14\%] Cond.\ imp.\ & MC-4000: $5.23\%$; MC-400: $15.06\%$; $v=0$: $71.4\text{--}85.4\%$ ($76.5\%$ pooled) & $11/11$ cells ($2.6\times$ lower vs.\ MC-4000) \\
\addlinespace[0.6pt]
3.\ Single-Core Latency & Batch / single-record $\text{ms}$ per patient & \textbf{0.62 ms} quad.\ (\textbf{1.23 ms} end-to-end) / \textbf{1.80 ms} single & MC-400: $20.8\text{ ms}$; MC-4000: $203.8\text{ ms}$; MC-$10^5$: $5.0\text{ s}$ & $329\times$ quad.\ speedup vs.\ MC-4000 \\
\addlinespace[0.6pt]
4.\ Counterfactual Monotonicity & Spurious risk rises when setting CAC/emph $=0$ & \textbf{0 / 3,990} ($\mathbf{0.0\%}$) violations & MC-400: $197$ (CAC) \& $449$ (emphysema) rises & $0/3{,}990$ ($0.0\%$) sign violations \\
\addlinespace[0.6pt]
5.\ Reliability Calibration Slope & Slope across $11$ MCAR/MAR cells ($0\%\text{--}60\%$) & \textbf{0.959--1.047} (reliability score); ECE $0.0132 \to 0.0053$ & Untuned gate: $0.76\text{--}0.86$; Pop.-mean: $0.78\text{--}0.89$; Plug-in: $0.71\text{--}0.78$ & Lowest log-loss in $11/11$ cells ($95\%$ CI $< 0$) \\
\addlinespace[0.6pt]
6.\ $40\%$ Missingness Retention & Smaller $\text{AUC}$ loss ($0\% \to 40\%$) vs.\ baselines & \textbf{Higher retention vs.\ all 6} trained baselines at $40\%$ MCAR/MAR & Retention gain at MCAR $40\%$: Cox $+0.008$, XGB-Cox $+0.015$, XGB $+0.018$, RSF $+0.022$, Tree $+0.026$, LR $+0.115$ & $6/6$ trained baselines ($95\%$ CI above zero)$^*$ \\
\addlinespace[0.6pt]
7.\ Overconfident False Negatives & Events predicted below half the training event rate, per $1{,}000$ ($\text{O/E}$) in top-$20\%$ $\sigma_{\text{input}}$ & \textbf{2.27 / 1,000} ($\text{O/E} = \mathbf{0.78}$ pooled; $0.00\text{--}0.96$) & RSF $0.8$, XGB-Cox $4.7$, Cox $5.1$, XGB $9.1$, Tree $19.4$, LR $22.3 / 1{,}000$ & $2.1\text{--}9.8\times$ lower vs.\ five of six baselines \\
\addlinespace[0.6pt]
8.\ Uncertainty-Guided Triage & Retained $\Delta\text{AUC}$ across $0\%\text{--}30\%$ review & \textbf{+0.011} ($20\%$) to \textbf{+0.013} ($30\%$ $\sigma_{\text{input}}$) & Total width $\hat{W}_{90}$ ($\rho = 0.999$ with $\hat{p}$): $-0.057$ ($20\%$) to $-0.069$ ($30\%$) & $+0.068$ to $+0.082$ vs.\ $\hat{W}_{90}$ triage \\
\botrule
\end{tabularx}%
}
\par\vspace{2pt}{\tabnotefont\leftskip=0pt\rightskip=0pt\parfillskip=0pt plus 0.72\linewidth\emergencystretch=1.5em\noindent Note: Bold values highlight primary CE-Agent metrics. Abbreviations: MCAR, missing completely at random; MAR, missing at random; Rel.\ MAE, relative mean absolute error; MC, Monte Carlo; $r$, Pearson correlation; Cond.\ imp., out-of-fold conditional Gaussian imputation; Pop.-mean, population-mean fill; Tree, decision tree; quad., Gauss--Hermite quadrature ($1.80\text{ ms}$ single-record); emph, emphysema; ECE, expected calibration error; LR, logistic regression; XGB, XGBoost; XGB-Cox, Cox-boosted trees; RSF, random survival forest; O/E, observed-to-expected ratio; $\rho$, Spearman rank correlation. $^*$Row 6 reports difference-in-degradation $(\text{AUC}_{0\%}-\text{AUC}_{40\%})_{\text{base}} - (\text{AUC}_{0\%}-\text{AUC}_{40\%})_{\text{CE}}$, with $95\%$ CI $> 0$ across all $6$ baselines under $40\%$ MCAR and MAR (raw $40\%$ MCAR $\text{AUC} = 0.681$, $+0.005\text{ to }+0.109$ over baselines).\par}
\end{table}

\begin{table}[!t]
\centering
\caption{\textbf{Uncertainty propagation with missing data in MESA: CE-Agent versus $40$ model-and-imputer configurations ($n = 1{,}974$).}}\label{tab:app_mesa_uncertainty}
{\fontsize{6.8bp}{8.2bp}\selectfont
\setlength{\tabcolsep}{3.0pt}
\renewcommand{\arraystretch}{1.02}
\begin{tabularx}{\textwidth}{@{}L{0.18\textwidth}YYYY@{}}
\toprule
\colh{Dimension} & \colh{Metric / Setup} & \colh{CE-Agent} & \colh{Baselines} & \colh{Summary} \\
\midrule
1.\ $\hat{W}_{90}$ Interval Accuracy & Rel.\ MAE vs.\ $10^5\text{-draw}$ MC ($r$) & \textbf{0.50\%} med.\ (\textbf{0.52\%} mean) [0.43--0.64\%] ($r \ge 0.99998$) & MC-4000: $2.13\text{--}2.90\%$; MC-400: $5.9\text{--}10.1\%$; $v=0$: $69.3\%$ & $11/11$ cells ($4.8\times$ lower vs.\ MC-4000) \\
2.\ Predictive SD $\hat{\sigma}_p$ & Rel.\ MAE vs.\ $10^5\text{-draw}$ MC & \textbf{1.28\%} med.\ (\textbf{1.95\%} mean) [0.85--4.45\%] & MC-4000: $3.1\text{--}13.6\%$ ($6.3\%$ mean); MC-400: $8.8\text{--}28.0\%$ & $11/11$ cells ($3.3\times$ lower vs.\ MC-4000) \\
3.\ Single-Core Latency & Quadrature / end-to-end $\text{ms}$ per patient & \textbf{0.72 ms} quad.\ / \textbf{1.13 ms} end-to-end & MC-400: $18.9\text{ ms}$; MC-4000: $184.8\text{ ms}$ & $257\times$ quad.\ ($164\times$ end-to-end) vs.\ MC-4000 \\
4.\ Counterfactual Monotonicity & Spurious risk rises across $17$ non-zero edges & \textbf{0 / 2,720} ($\mathbf{0.0\%}$) violations & Monte Carlo sampling produces rank reversals & $0/2{,}720$ ($0.0\%$) sign violations \\
5.\ Reliability Calibration Slope & Slope across $11$ cells (CVD death / Hard ASCVD) & \textbf{0.881--1.083} (CVD) / \textbf{0.92--1.17} (ASCVD) & Untuned: $0.39\text{--}0.46$; Pop.-mean: $0.45\text{--}0.60$; Plug-in: $0.45\text{--}0.51$ & $11/11$ cells within $[0.88, 1.17]$ \\
6.\ $40\%$ Missingness Retention & Smaller $\text{AUC}_{10}$ loss ($0\% \to 40\%$) vs.\ baselines & \textbf{Higher retention vs.\ all 6} main baselines; on par or better vs.\ $\mathbf{40/40}$ under MAR & Gain (MCAR/MAR): RSF $0.021$/$0.044$; Cox $0.103$/$0.051$; LR $0.235$/$0.191$ & Point estimate higher: $6/6$ main; $30/40$ MAR, $18/40$ MCAR Ext.; on par or better, MCAR: $39/40$ ASCVD, $27/40$ CVD \\
7.\ Overconfident False Negatives & Events predicted below half the training event rate, per $1{,}000$ & \textbf{2.07--3.10 / 1,000} (CVD); \textbf{1.7--9.6 / 1,000} (ASCVD) & Untuned: $8.8\text{--}19.7$ (CVD), $24.3\text{--}50.3$ (ASCVD); Ext.: $\le 24.3 / 72.4$ & $2.5\text{--}11.7\times$ lower at $40\%$ MCAR \\
8.\ Uncertainty-Guided Triage & Retained $\Delta\text{AUC}_{10}$ across $0\%\text{--}30\%$ review & \textbf{+0.061} ($30\%$ $\sigma_{\text{input}}$); \textbf{+0.083 / +0.104} (USPSTF $\sigma_{\text{miss}}$) & Triage by $\hat{W}_{90}$ ($\rho \ge 0.990$): $-0.046$ ($15\%$), $-0.036$ ($20\%$), $-0.004$ ($30\%$) & $0.8\%$ flagged at $0\%$ missingness \\
9.\ Targeted CT Triggering & CT budget $0\%\text{--}100\%$ (Stage I graph) & \textbf{30\% CT budget} by $\hat{W}_{90}$ or distance to $7.5\%$ risk $\to \text{AUC}_{10}=\mathbf{0.794}\text{--}\mathbf{0.795}$ & No CT: $0.776$; Random $30\%$ CT: $0.777$ (needs $70\%$ for $0.794$) & $70\%$ CT scans spared ($\Delta\text{AUC}_{10} = 0.000$) \\
\botrule
\end{tabularx}%
}
\tabnote{Note: Bold, primary CE-Agent metrics. MCAR, missing completely at random; MAR, missing at random; Rel.\ MAE, relative mean absolute error; MC, Monte Carlo; $r$, Pearson correlation; SD, standard deviation; quad., Gauss--Hermite quadrature ($2.45\text{ ms}$ single-record end-to-end); Untuned, reliability gate before tuning; Pop.-mean, population-mean fill; on par or better, $95\%$ CI of the loss difference includes zero or favors CE-Agent; main baselines, Cox PH, RSF, GBSA, XGBoost, LR, and DeepSurv (Fig.~\ref{fig:fig6}g: RSF, Cox PH, XGB-Cox, XGBoost, decision tree, LR; all eight models favor CE-Agent by point estimate); Ext., all $40$ model-and-imputer configurations; USPSTF, lung-screening-eligible subcohort~\cite{uspstf2021lung}; $\sigma_{\text{miss}}$, missing-predictor log-odds uncertainty; $\rho$, Spearman rank correlation. Row 9 ranks participants by the $90\%$ predictive-interval width $\hat{W}_{90}$ under a fixed CT budget, a rule distinct from the CT-uncertainty-interval trigger of two-stage triage (Table~\ref{tab:app_interventions_triage}).}
\end{table}

\FloatBarrier
\subsection{Penalty scaling, robust comparators, and CT triage}

\begin{figure}[!htbp]
\centering
\includegraphics[width=0.94\textwidth]{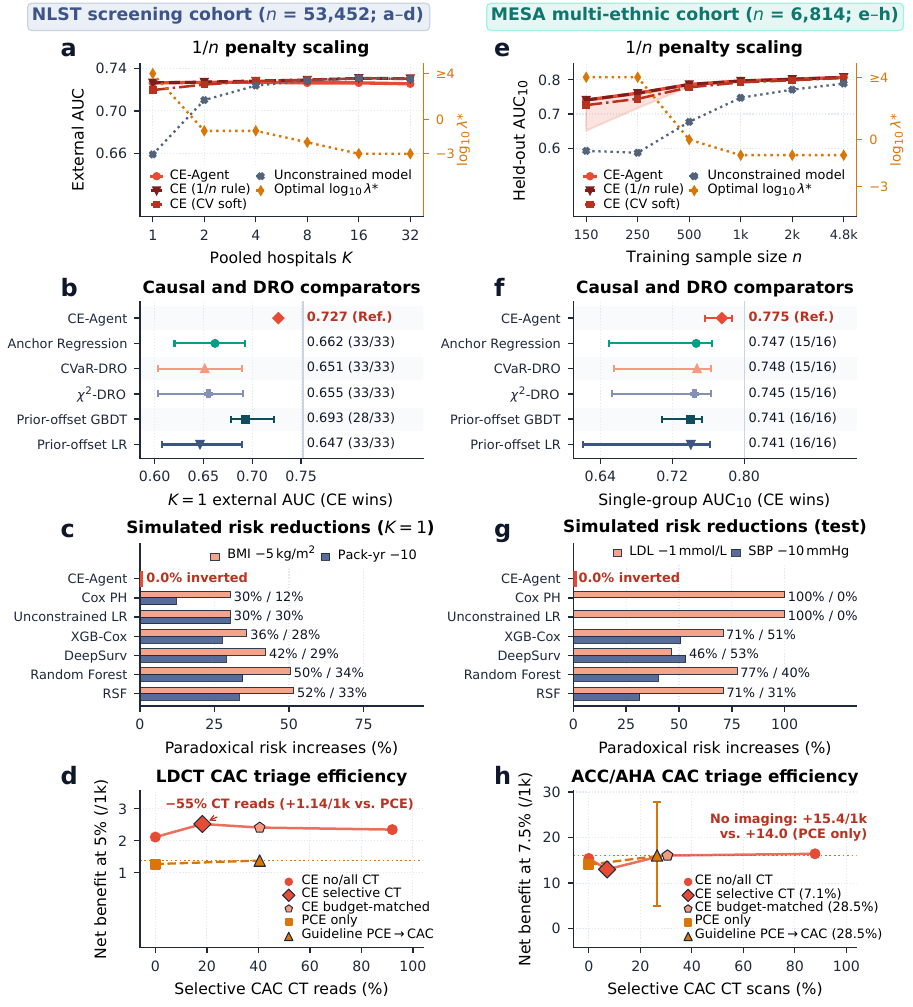}
\caption{\textbf{Penalty scaling, robust-learning comparators, simulated risk-factor reductions, and two-stage CT triage in NLST (a--d) and MESA (e--h).}
\textbf{a},\textbf{e}, External discrimination of CE-Agent (CE) with hard bounds, the $1/n$ penalty (adaptive $1/n\text{-to-CV}$ rule in \textbf{a}; pure $\Lambda_0/n$ in \textbf{e}), or cross-validated (CV) soft bounds against the unconstrained model (left axis), with the AUC-optimal penalty weight $\lambda^*$ (amber, right axis), across pooled NLST hospitals (\textbf{a}) and MESA training subsamples (\textbf{e}; shaded band, small-sample gain from the $1/n$ penalty).
\textbf{b},\textbf{f}, Single-environment external discrimination of CE against Anchor Regression~\cite{rothenhausler2021anchor}, $\chi^2$ and conditional value-at-risk (CVaR) distributionally robust optimization (DRO)~\cite{duchi2021learning}, and prior-offset logistic (LR) model updating~\cite{steyerberg2004validation} and its gradient-boosted decision tree (GBDT) extension across $33$ NLST hospitals (\textbf{b}) and $16$ MESA demographic and scanner transfers (\textbf{f}).
\textbf{c},\textbf{g}, Share of patients whose predicted risk rose under a simulated risk-factor reduction in seven models: BMI and smoking (pack-years, pack-yr) reductions in single-hospital NLST fits (\textbf{c}) and LDL cholesterol and systolic blood pressure (SBP) reductions in MESA (\textbf{g}).
\textbf{d},\textbf{h}, Net benefit per $1{,}000$ patients (/1k) against the share with CAC read, in NLST screening (\textbf{d}) and in MESA adults not on lipid-lowering therapy (\textbf{h}; error bar, guideline $95\%$ CI; markers at $28.5\%$ offset for visibility); dotted line, guideline net benefit. CE reads no or all CAC, reads CAC only when the $90\%$ CT-uncertainty interval crosses the threshold (selective), or reads as many scans as the guideline (budget-matched); PCE is used alone (recalibrated in NLST) or refers intermediate-risk patients for CAC (guideline).}\label{fig:fig9}
\end{figure}

Tables~\ref{tab:app_lambda_rivals}--\ref{tab:app_interventions_triage} and Fig.~\ref{fig:fig9} show how the soft-bound penalty $\lambda(n)$ scales with sample size, compare CE-Agent with robust-learning and model-updating methods~\cite{rothenhausler2021anchor,duchi2021learning,steyerberg2004validation} and with Gaussian priors of matched width, and report simulated risk-factor reductions~\cite{mons2015impact,psc2009body,ctt2010efficacy,ettehad2016blood} and two-stage CT and statin triage~\cite{grundy2019aha,conroy2003estimation,debacker2003european}.

\begin{table}[!htbp]
\centering
\caption{\textbf{Scaling the penalty with sample size and comparison with robust-learning, model-updating, and Gaussian-prior methods in NLST and MESA.}}\label{tab:app_lambda_rivals}
{\fontsize{7.2bp}{8.7bp}\selectfont
\setlength{\tabcolsep}{2.8pt}
\renewcommand{\arraystretch}{1.04}
\begin{tabularx}{\textwidth}{@{}L{0.185\textwidth}YYYY@{}}
\toprule
\colh{Benchmark} & \colh{Protocol} & \colh{CE-Agent} & \colh{Baseline} & \colh{Effect Size} \\
\midrule
\multicolumn{5}{@{}l}{\textit{$(N \times \lambda)$ inverse-sample-size scaling, adaptive weight selection, and separation control}} \\
1.\ NLST $1/n$ Weight Decay & $K \in \{1, \allowbreak 2, \allowbreak 4, \allowbreak 8, \allowbreak 16, \allowbreak 32\}$ centers (up to $36.4\text{k}$ training participants) & $\lambda^*(K)$: $\mathbf{1} \to \allowbreak \mathbf{0.1} \to \allowbreak \mathbf{10^{-2}} \to \allowbreak \mathbf{10^{-3}} \to \allowbreak \mathbf{10^{-4}}$ & Subsample grid $n=250 \to \allowbreak 37.5\text{k}$: $\lambda^* = 1 \to 0$ & Slope rises $0.81 \to 1.00$ as $\lambda^* \propto 1/n$ \\
2.\ NLST Sample-Size Selector & $\lambda_{\text{lit}}(n) = \allowbreak \Lambda_0 / n$ ($\Lambda_0 = 1{,}576.5$) if $n < N_{\min}$ & Oracle gap $\mathbf{\le +0.19 \times 10^{-3}}$ across all $K=1\dots 32$ & Pure inner CV ($K=1$): $+1.06 \times 10^{-3}$ ($\text{AUC } 0.720$) & $+0.0049$ $\text{AUC}$ vs.\ fixed hard at $K=32$ \\
3.\ NLST Prior-Shift Resilience & Synthetic shift $\Delta \in \{0.10, \allowbreak 0.25, \allowbreak 0.50\}$ SD ($28$ cells) & Inner CV (row 2 rule at $n \ge N_{\min}$): oracle gap $\mathbf{\le +7.2 \times 10^{-3}}$ & Fixed hard bounds: up to $+102.5 \times 10^{-3}$ & Inner CV detects shift in $28/28$ cells \\
4.\ MESA Free-Column $\mathcal{N}(0, 1)$ & Unit-normal prior ($\lambda_{\text{free}} = 1/n$) on unpriced cols & $\mathbf{0/50}$ separated; $n=150$: $\mathbf{0.7407}$, $n=250$: $\mathbf{0.7607}$ & Unreg.\ cols: $29/50$ ($0.6512$), $13/50$ ($0.7171$) & $+0.0895$ ($n=150$), $+0.093$ Chinese fold, $+0.101$ scanner fold \\
5.\ MESA Adaptive $1/n$ Schedule & $\Lambda_{0,\text{fr}} / n$ ($\Lambda_{0,\text{fr}} = 31.6$) across $12$ cells & Oracle gap $[\mathbf{-0.9}, \allowbreak \mathbf{+0.7}] \times \mathbf{10^{-3}}$ ($12/12$ CIs cover $0$) & Pure inner CV ($n=250$): $+521 \times 10^{-3}$ oracle gap & Hard ASCVD ($\lambda^*=0.01$): $|\text{slope}-1|$ $0.294 \to 0.096$ \\
\midrule
\multicolumn{5}{@{}l}{\textit{Causal invariance, DRO, Steyerberg updating, and Gaussian MAP controls}} \\
6.\ Causal Anchor Regression~\cite{rothenhausler2021anchor} & $g \in [0, 10^3]$ across $33$ NLST \& $16$ MESA OOD folds & NLST med.\ $\mathbf{0.7268}$; MESA med.\ $\mathbf{0.7752}$ & Anchor best $g$: NLST $0.6619$; MESA $0.7466$ & $33/33$ NLST ($+0.0649$ in medians); $15/16$ MESA (med.\ paired $+0.0345$) \\
7.\ Label-Free $\chi^2$ \& CVaR DRO~\cite{duchi2021learning} & Oracle across $7$ $\chi^2$ radii \& $7$ CVaR tail levels & NLST med.\ $\mathbf{0.7268}$; MESA med.\ $\mathbf{0.7752}$ & Oracle DRO: NLST $0.6460$; MESA $0.7446$ / $0.7475$ & $33/33$ NLST; $15/16$ MESA (med.\ paired $+0.0214$ / $+0.0262$) \\
8.\ Steyerberg Offset Updating~\cite{steyerberg2004validation} & Prior-offset LR \& base-margin XGBoost on CE-Agent's predicted log-odds & Prevalence-only envelope: $\mathbf{0.7268}$ / $\mathbf{0.7752}$ & Offset LR / XGB: NLST $0.6465$ / $0.6929$; MESA $0.7406$ & Outperforms LR / XGB in $33/33$ \& $28/33$ NLST, $16/16$ MESA \\
9.\ Gaussian Point-Prior MAP & Width-matched $\mathcal{N}(c_j^{\text{pub}}, \sigma_j^2)$, $\sigma_j = \allowbreak (u_j-\ell_j)/3.92$ & KKT exact $\hat{c}_j=0$: $\mathbf{95.1\%\text{--}98.6\%}$ of fits & Gaussian MAP exact $\hat{c}_j=0$: $0.0\%$ ($68\text{--}115/200$ TAC flip) & Retains full $\text{AUC}$ ($|\Delta|\le 0.001$), preserves sparsity \\
\botrule
\end{tabularx}%
}
\tabnote{Note: Bold values highlight primary CE-Agent metrics. Abbreviations: $\lambda^*$, best-performing penalty weight (lowest out-of-sample log loss; Fig.~\ref{fig:fig9}a,e plot the weight with the highest $\text{AUC}$); $\lambda_{\text{lit}}$, penalty weight set from published interval widths; Oracle gap, out-of-sample log-loss excess over $\lambda^*$; CV, cross-validation; cols, columns; Unreg., unregularized; $\Lambda_{0,\text{fr}}$, development-calibrated prior precision scale ($31.6$; closed-form $\Lambda_0 = 2{,}289.5$ in Fig.~\ref{fig:fig9}e) under free-column $\mathcal{N}(0, 1)$ stabilization; $g$, Anchor Regression invariance weight; DRO, distributionally robust optimization (Row 7 reports the best single fixed-hyperparameter median across the $33$ NLST hospitals, $0.6460$, whereas Fig.~\ref{fig:fig9}b plots per-hospital hyperparameter oracles, $0.651$ and $0.655$); CVaR, conditional value-at-risk; med., median across $33$ single-hospital NLST OOD folds or all $16$ MESA single-group OOD folds ($4$ race and $4$ scanner groups across cardiovascular mortality and hard ASCVD); LR, logistic regression; XGB, XGBoost; Prevalence-only envelope, prevalence-recalibrated envelope without slope refitting; Offset LR / XGB, Steyerberg prior-offset logistic revision~\cite{steyerberg2004validation} and base-margin boosted-tree updating (prior-offset GBDT in Fig.~\ref{fig:fig9}b,f); MAP, maximum a posteriori; KKT, Karush--Kuhn--Tucker; TAC, thoracic aortic calcification.}
\end{table}

\begin{table}[!htbp]
\centering
\caption{\textbf{Simulated risk-factor reductions and two-stage triage in NLST and MESA.}}\label{tab:app_interventions_triage}
{\tabfont
\setlength{\tabcolsep}{2.8pt}
\renewcommand{\arraystretch}{1.18}
\begin{tabularx}{\textwidth}{@{}L{0.185\textwidth}YYYY@{}}
\toprule
\colh{Benchmark} & \colh{Protocol} & \colh{CE-Agent} & \colh{Baseline} & \colh{Clinical Effect} \\
\midrule
\multicolumn{5}{@{}l}{\textit{Counterfactual interventional direction consistency across single-environment and pooled fits}} \\
1.\ NLST Smoking \& BMI ($K=1$) & Cessation, $-10\text{/}{-20}\text{ pack-yr}$, $\text{BMI }-5$~\cite{mons2015impact,psc2009body} & $\mathbf{0.0\%}$ rises ($\mathbf{0/33}$ $K=1$, $\mathbf{0/50}$ subsample rev.) & $K=1$ rev.: $9\%$ quit, $21\%$ pack-yr, $41\%$ BMI ($51.6\%$ rises) & LOCO BMI rev.\ in $4/8$ ML ($55\%\text{--}62\%$ rises)$^*$ \\
2.\ MESA LDL ($-1\text{ mmol/L}$) & Counterfactual $\Delta\text{LDL}$ ($n=1{,}968$ with recorded lipids) vs.\ CTT RCT~\cite{ctt2010efficacy} & $\mathbf{0.0\%}$ discordant rises ($\text{RR} = \mathbf{0.865}$, $0/48$ fits) & $20/24$ ML configs reversed ($46\%\text{--}100\%$ rises) & Eliminates statin-confounded lipid inversion \\
3.\ MESA SBP ($-10\text{ mmHg}$) & Counterfactual $\Delta\text{SBP}$ vs.\ Ettehad RCT~\cite{ettehad2016blood} & $\mathbf{0.0\%}$ discordant rises ($\text{RR} = \mathbf{0.876}$, $0/48$ fits) & RSF $31.3\%$, RF $40.2\%$, XGB-Cox $50.8\%$, DeepSurv $53.2\%$ rises & Prevents tree/neural local non-monotonicity \\
\midrule
\multicolumn{5}{@{}l}{\textit{Two-stage clinical CT reading and statin triage efficiency}} \\
4.\ NLST Opportunistic LDCT Triage & Primary-prev.\ LDCT train ($n=15{,}727$, $31$ non-zero-event hospital fits, $p_t=5.0\%$)~\cite{conroy2003estimation,debacker2003european} & LOCO $\mathbf{+2.52\text{/1k}}$ ($18.3\%$ CT); $K=1$ $\mathbf{+1.48\text{/1k}}$ & Intermediate $\text{PCE}\to\text{CAC}$: $+1.38\text{/1k}$ ($40.5\%$ CT); $K=1$ ML $<0$ & $-55\%$ reads ($+1.14\text{/1k}$, $95\%$ CI $0.23\text{--}2.13$, $P=0.019$); $27/31$ vs.\ $K=1$ ML$^\dagger$ \\
5.\ MESA 2018 ACC/AHA CAC Triage & Lipid-therapy-free test ages $45\text{--}75$ ($n=1{,}465$, $p_t=7.5\%$)~\cite{grundy2019aha} & Equal-budget ($28.5\%$ CT): $\mathbf{+16.0\text{/1k}}$; Two-stage selective CT ($7.1\%$ CT, $24.9\%$ statin): $\mathbf{+13.0\text{/1k}}$ & Guideline $\text{PCE}\to\text{CAC}$: $+16.1\text{/1k}$ ($28.5\%$ CT, $42.2\%$ statin)$^\dagger$ & Parity at $28.5\%$ CT; $-75\%$ CT \& $-41\%$ statin within $95\%$ CI \\
\botrule
\end{tabularx}%
}
\tabnote{Note: Bold highlights primary CE-Agent metrics. pack-yr, pack-years; rises, predicted risk increases under a beneficial risk-factor reduction; rev., cohort-level direction reversals; LOCO, leave-one-center-out; CTT, Cholesterol Treatment Trialists' Collaboration; CHD, coronary heart disease; RCT, randomized controlled trial; RR, risk ratio; SBP, systolic blood pressure; RSF, random survival forest; RF, random forest; XGB-Cox, Cox-boosted trees; $p_t$, decision risk threshold ($5.0\%$ fatal-CVD risk~\cite{conroy2003estimation,debacker2003european} in NLST; $7.5\%$ hard ASCVD risk~\cite{grundy2019aha} in MESA); /1k, net benefit per $1{,}000$ patients; Intermediate $\text{PCE}\to\text{CAC}$, CAC referral at intermediate risk by recalibrated PCE; Two-stage selective CT, CAC read only when the $90\%$ CT-uncertainty interval crosses $p_t$. $^*$With point anchors (or soft $1/n$ bounds), all $33$ single-hospital fits move risk in the expected direction under the four NLST perturbations ($19/33\text{--}29/33$ without point anchors). Rows 2--3: CTT~\cite{ctt2010efficacy} reports CHD mortality $\text{RR} = 0.80$ ($99\%$ CI $0.74\text{--}0.87$) and vascular mortality $\text{RR} = 0.86$ ($95\%$ CI $0.82\text{--}0.90$) per $1\text{ mmol/L}$ LDL reduction, and Ettehad et al.~\cite{ettehad2016blood} report major CVD $\text{RR} = 0.80$ ($0.77\text{--}0.83$) and all-cause mortality $\text{RR} = 0.87$ ($0.84\text{--}0.91$) per $10\text{ mmHg}$ SBP reduction. $^\dagger$Rows 4--5 use the Stage I propagation graph (published weights). In NLST, selective CT with Stage II hard-bound weights gives $+2.34\text{/1k}$ at $6.7\%$ CT reads, $+0.96\text{/1k}$ above intermediate $\text{PCE}\to\text{CAC}$ referral ($P = 0.057$; $+2.42\text{/1k}$ at $0\%$ CT); in MESA, CE-Agent gives $+15.4\text{/1k}$ and PCE alone $+14.0\text{/1k}$ at $0\%$ CT scans.}
\end{table}

\FloatBarrier
\subsection{Lung cancer and joint cardiopulmonary triage}

Tables~\ref{tab:app_lung_main}--\ref{tab:app_dual_triage} report held-out test ($n = 15{,}931$; LDCT arm $n = 7{,}786$) and $33\text{-hospital}$ out-of-center ($n = 37{,}521$; LDCT arm $n = 18{,}519$) benchmarks for $6\text{-year}$ competing-risk lung cancer incidence and $10\text{-year}$ lung cancer mortality, complementing Fig.~\ref{fig:fig_lung_dual}. Evaluations span single-hospital generalization ($K = 1 \to 32$), PLCOm2012-free sign-only controls, progressive LDCT integration, smoking counterfactuals, missing-data uncertainty propagation, and joint cardiopulmonary triage.

\begin{table}[!htbp]
\centering
\caption{\textbf{NLST lung cancer: CE-Agent, PLCOm2012, and ML baselines in the test cohort ($n = 15{,}931$) and in leave-one-center-out validation in the LDCT arm ($n = 18{,}519$).} Six-year competing-risk $\text{AUC}_{\text{cr}}$ ($575$ held-out and $1{,}350$ training $6\text{-year}$ incident lung cancers, with $618$ and $1{,}440$ in total; LDCT arm $317$ and $694$ $6\text{-year}$ cases, with $338$ and $732$ total), held-out calibration slope and $\text{E/O}_{\text{cr}}$, and $10\text{-year}$ competing-risk lung cancer mortality $\text{AUC}_{\text{cr}}$ ($535$ held-out and $1{,}242$ training $10\text{-year}$ deaths; $747$ and $1{,}663$ total).}\label{tab:app_lung_main}
{\fontsize{7.0bp}{9.0bp}\selectfont
\setlength{\tabcolsep}{0pt}
\renewcommand{\arraystretch}{1.12}
\begin{tabular*}{\textwidth}{@{\extracolsep{\fill}}lcccc@{}}
\toprule
\colh{Model / Specification} & \colh{LOCO LDCT} & \colh{Test $\text{AUC}_{\text{cr}}$} & \colh{Slope ($\text{E/O}_{\text{cr}}$)} & \colh{10-yr Death$^\S$} \\
\midrule
CE-Agent (hard, primary) & 0.727 (0.708--0.745)$^*$ & 0.708 (0.689--0.728) & \textbf{0.99} (0.99) & 0.714 (0.693--0.735) \\
CE-Agent (soft, $\Lambda_0/n$) & \textbf{0.728} (0.709--0.746)$^*$ & 0.709 (0.691--0.729) & \textbf{0.99} (0.99) & 0.716 (0.696--0.738) \\
CE-Agent (sign-only)$^\dagger$ & \textbf{0.728} (0.708--0.746)$^*$ & 0.710 (0.691--0.729) & 0.98 (0.99) & 0.715 (0.695--0.737) \\
\midrule
Cox PH (CIF, 11 pred.) & 0.727 (0.707--0.746) & \textbf{0.711} (0.691--0.731) & 2.76 (\textbf{1.00}) & 0.713 (0.693--0.735) \\
Logistic Reg.\ (11 pred.) & 0.727 (0.708--0.745) & 0.710 (0.690--0.729) & 0.92 (1.10) & 0.712 (0.692--0.734) \\
XGBoost (11 pred.) & 0.726 (0.707--0.744) & 0.709 (0.688--0.729) & 0.87 (1.08) & 0.706 (0.685--0.727) \\
DeepSurv (11 pred.) & 0.725 (0.705--0.744) & 0.708 (0.686--0.728) & 0.92 (1.01) & --- \\
RSF (11 pred.) & 0.710 (0.690--0.729) & 0.695 (0.674--0.715)$^\ddagger$ & 0.90 (1.02) & --- \\
GBSA (11 pred.) & --- & 0.666 (0.644--0.689)$^\ddagger$ & 4.19 (\textbf{1.00}) & --- \\
\midrule
PLCOm2012 (recal.)~\cite{tammemagi2013selection} & 0.704 (0.685--0.725) & 0.704 (0.684--0.724) & 0.92 (0.99) & \textbf{0.723} (0.703--0.742) \\
PLCOm2012 (published)~\cite{tammemagi2013selection} & 0.704 (0.685--0.725) & 0.704 (0.684--0.724) & 0.92 (0.86) & \textbf{0.723} (0.703--0.742) \\
\botrule
\end{tabular*}%
}
\tabnote{Note: Bold marks the best discrimination or calibration in each column ($|\text{slope} - 1|$ and $|\text{E/O}_{\text{cr}} - 1|$ closest to zero). LOCO, $33\text{-hospital}$ leave-one-center-out development validation ($n = 18{,}519$ LDCT); CIF, cumulative incidence function; LR, logistic regression; RSF, random survival forest; GBSA, gradient-boosted survival analysis; recal., prevalence-recalibrated. $^*$Paired development (LOCO LDCT; not preregistered) difference from recalibrated PLCOm2012: $+0.023$ ($95\%$ CI $+0.010\text{ to }+0.036$) for CE-Agent and $+0.024$ ($+0.011\text{ to }+0.038$) for sign-only CE-Agent; on the held-out test, $\Delta\text{AUC}_{\text{cr}} = +0.004$ ($95\%$ CI $-0.008\text{ to }+0.017$; $P = 0.542$; preregistered noninferiority margin $-0.02$). $^\dagger$Uses only the direction of each risk factor, with no PLCOm2012 regression coefficients; the preregistered comparison of compiled with sign-only bounds gave $\Delta\text{AUC}_{\text{cr}} = -0.0017$ ($95\%$ CI $-0.0040\text{ to }+0.0005$), within the $\pm 0.01$ equivalence margin. $^\ddagger$Held-out comparisons of CE-Agent with $11\text{-predictor}$ baselines: RSF, $P = 0.040$; GBSA, $P < 0.001$. Against baselines given an expanded $19\text{-node}$ encoding that adds visual emphysema and missingness indicators, the $11\text{-predictor}$ CE-Agent stays calibrated (slope $0.99$, $\text{E/O}_{\text{cr}} = 0.99$), whereas pooled Cox CIF ($\text{AUC}_{\text{cr}} = 0.724$; slope $2.59$) and LR ($0.715$; $\text{E/O}_{\text{cr}} = 1.10$) gain discrimination with poorer calibration. $^\S$The same $11\text{-predictor}$ envelope fitted to $10\text{-yr}$ lung cancer mortality ($\text{E/O}_{\text{cr}} = 0.98$ vs.\ $0.93$ for published PLCOm2012; noninferior to recalibrated PLCOm2012: $\Delta\text{AUC}_{\text{cr}} = -0.009$, $95\%$ CI $-0.019\text{ to }+0.001$, within the preregistered $-0.02$ margin; two-sided $P = 0.082$). Refitting on the mortality label raised $\text{AUC}_{\text{cr}}$ by $0.011$ ($P < 0.001$) over the incidence weights shifted to the $10\text{-year}$ horizon ($0.702$).}
\end{table}

\begin{table}[!htbp]
\centering
\caption{\textbf{NLST lung cancer: transfer across hospitals, added CT findings, simulated smoking reductions, and missing data.}}\label{tab:app_lung_ood_mech}
{\fontsize{6.8bp}{8.0bp}\selectfont
\setlength{\tabcolsep}{2.7pt}
\renewcommand{\arraystretch}{1.01}
\begin{tabularx}{\textwidth}{@{}L{0.20\textwidth}YYYY@{}}
\toprule
\colh{Evaluation Dimension} & \colh{Protocol / Cohort} & \colh{CE-Agent} & \colh{Comparators} & \colh{Key Finding} \\
\midrule
1.\ Single-Hospital OOD ($K=1 \to 32$) & $33$ single-site fits $\times 8$ ML baselines ($264$ cells) & \textbf{246 / 264} (\textbf{93.2\%}) wins; med.\ per-site $\text{AUC}_6 = \mathbf{0.665}$ ($0.679$ pooled external) & Sign-only: $\mathbf{244/264}$ ($92.4\%$, med.\ per-site $0.657$); RSF $0.633$, Cox $0.635$, GBSA $0.631$, LR $0.620$, XGB $0.611$, DeepSurv $0.586$ & $+0.052$ med.\ $\Delta\text{AUC}_6$ across external sites ($P < 0.001$) \\
\addlinespace[1.0pt]
2.\ Multi-Center Scaling & $K = 1, 2, 4, 8, 16, 32$ hospitals ($33$ subsets/level) & $\text{AUC}_6$: $\mathbf{0.679} \to \mathbf{0.702} \to \mathbf{0.708} \to \mathbf{0.713} \to \mathbf{0.718} \to \mathbf{0.715}$ & Unconstrained $11\text{-pred.}$: $0.653 \to 0.694 \to 0.704 \to 0.713 \to 0.718 \to 0.708$; XGB: $0.629 \to 0.712$ & $\lambda_{\text{lit}}(n)$ decays $0.283 \to 0.006$; slope within $[0.80, 1.25]$ in $36\% \to 70\% \to 88\% \to 100\% \to 100\% \to 70\%$ \\
\addlinespace[1.0pt]
3.\ Domain-Gen.\ \& Riley $N_{\min}$ & $K=1$ external sites vs.\ DRO / Anchor / updating & Med.\ $\mathbf{0.679}$; Riley $N_{\min} = \mathbf{3{,}296}$ ($p_{\text{eff}} = 6$) & Anchor: $0.633$; $\chi^2\text{-DRO}$: $0.617$; CVaR-DRO: $0.629$; Offset LR/XGB: $0.625/0.650$; Unconstr.\ $11\text{-param.}$: $8{,}383$; $34\text{-col.}$ LR: $37{,}183$ & $2.5\text{--}11.3\times$ lower $N_{\min}$ ($5.7\times$ at matched $C = 0.679$); LOCO NB $>0$ in $27/33$ sites \\
\addlinespace[1.0pt]
4.\ Progressive LDCT Integration & Clinical only $\to$ + LDCT nodule $\to$ + Visual emphysema & Held-out ($n=7{,}786$): $0.685 \to 0.700 \to \mathbf{0.708}$ ($\mathbf{0.710}$ sign-only); LOCO: $0.700 \to 0.726 \to \mathbf{0.729}$ & Held-out full LDCT: LR $0.701$ ($P=0.036$), XGB $0.696$, PLCOm2012 $0.696$; LOCO full LDCT: PLCOm2012 $0.704$$^*$ & $+0.023$ held-out CT gain ($P = 0.036$; nodule step $+0.015$, $P=0.102$; full cohort $+0.009$, $P=0.098$) \\
\addlinespace[1.0pt]
5.\ Smoking Counterfactuals & Quit smoking ($18{,}164$), $-10 / -20$ pack-yr ($27{,}818 / 17{,}798$) & \textbf{0 / 33} ($\mathbf{0.0\%}$) hospital fits with any risk inversion (quit $\text{RR} = 0.65$) & Sign-only: $\mathbf{0/33}$ ($0.0\%$); Unconstr.\ LR/Cox: $3\text{--}5/33$; XGB: $16\text{--}27/33$; RSF: $31\text{--}32/33$; DeepSurv: $33/33$ & Exact smoking monotonicity across $100\%$ of single-site fits$^\dagger$ \\
\addlinespace[1.0pt]
6.\ Missing-Data Uncertainty & $11$ MCAR/MAR cells ($0\%\text{--}60\%$, $n=37{,}521$, Stage I lung graph) & $\hat{W}_{90}$ rel.\ MAE \textbf{0.64--0.79\%} ($r \ge \mathbf{0.9999}$); \textbf{1.21 ms}/pt; gated slope \textbf{0.906--1.071} & MC-4000: $1.55\text{--}2.31\%$ ($93.7\text{ ms}$, $77.5\times$ slower); Plug-in slope: $0.512\text{--}0.607$; Pop.-mean: $0.519\text{--}0.650$ & ECE reduced $3\text{--}4\times$ ($0.0029\text{--}0.0063$); full-CT risk inside the no-CT PI under unattenuated projection: $99.98\%$ \\
\botrule
\end{tabularx}%
}
\tabnote{Note: Bold highlights primary CE-Agent metrics. OOD, out-of-distribution; med., median; DRO, distributionally robust optimization; CVaR, conditional value-at-risk; Anchor, Anchor Regression~\cite{rothenhausler2021anchor}; Offset LR/XGB, prior-offset logistic updating~\cite{steyerberg2004validation} and base-margin XGBoost; NB, net benefit; pt, patient; Pop.-mean, population-mean fill; ECE, expected calibration error; PI, $90\%$ predictive interval. Rows 1--3, 5, and 6 and leave-one-center-out (LOCO) values are post hoc training-cohort analyses; the held-out CT gain in row 4 is exploratory. Row 1: per-site values are medians across the $33$ single-hospital fits (median $n \approx 792$) of mean $\text{AUC}_6$ at the other $32$ hospitals; pooled external values (also rows 2--3) score those hospitals together. CE-Agent meets Riley's $N_{\min} = 3{,}296$ in the two largest centers ($0/33$ unconstrained). Pooling two or more centers raises CE-Agent to $0.702\text{--}0.718$; published PLCOm2012, which needs no local fitting, leads single-site fits (per-site median $0.696$, $27/33$ centers). $^*$Lung models use radiologist reads of the baseline scan (nodules; visual emphysema in the LDCT-arm extension) and are independent of Sybil~\cite{mikhael2023sybil} (Appendix~\ref{app:protocols}). $^\dagger$The inverse BMI--lung cancer association reported for PLCOm2012~\cite{tammemagi2013selection} is a pre-diagnostic association, not a weight-loss intervention target.}
\end{table}

\begin{sidewaystable}
\centering
\caption{\textbf{Joint cardiopulmonary triage in the NLST test cohort: combined net benefit of eight strategies and risk quadrants ($n = 15{,}931$; LDCT arm $n = 7{,}786$).}}\label{tab:app_dual_triage}
{\fontsize{7.6bp}{9.6bp}\selectfont
\setlength{\tabcolsep}{0pt}
\renewcommand{\arraystretch}{1.15}
\begin{tabular*}{\textheight}{@{\extracolsep{\fill}}lccccc@{}}
\toprule
\colh{Strategy / Quadrant} & \colh{CT / Flag} & \colh{All ($1.5\%/5\%$)} & \colh{LDCT ($1.5\%/5\%$)} & \colh{All ($2\%/5\%$)} & \colh{LDCT ($2\%/5\%$)} \\
\midrule
\multicolumn{6}{@{}l}{\textit{Joint cardiopulmonary triage net benefit $\text{NB}_{\text{sum}}$ per $1{,}000$ participants (95\% CI)}} \\
Dual CE-Agent (full CT) & 100\% / 83.1\% & \textbf{+30.2}\,(26.8--33.9) & \textbf{+33.4}\,(28.6--38.9) & \textbf{+26.6}\,(23.2--30.0) & \textbf{+29.2}\,(24.4--34.5) \\
Dual CE-Agent (sign-only) & 100\% / 82.0\% & +30.0\,(26.5--33.6) & +32.8\,(27.9--38.3) & +26.4\,(23.0--29.9) & +29.0\,(24.1--34.2) \\
Two-stage selective CT$^\dagger$ & \textbf{61.1\%} / 84.9\% & +29.6\,(26.2--33.3) & +32.6\,(27.7--37.9) & +26.1\,(22.8--29.6) & +28.5\,(24.0--33.8) \\
PLCOm2012 + PCE (recal.) & 0\% / 80.4\% & +28.6\,(25.3--32.1)$^*$ & +31.8\,(27.3--37.1) & +25.2\,(22.0--28.7) & +27.1\,(22.5--32.7) \\
PLCOm2012 (pub.) + PCE (recal.) & 0\% / 73.6\% & +28.7\,(25.4--32.3)$^*$ & +31.6\,(26.9--37.0) & +24.7\,(21.6--28.1)$^*$ & +26.4\,(21.9--31.4)$^*$ \\
Clinic-only (no CT) & 0\% / 87.4\% & +28.4\,(25.1--32.0)$^*$ & +31.4\,(26.7--36.8) & +24.4\,(21.1--28.0)$^*$ & +26.6\,(22.0--31.8) \\
USPSTF + PCE (recal.) & 0\% / 99.4\% & +27.5\,(24.2--31.2)$^*$ & +30.8\,(26.2--36.2)$^*$ & +22.5\,(19.2--26.3)$^*$ & +25.9\,(21.2--31.3)$^*$ \\
USPSTF + PCE (pub., $7.5\%$) & 0\% / 99.4\% & +18.4\,(14.6--22.4)$^*$ & +20.8\,(15.7--26.8)$^*$ & +13.5\,(9.6--17.5)$^*$ & +15.9\,(10.7--21.9)$^*$ \\
\midrule
\multicolumn{6}{@{}l}{\textit{Held-out dual CE-Agent risk quadrants at $1.5\%$ lung cancer \& $5.0\%$ fatal-CVD cutoffs ($n = 15{,}931$)}} \\
Dual high ($\text{Lung}^+ \text{CVD}^+$) & 17.8\% ($2{,}832$) & \multicolumn{4}{l@{}}{$170$ ($29.5\%$) 6-yr lung cancers; $226$ ($42.9\%$) 10-yr CVD deaths} \\
Lung-only ($\text{Lung}^+ \text{CVD}^-$) & 65.3\% ($10{,}401$) & \multicolumn{4}{l@{}}{$385$ ($\mathbf{67.0\%}$) lung cancers (CVD-missed); $243$ ($46.5\%$) CVD deaths} \\
CVD-only ($\text{Lung}^- \text{CVD}^+$) & 0.8\% ($135$) & \multicolumn{4}{l@{}}{$2$ ($0.3\%$) lung cancers; $12$ ($2.3\%$) CVD deaths ($\mathbf{45.2\%}$ $\text{CVD}^+$ total)} \\
Dual low ($\text{Lung}^- \text{CVD}^-$) & 16.1\% ($2{,}563$) & \multicolumn{4}{l@{}}{$18$ ($3.2\%$) 6-yr lung cancers; $44$ ($8.3\%$) 10-yr CVD deaths} \\
\botrule
\end{tabular*}%
}
\tabnote{Note: Combined net benefit $\text{NB}_{\text{sum}} = \text{NB}_{\text{lung}} + \text{NB}_{\text{CVD}}$ per $1{,}000$ participants at the preregistered $1.5\%$ (sensitivity $2.0\%$) $6\text{-year}$ competing-risk lung cancer incidence and $5.0\%$ $10\text{-year}$ fatal-CVD incidence cutoffs~\cite{tammemagi2014evaluation,tammemagi2021ontario,conroy2003estimation,debacker2003european}; Dual CE-Agent, the lung cancer and cardiovascular CE-Agent models applied together; Clinic-only, lung cancer CE-Agent without CT inputs paired with recalibrated PCE; pub., published; recal., prevalence-recalibrated. CT / Flag, fraction with CT features read / fraction flagged for lung-cancer follow-up in the $n = 15{,}931$ cohort at $1.5\%/5.0\%$ ($99.4\%$ meet USPSTF 2021 eligibility by NLST enrollment design). Quadrant cells give uncensored $6\text{-year}$ lung cancer ($n = 575$) and $10\text{-year}$ cardiovascular death ($n = 525$) counts with inverse-probability-of-censoring weighted competing-risk shares. $^*$Two-sided paired bootstrap $P < 0.05$ versus Dual CE-Agent in the same column; the full-cohort comparison with USPSTF+PCE (recal.) at $1.5\%/5.0\%$ is the preregistered test ($P < 0.001$); the two-stage comparisons$^\dagger$ at $1.5\%/5.0\%$ are also preregistered, and all others are exploratory. In the LDCT arm ($n = 7{,}786$), where every imaging input is read from the screening scan itself, Dual CE-Agent exceeds USPSTF+PCE (recal.) by $+2.6\text{/1k}$ ($P = 0.024$). $^\dagger$CT features are read when the $90\%$ CT-uncertainty interval (CT-derived predictors treated as unmeasured) contains either cutoff: $61.1\%$ reads at $1.5\%/5.0\%$, $-0.58\text{/1k}$ vs.\ full CT ($95\%$ CI $-1.26\text{ to }0.00$; within the preregistered $-2.0\text{/1k}$ noninferiority margin) and $+0.32\text{/1k}$ vs.\ equal-budget random CT reading ($P = 0.51$).}
\end{sidewaystable}

\FloatBarrier
\clearpage
\section{Mathematical details and proofs}\label{app:math}
\setcounter{equation}{0}

\subsection{Domain graph, binary biomarkers, and composite indices}

In structural causal screening (Methods), the candidate domain graph is a directed acyclic graph $\mathcal{G}_{\text{CE}} = (\mathcal{V}, \mathcal{E}, \tau, \rho)$ with node set $\mathcal{V}$ and directed edge set $\mathcal{E}$, in which each edge is marked as an established dependency (supported by at least one cited reference) or a plausible, context-specific one. Each node $v \in \mathcal{V}$ carries a timing index $\tau(v) \in \{T_0, T_{>0}\}$ distinguishing baseline $T_0$ from post-baseline $T_{>0}$ variables and a clinical role $\rho(v) \in \{\text{exposure},\allowbreak\ \text{mediator},\allowbreak\ \text{outcome},\allowbreak\ \text{decision},\allowbreak\ \text{unmeasured}\}$ (decision nodes are clinical actions, such as starting a drug, and are excluded from baseline predictors). Each derived linear composite $C_\ell$ is formed from its standardized baseline parents $\operatorname{pa}(C_\ell)$ with fixed weights $w_{p\ell} \in \mathbb{R}$ as $C_\ell = \allowbreak \sum_{p \in \operatorname{pa}(C_\ell)} w_{p\ell} Z_p$. Adding $\gamma_\ell C_\ell$ for $\gamma_\ell \in \mathbb{R}$ to the baseline linear predictor $c_0 + \sum_{j \in \mathcal{V}_{\text{adm}}} c_j Z_j$ shifts the parent weights to $\tilde{c} \in \mathbb{R}^d$:
\begin{equation}\label{eq:mediator_shift}
\tilde{c}_p = c_p + \gamma_\ell w_{p\ell} \quad \forall p \in \operatorname{pa}(C_\ell), \qquad \tilde{c}_j = c_j \quad \forall j \in \mathcal{V}_{\text{adm}} \setminus \operatorname{pa}(C_\ell).
\end{equation}
For missing binary intermediate biomarkers $M_m \in \{0, 1\}$ with training prevalence $\hat{\pi}_m$ and standard deviation $\hat{s}_m = \sqrt{\hat{\pi}_m(1 - \hat{\pi}_m)}$, sign-constrained logistic regression on standardized parent moments $(\mu_{ip}, v_{ip})$ yields conditional probability $q_{im} = \operatorname{expit}\!\bigl(\hat{b}_m + \sum_{p \in \operatorname{pa}(M_m)} \hat{w}_{pm} \mu_{ip}\bigr)$, standardized conditional mean $\mu_{im} = (q_{im} - \hat{\pi}_m)/\hat{s}_m$, and delta-method variance $v_{im} = \bigl[q_{im}(1 - q_{im}) + q_{im}^2(1 - q_{im})^2 \sum_{p \in \operatorname{pa}(M_m)} \hat{w}_{pm}^2 v_{ip}\bigr] / \hat{s}_m^2$, complementing continuous linear projection in equation~(\ref{eq:projection}). For latent mediators $L_r \in \mathcal{L}$ inferred linearly from standardized upstream parents with residual variance $v_L = 0.5$, the downstream edge into the outcome carries reliability gate $g_{iL_r}(\alpha_g) = \operatorname{expit}(\alpha_0 - \alpha_m - \alpha_v v_{iL_r})$.

\begin{proposition}[Shared-parent composite rank deficiency]\label{prop:mediator}
Let the standardized baseline feature matrix $Z \in \mathbb{R}^{n \times d}$ over $\mathcal{V}_{\mathrm{adm}}$ have full column rank $\operatorname{rank}(Z) = d$, and let $C_\ell = \sum_{p \in \operatorname{pa}(C_\ell)} w_{p\ell} Z_p$. If $\operatorname{pa}(C_\ell) \subseteq \mathcal{V}_{\mathrm{adm}}$, then $\operatorname{rank}([Z, C_\ell]) = \operatorname{rank}(Z) = d$, and for any diagonal weight matrix $W \succeq 0$ the unregularized Fisher matrix $\widetilde{Z}^\top W \widetilde{Z}$ of $\widetilde{Z} = [Z, C_\ell] \in \mathbb{R}^{n \times (d+1)}$ is singular. Moreover, for all-ones $\mathbf{1} \in \mathbb{R}^n$, $\eta = c_0 \mathbf{1} + Zc + \gamma_\ell C_\ell$ equals $\eta = c_0 \mathbf{1} + Z\tilde{c}$ on every participant under equation~(\ref{eq:mediator_shift}). Conversely, $\operatorname{rank}([Z, C_\ell]) = d + 1$ if and only if $\sum_{p \in \operatorname{pa}(C_\ell) \setminus \mathcal{V}_{\mathrm{adm}}} w_{p\ell} Z_p \notin \operatorname{col}(Z)$.
\end{proposition}
\begin{proof}
When $\operatorname{pa}(C_\ell) \subseteq \mathcal{V}_{\text{adm}}$, $C_\ell$ lies in the column space $\operatorname{col}(Z)$ of the parent columns, giving $\operatorname{rank}([Z, C_\ell]) = \operatorname{rank}(Z) = d$. Defining $u_{\text{null}} = [w_\ell^\top, -1]^\top \in \mathbb{R}^{d+1} \setminus \{0\}$, where $w_\ell \in \mathbb{R}^d$ has entries $w_{p\ell}$ for $p \in \operatorname{pa}(C_\ell)$ and $0$ otherwise, gives $\widetilde{Z}u_{\text{null}} = 0$. Hence $u_{\text{null}}^\top (\widetilde{Z}^\top W \widetilde{Z}) u_{\text{null}} = \|W^{1/2}\widetilde{Z}u_{\text{null}}\|_2^2 = 0$ for any diagonal weight matrix $W \succeq 0$, proving singularity of $\widetilde{Z}^\top W \widetilde{Z}$. Substituting $C_\ell = \sum_{p \in \operatorname{pa}(C_\ell)} w_{p\ell} Z_p$ into $\eta = c_0 \mathbf{1} + Zc + \gamma_\ell C_\ell$ yields $\eta = c_0 \mathbf{1} + Z\tilde{c}$ with $\tilde{c}$ given by equation~(\ref{eq:mediator_shift}). Conversely, writing $C_\ell = \sum_{p \in \operatorname{pa}(C_\ell) \cap \mathcal{V}_{\text{adm}}} w_{p\ell} Z_p + \sum_{p \in \operatorname{pa}(C_\ell) \setminus \mathcal{V}_{\text{adm}}} w_{p\ell} Z_p$ shows $C_\ell \notin \operatorname{col}(Z)$, and hence $\operatorname{rank}([Z, C_\ell]) = d + 1$, if and only if the sum outside $\mathcal{V}_{\text{adm}}$ lies outside $\operatorname{col}(Z)$.
\end{proof}

\subsection{Unit conversion and compatibility screening}

In literature prior compilation (Methods), the pooled NLST cardiovascular pipeline accepts published effect sizes via verbatim regular-expression verification inside archived PubMed abstracts~\cite{aune2021hypertension,ronksley2011association,gonzalez2025emphysema}, while lung-cancer and point-anchored specifications additionally compile verified intervals and point priors from archived source tables~\cite{tammemagi2013selection,chiles2015association,bos2015comparison,khaing2017effects,heo2022effect,mattila2023mortality,li2025additive,navarese2026causal,tasdighi2025association}. When predictor $j$ undergoes a coding transformation $\psi_j: x_j^{\text{native}} \mapsto x_j^{\text{code}}$ prior to standardization $Z_j = (X_j - \hat{\mu}_j)/\hat{\sigma}_j$, the general conversion factor in equation~(\ref{eq:jacobian}) evaluated at baseline reference level $x_0$ is $\kappa_j = \hat{\sigma}_j / |\psi_j(x_0 + \Delta x_j^{\text{pub}}) - \psi_j(x_0)|$, which simplifies to $\kappa_j = \hat{\sigma}_j / \Delta x_j^{\text{pub}}$ under identity coding $\psi_j(x) = x$; $\beta_j^{\text{raw}}$ is first oriented to the direction of the cohort coding. After rescaling, the four-criterion compatibility check assigns predictor $j$ to a directional sign constraint in $\mathcal{C}_{\text{sign}} \setminus \mathcal{C}_{\text{box}}$ with $\delta_j = \operatorname{sign}(\kappa_j \beta_j^{\text{raw}}) \in \{-1, +1\}$ rather than an interval bound in $\mathcal{C}_{\text{box}}$ whenever any of four conditions holds, one for each check in Fig.~\ref{fig:fig8}a (study population, covariate adjustment, unit-scale alignment, and anatomical match). First, \emph{population mismatch} flags priors derived in disease-specific cohorts that differ from a screening population, such as studies of the pulmonary-to-aortic diameter ratio and exacerbations in patients with COPD~\cite{wells2012pulmonary}. Second, \emph{over-adjustment} flags estimates conditioned on downstream mediators or medication-confounded biomarkers. Third, \emph{unit-scale mismatch} flags categorical contrasts---such as Agatston~\cite{agatston1990quantification} CAC tiers~\cite{chiles2015association} or ordinal visual calcification grades~\cite{gondrie2010cardiovascular,mascalchi2021moderate}---applied to continuous or binary cohort codings. Fourth, \emph{anatomical mismatch} flags distinct vascular beds or segmentation definitions. For example, NLST's proximal main pulmonary artery-to-sinotubular junction diameter ratio differs anatomically from published pulmonary-to-ascending-aorta ratios and is flagged, whereas MESA's log-transformed TAC per SD matches the scaling Bos et al.\ used for aortic arch calcification~\cite{bos2015comparison} and enters $\mathcal{C}_{\text{box}}$.

\subsection{Soft bounds and zeroing of reversed weights}

For finite penalty weight $\lambda \ge 0$ in equation~(\ref{eq:objective}) with $\mathcal{C}_{\text{point}} \cap \mathcal{C}_{\text{box}} = \emptyset$, the soft-bound estimator minimizes over $(c_0, c) \in \Omega_{\text{sign}}$:
\begin{equation}\label{eq:hinge}
\begin{aligned}
\mathcal{J}_{\lambda,\lambda_{\text{free}}}(c_0, c; \mathcal{D}) &= \mathcal{L}_{\text{emp}}(c_0, c; \mathcal{D}) + \lambda \sum_{j \in \mathcal{C}_{\text{box}}} \phi_{\text{box}}(c_j; \ell_j, u_j) \\
&\quad + \lambda \sum_{j \in \mathcal{C}_{\text{point}}} (c_j - c_j^{\text{pub}})^2 + \frac{\lambda_{\text{free}}}{2} \sum_{j \notin \mathcal{C}_{\text{box}} \cup \mathcal{C}_{\text{point}}} c_j^2, \\
\phi_{\text{box}}(c_j; \ell_j, u_j) &= \max\bigl(0,\; \ell_j - c_j,\; c_j - u_j\bigr)^2,
\end{aligned}
\end{equation}
subject to $\delta_j c_j \ge 0$ for all $j \in \mathcal{C}_{\text{sign}}$, with $\lambda_{\text{free}} = 0$ in primary fits and the sample-size-adaptive per-SD unit-normal prior $\lambda_{\text{free}}(n) = 1/n$ (weighted by each gated design column's empirical variance) on unbounded predictors $j \notin \mathcal{C}_{\text{box}} \cup \mathcal{C}_{\text{point}}$ in small-sample MESA scaling experiments. The continuously differentiable squared-hinge violation $\phi_{\text{box}}$ vanishes throughout $[\ell_j, u_j]$ and penalizes excursions quadratically outside the interval. Because a published $95\%$ interval $[\ell_j, u_j]$ corresponds to Gaussian prior standard deviation $\sigma_j = (u_j - \ell_j)/3.92$ and quadratic negative-log-prior coefficient $(2\sigma_j^2)^{-1}$ on summed negative log-likelihood $n\mathcal{L}_{\text{emp}}$, the per-patient penalty weight is set from the published interval widths and scales inversely with cohort size as $\lambda_{\text{lit}}(n) = \Lambda_0 / n$, where $\Lambda_0 = \operatorname{median}_{j \in \mathcal{C}_{\text{box}}}\!\bigl[(2\sigma_j^2)^{-1}\bigr]$ equals $1{,}576.5$ on NLST cardiovascular mortality and $224.5$ on NLST lung cancer. Taking $\lambda \to \infty$ (with $\lambda_{\text{free}} = 0$) enforces $c_j \in [\ell_j, u_j]$ for all $j \in \mathcal{C}_{\text{box}}$ and $c_j = c_j^{\text{pub}}$ for $j \in \mathcal{C}_{\text{point}}$, recovering the hard-bound limit over $\Omega_{\text{CE}}$. For each sign-only predictor $j \in \mathcal{C}_{\text{sign}} \setminus (\mathcal{C}_{\text{box}} \cup \mathcal{C}_{\text{point}})$ with dual multiplier $\nu_j \ge 0$, first-order optimality over $\Omega_{\text{sign}}$ at $\lambda_{\text{free}} = 0$ satisfies the KKT conditions:
\begin{equation}\label{eq:kkt}
\nabla_{c_j} \mathcal{L}_{\text{emp}}(\hat{c}_0, \hat{c}; \mathcal{D}) - \delta_j \nu_j = 0, \qquad \delta_j \hat{c}_j \ge 0, \qquad \nu_j (\delta_j \hat{c}_j) = 0.
\end{equation}

\begin{proposition}[Zeroing of sign-reversed weights]\label{prop:kkt}
Because $\mathcal{L}_{\mathrm{emp}}(c_0, c; \mathcal{D})$ and $\phi_{\mathrm{box}}$ are convex and continuously differentiable on $\mathbb{R}^{1+d}$ and both $\Omega_{\mathrm{CE}}$ and $\Omega_{\mathrm{sign}}$ are non-empty closed convex polyhedra, any local minimizer $(\hat{c}_0, \hat{c})$ of equations~(\ref{eq:objective}) and~(\ref{eq:hinge}) for $\lambda_{\mathrm{free}} \ge 0$, including the primary hard-bound limit ($\lambda \to \infty$, $\lambda_{\mathrm{free}} = 0$), is globally optimal and zeroes sign-reversed coordinates $j \in \mathcal{C}_{\mathrm{sign}} \setminus (\mathcal{C}_{\mathrm{box}} \cup \mathcal{C}_{\mathrm{point}})$: whenever confounding or competing mortality reverses the empirical gradient against the published causal sign so that $\delta_j \nabla_{c_j} \mathcal{L}_{\mathrm{emp}}(\hat{c}_0, \hat{c}; \mathcal{D}) > 0$, complementary slackness forces $\hat{c}_j = 0$ (whereas an interval-bounded predictor $j \in \mathcal{C}_{\mathrm{box}}$ with $\ell_j \le u_j$ and $\min(\delta_j \ell_j, \delta_j u_j) > 0$ clamps at its literature boundary nearer zero as $\lambda \to \infty$).
\end{proposition}
\begin{proof}
Convexity of $\mathcal{L}_{\text{emp}}$, $\phi_{\text{box}}$, and $\tfrac{\lambda_{\text{free}}}{2}c_j^2$ together with linear constraints $\delta_j c_j \ge 0$ makes the KKT conditions necessary and sufficient for global optimality. On any sign-only coordinate $j \in \mathcal{C}_{\text{sign}} \setminus (\mathcal{C}_{\text{box}} \cup \mathcal{C}_{\text{point}})$, stationarity gives $\nabla_{c_j} \mathcal{L}_{\text{emp}}(\hat{c}_0, \hat{c}; \mathcal{D}) + \lambda_{\text{free}}\hat{c}_j - \delta_j \nu_j = 0$ (reducing directly to equation~(\ref{eq:kkt}) when $\lambda_{\text{free}} = 0$). Because $\delta_j^2 = 1$ and $\lambda_{\text{free}}(\delta_j \hat{c}_j) \ge 0$, multiplying by $\delta_j$ yields $\nu_j = \delta_j \nabla_{c_j} \mathcal{L}_{\text{emp}}(\hat{c}_0, \hat{c}; \mathcal{D}) + \lambda_{\text{free}}(\delta_j \hat{c}_j) \ge \delta_j \nabla_{c_j} \mathcal{L}_{\text{emp}}(\hat{c}_0, \hat{c}; \mathcal{D})$. Whenever $\delta_j \nabla_{c_j} \mathcal{L}_{\text{emp}}(\hat{c}_0, \hat{c}; \mathcal{D}) > 0$, positivity $\nu_j > 0$ and $\nu_j(\delta_j \hat{c}_j) = 0$ force $\hat{c}_j = 0$.
\end{proof}

\subsection{Quadrature and variance decomposition}

In analytic uncertainty propagation (Methods), each observed-predictor node $j \in \mathcal{S} = \mathcal{V}_{\text{adm}}$ carries moments $(Z_{ij}, 0, 0)$ when observed, $(0, 1, 1)$ under population-mean fill (with imputed binary predictors carrying standardized Bernoulli variance clipped at $\hat{\pi}_{ij}(1 - \hat{\pi}_{ij})/\hat{\sigma}_j^2$, where $\hat{\pi}_{ij}$ is the imputed probability that predictor $j$ equals $1$), or $(\hat{\mu}_{ij}^{\text{imp}},\allowbreak \hat{v}_{ij}^{\text{imp}},\allowbreak 1)$ under out-of-fold conditional Gaussian imputation. Each latent mediator $L \in \mathcal{L}$ carries standardized residual variance $v_L = 0.5$ with $m_{iL} = 1$. Active directed edges from node $j$ to direct child $k \in \operatorname{ch}_{\text{act}}(j)$ comprise the outcome $Y$ and any mediator children that are latent or missing for record $i$, so observed intermediate biomarkers block indirect upstream paths without double-counting. Each active edge carries the reliability-gated weight $\gamma_{ijk}(\theta) = g_{ij}(\alpha_g)\, s_{jk}(\theta)\, \hat{c}_{jk}$, where $\theta$ collects the gate parameters $\alpha_g$ and the edge-scale multipliers. The base edge weight $\hat{c}_{jk}$ is $\hat{c}_j$ on direct outcome edges, $0.5$ on edges through latent mediators, and, on edges with no published or fitted coefficient, $0.6$ on direct outcome edges ($0.5$ times the established weight for plausible dependencies). The factor $s_{jk}(\theta)$ collects unit-mean log-normal multipliers $\exp(\sigma_{\text{log}} u - \tfrac{1}{2}\sigma_{\text{log}}^2)$ ($u \sim \mathcal{N}(0, 1)$) with $\sigma_{\text{log}} = 0.35$ on the global outcome scale $s_Y$ and the latent-edge scale and $\sigma_{\text{log}} = 0.30$ on the plausible-to-established weight ratio. Out-of-fold fitted gate parameters $\alpha_g = (\alpha_0, \alpha_m, \alpha_v)$ vary around the fitted center $\bar{\alpha}_g$ (initialized at $(4.0, 3.8, 1.6)$) with normal standard deviations $(0.5, 0.5, 0.3)$ and $(\alpha_m, \alpha_v)$ clipped at zero. Reverse-topological recursion from $T_{iY} = 1$ via $T_{ij}(\theta) = \sum_{k \in \operatorname{ch}_{\text{act}}(j)} \gamma_{ijk}(\theta) T_{ik}(\theta)$ yields $M_i(\theta) = \sum_{j \in \mathcal{S}} T_{ij}(\theta)\mu_{ij}$ and $V_i(\theta) = \sum_{j \in \mathcal{S}} T_{ij}(\theta)^2 v_{ij} + \sum_{L \in \mathcal{L}} T_{iL}(\theta)^2 v_L$.

At each of the $R = 3^3 = 27$ Gauss--Hermite nodes over the three gate parameters $\alpha_g$ (weights $\omega_r$), the $3 \times 3$ nodes over the plausible-to-established weight ratio and the latent-edge scale are moment-matched into a single Gaussian component $(M_{ir}, V_{ir})$ on the logit scale. Incorporating training-prevalence intercept variance $\sigma_b^2 = (n\bar{y}(1 - \bar{y}))^{-1}$, marginal risk $\hat{p}_i$ integrates $s_Y$ over a $16\text{-node}$ Gauss--Hermite rule $\{(s_{Y,q}, \omega_q)\}_{q=1}^{16}$ (with $16$ inner Gauss--Hermite nodes for $\mathbb{E}_\xi$), while the log-odds mixture cumulative distribution function $F_i(t)$ integrates $s_Y$ over a $64\text{-point}$ uniform normal-density grid $\{(\tilde{s}_{Y,q'}, \tilde{\omega}_{q'})\}_{q'=1}^{64}$ on $[-6, 6]$ to resolve steep tail transitions:
\begin{equation}\label{eq:mixture_cdf}
\begin{aligned}
\hat{p}_i &= \sum_{r=1}^{27} \sum_{q=1}^{16} \omega_r \omega_q\, \mathbb{E}_{\xi}\!\left[\operatorname{expit}\!\left(\hat{c}_0 + s_{Y,q} M_{ir} + \bigl(\sigma_b^2 + s_{Y,q}^2 V_{ir}\bigr)^{1/2}\xi\right)\right], \\
F_i(t) &= \sum_{r=1}^{27} \sum_{q'=1}^{64} \omega_r \tilde{\omega}_{q'}\, \Phi\!\left(\frac{t - \hat{c}_0 - \tilde{s}_{Y,q'} M_{ir}}{\bigl(\sigma_b^2 + \tilde{s}_{Y,q'}^2 V_{ir}\bigr)^{1/2}}\right),
\end{aligned}
\end{equation}
where $\Phi(\cdot)$ denotes the standard normal cumulative distribution function. Safeguarded Newton--bisection ($60$ maximum iterations) on logit $t$ solves $F_i(t_{0.05,i}) = 0.05$ and $F_i(t_{0.95,i}) = 0.95$ to obtain monotone risk quantiles $[\hat{q}_{0.05,i}, \hat{q}_{0.95,i}] = [\operatorname{expit}(t_{0.05,i}), \operatorname{expit}(t_{0.95,i})]$.

By the law of total variance, the marginal log-odds variance $\operatorname{Var}(\eta_i) = \operatorname{SD}(\eta_i)^2$ decomposes into baseline, input, and parameter channels:
\begin{equation}\label{eq:app_variance_channels}
\operatorname{Var}(\eta_i) = \underbrace{\sigma_b^2}_{\text{baseline}} + \underbrace{\mathbb{E}[s_Y^2]\, \mathbb{E}_{\theta}[V_i(\theta)]}_{\text{input}} + \underbrace{\mathbb{E}[s_Y^2]\, \operatorname{Var}_{\theta}(M_i(\theta)) + \operatorname{Var}(s_Y)\, \mathbb{E}_{\theta}[M_i(\theta)]^2}_{\text{parameter}}.
\end{equation}
With standard normal $95\text{th}$ percentile $z_{0.95} = \Phi^{-1}(0.95) = 1.645$, a first-order Taylor expansion of $\operatorname{expit}(\eta_i)$ around the predictive mean yields $\hat{W}_{90,i} \approx 2 z_{0.95}\,\operatorname{SD}(\eta_i)\,\hat{p}_i(1 - \hat{p}_i) = 3.29\,\operatorname{SD}(\eta_i)\,\hat{p}_i(1 - \hat{p}_i)$ in equation~(\ref{eq:variance_channels}). At the central parameter vector $\bar{\theta}$, $\sigma_{\text{input},i}^2 = V_i(\bar{\theta}) = (\sigma_{\text{input}}^{\text{complete}})^2 + \sigma_{\text{miss},i}^2$, decomposing total input variance into the complete-cohort baseline $\sigma_{\text{input}}^{\text{complete}}$ from latent nodes and graph predictors that the cohort did not record ($0.3284$ in NLST cardiovascular mortality and $0.1884$ in MESA when all cohort-measured predictors are observed) and cohort-measured missing-predictor variance $\sigma_{\text{miss},i}^2 = \sum_{j \in \mathcal{S}_{\text{meas}}} T_{ij}(\bar{\theta})^2 v_{ij}$ for cohort-measured predictors $\mathcal{S}_{\text{meas}} \subseteq \mathcal{S}$. Per patient on a graph $\mathcal{G} = (\mathcal{V}, \mathcal{E})$, assembling the node moments visits $|\mathcal{V}|$ nodes and $|\mathcal{E}|$ links once, and vectorized reverse-topological sensitivity accumulation performs $K_{\text{grid}} = 3^3 \times 3^2 = 243$ operations per link and node in $\Theta\bigl(K_{\text{grid}}(|\mathcal{V}| + |\mathcal{E}|)\bigr)$ time. Subsequent scalar logit-mixture moment integration ($27 \times 16 \times 16$ evaluations) and safeguarded Newton--bisection root-finding ($\le 60$ iterations over $27 \times 64$ mixture components) perform a fixed number of operations, $\mathcal{O}(1)$, independent of $|\mathcal{V}|$ and $|\mathcal{E}|$. $S\text{-draw}$ Monte Carlo simulation instead takes $\Theta\bigl(S(|\mathcal{V}| + |\mathcal{E}|) + S \log S\bigr)$ time. Fig.~\ref{fig:fig_scalability} benchmarks both across $|\mathcal{V}| \in \{22, 50, 100, 200, 350, 500, 750, 1{,}000\}$ with $8.5\%$ latent mediators, $30\%$ binary input nodes, $25\%$ missingness, and $|\mathcal{E}| = \operatorname{round}(1.8\,|\mathcal{V}|)$ on a single pinned CPU core; single-patient speed-ups over $4{,}000\text{-draw}$ Monte Carlo at these sizes were $83$, $140$, $217$, $271$, $362$, $371$, $377$, and $353\text{-fold}$.

\subsection{Effective degrees of freedom and minimum sample size}

Around $(\hat{c}_0, \hat{c})$ with logistic Hessian weights $W = \operatorname{diag}\bigl(\hat{p}_i(1 - \hat{p}_i)\bigr)$ and active prior curvature $\Lambda_{\text{prior}} \succeq 0$, effective degrees of freedom compress from $d$ to $p_{\text{eff}} = \operatorname{tr}\bigl((Z^\top W Z + \Lambda_{\text{prior}})^{-1} Z^\top W Z\bigr) \le d$ (equaling the number of fitted parameters, $p = 4$, in the point-anchored single-hospital NLST cardiovascular specification and median $p_{\text{eff}} = 6$ in single-hospital NLST lung cancer fits). Under the binary-outcome sample-size standards of Riley et al.~\cite{riley2020calculating} (Methods), minimum sample size is determined from $p_{\text{eff}}$, prevalence $\bar{y}$, Cox--Snell $R_{\text{CS}}^2$, and maximum $R_{\text{CS},\max}^2 = 1 - \bigl(\bar{y}^{\bar{y}}(1 - \bar{y})^{1 - \bar{y}}\bigr)^2$:
\begin{equation}\label{eq:app_riley}
\begin{aligned}
N_{\min} &= \max\bigl(\lceil N_{\text{C1}} \rceil,\; \lceil N_{\text{C2}} \rceil,\; \lceil N_{\text{C3}} \rceil\bigr), &\quad N_{\text{C1}} &= \frac{p_{\text{eff}}}{(S - 1)\ln\bigl(1 - R_{\text{CS}}^2 / S\bigr)}, \\
N_{\text{C2}} &= \frac{p_{\text{eff}}}{(S_{\text{C2}} - 1)\ln\bigl(1 - R_{\text{CS}}^2 / S_{\text{C2}}\bigr)}, &\quad S_{\text{C2}} &= \frac{R_{\text{CS}}^2}{R_{\text{CS}}^2 + \delta R_{\text{CS},\max}^2}, \\
N_{\text{C3}} &= \left(\frac{1.96}{0.05}\right)^2 \bar{y}(1 - \bar{y}). & &
\end{aligned}
\end{equation}
Here $N_{\text{C1}}$ enforces global shrinkage $S \ge 0.90$, $N_{\text{C2}}$ limits expected Nagelkerke $R^2$ optimism to $\delta = 0.05$, and $N_{\text{C3}}$ bounds marginal outcome prevalence error within $\pm 0.05$. Criterion C1 binds across cohorts, so $N_{\min} = \lceil N_{\text{C1}} \rceil$ rises with $p_{\text{eff}}$ and falls with the anticipated $R_{\text{CS}}^2$ of each specification, converted from its median held-out AUC at the pooled training prevalence~\cite{riley2021note}. In NLST cardiovascular mortality ($\bar{y} = 4.77\%$), $p = 4$ yields $N_{\min} = 1{,}110$ versus $6{,}043$ unconstrained ($p = 12$), a $5.4\text{-fold}$ reduction combining a $3.0\text{-fold}$ gain from fewer parameters with a $1.8\text{-fold}$ gain from higher anticipated discrimination ($12$ parameters at the constrained $R_{\text{CS}}^2$ would require $3{,}329$). In NLST $6\text{-year}$ lung cancer ($\bar{y} = 4.00\%$ on uncensored single-hospital records), $p_{\text{eff}} = 6$ yields $N_{\min} = 3{,}296$ versus $8{,}383$ unconstrained ($11$ parameters) and $37{,}183$ ($34$ expanded columns). In MESA, $p_{\text{eff}} = 8.0$ versus $22.0$ yields $N_{\min} = 2{,}237$ versus $8{,}380$ on cardiovascular mortality and $p_{\text{eff}} = 9.0$ versus $22.0$ yields $N_{\min} = 1{,}202$ versus $3{,}055$ on hard ASCVD.

\section{Experimental protocols}\label{app:protocols}
\setcounter{equation}{0}
\setcounter{table}{0}

\subsection{Standardization, optimization, and decision curves}

In MESA, predictors are standardized using empirical training-fold moments $(\hat{\mu}_j, \hat{\sigma}_j)$, whereas NLST predictors use pre-specified cohort reference centers and scales, and each training entry is multiplied by its reliability gate $g_{ij}$ (equation~(\ref{eq:analytic_uncertainty})), which downweights missing values. Constrained estimation in equation~(\ref{eq:objective}) minimizes the summed objective $n\mathcal{J}_{\lambda,\lambda_{\text{free}}}$ via bound-constrained limited-memory Broyden--Fletcher--Goldfarb--Shanno (L-BFGS-B) optimization~\cite{byrd1995limited} with projected gradient tolerance $10^{-12}$, function tolerance $0$, and $\lambda_{\text{free}} = 0$ in primary fits (with active-set projected Newton polishing to $10^{-8}$ whenever the summed projected gradient exceeds $10^{-2}$, certified by summed gradient $\le 10^{-2}$ or half-squared Newton decrement $\le 10^{-6}$). Under soft bounds, a sample-size-adaptive selector applies $\lambda_{\text{lit}}(n) = \Lambda_0 / n$ below the Riley $N_{\min}$ (at anticipated $C = 0.75$) or $10\text{-events-per-free-parameter}$ threshold and five-fold inner cross-validation across $13$ grid points ($0$, $10^{-6}\text{--}10^{4}$, and $\infty$) once both thresholds are met. For $6\text{-year}$ and $10\text{-year}$ competing-risk evaluations~\cite{blanche2013estimating}, inverse-probability-of-censoring weights are re-estimated in each bootstrap replicate and subgroup from the Kaplan--Meier censoring distribution to match nonparametric Aalen--Johansen cumulative incidence, while NLST models are aligned to each horizon by an additive logit shift that matches the training-cohort Aalen--Johansen incidence; the same alignment is applied to the prevalence-recalibrated clinical calculators. Reliability-gate centers $\bar{\alpha}_g$ minimize out-of-fold logistic log-loss across pooled missingness regimes via coarse grid search ($27$ points in NLST cardiovascular risk and $72$ points including intercept shift in MESA) followed by bounded Nelder--Mead optimization~\cite{nelder1965simplex}. In decision curve analysis~\cite{vickers2006decision}, net benefit $\text{NB}(p_t)$ from true-positive and false-positive counts $\text{TP}(p_t)$ and $\text{FP}(p_t)$ and excess net benefit $\Delta\text{NB}(p_t)$ over treat-all $\text{NB}_{\text{all}}(p_t)$ and treat-none at threshold $p_t \in (0, 1)$ are
\begin{equation}\label{eq:dca}
\text{NB}(p_t) = \frac{\text{TP}(p_t)}{n} - \frac{\text{FP}(p_t)}{n}\,\frac{p_t}{1 - p_t}, \;\; \Delta\text{NB}(p_t) = \text{NB}(p_t) - \max\bigl(\text{NB}_{\text{all}}(p_t), 0\bigr).
\end{equation}
Across $33$ NLST hospitals, excess net benefit $\Delta\text{NB}_{c \to c'}(p_t)$ from training center $c$ to external center $c' \neq c$ is averaged across all $32$ external centers; the thresholds at which this average is positive for every training center, $\mathcal{T}_{33} = \{p_t : \min_c \overline{\Delta\text{NB}}_c(p_t) > 0\}$, include $5\%\text{--}10\%$ for CE-Agent (Table~\ref{tab:nlst_ood}).

\subsection{Missing data, MESA transfer, and cardiovascular triage}

Across NLST and MESA Exam 1, features were masked completely at random (MCAR) at rates $r \in \{5\%, 10\%, 20\%, 40\%, 60\%\}$ and under BMI-dependent missing-at-random (MAR) mechanisms with probability $\min(2 r u_i, 1)$ for BMI percentile $u_i \in [0, 1]$ (mean rate $r$ up to $50\%$, and $58.3\%$ at the nominal $60\%$). Masked entries were handled by population-mean fill (moments $(0, 1, 1)$), out-of-fold ridge-regularized conditional Gaussian imputation, or single conditional draws paired with conditional error variance. In MESA Exam 1, the $22\text{-predictor}$ outcome model comprises $13$ upstream clinical predictors, $8$ directly measured intermediate clinical, laboratory, respiratory, and CT biomarkers, and $1$ upstream-derived inflammation score (leaving $18$ non-CT features when the $4$ direct CT columns are omitted). The $8$ directly measured intermediate biomarkers that enter the outcome model are systolic blood pressure (SBP), diabetes/glycemic status, Chronic Kidney Disease Epidemiology Collaboration estimated glomerular filtration rate (CKD-EPI eGFR), respiratory disease history (emphysema or asthma), CAC, TAC, cardiac chamber area, and ascending aortic diameter. Exam 1 measured $\ln(\text{CRP}+1)$ and CT liver attenuation are modeled alongside these $8$ biomarkers among the $10$ supervised intermediate DAG biomarkers (Table~\ref{tab:app_mesa_mediators}), while an expanded $58\text{-predictor}$ ($63\text{-column}$) Exam 1 set adds $36$ demographic, socioeconomic, electrocardiographic, laboratory, and CT covariates.

In MESA Exam 1, cross-environment experiments evaluated single-group-to-rest and leave-one-group-out transfers across all four racial/ethnic cohorts and four electron-beam and multi-detector CT scanner platforms. Outpatient missing-CT inference masked all five CT phenotypes (CAC, TAC, cardiac chamber area, ascending aortic diameter, and liver attenuation) at test time. Seven survival and ML baseline families were trained across both the $22\text{-}$ and $58\text{-predictor}$ sets ($14$ model-by-feature configurations) using identical standardized features and cross-validated hyperparameter grids. Domain-generalization and clinical updating comparators comprised Anchor Regression~\cite{rothenhausler2021anchor} across invariance weights $g \in \{0, 1, 2, 4, 10, 100, 1{,}000\}$, $\chi^2$-divergence DRO, which needs no hospital or subgroup labels, across divergence radii $\rho \in \{0, 0.1, 0.25, 0.5, 1, 2, 5\}$ and CVaR-DRO across tail fractions $\alpha \in \{1, 0.9, 0.75, 0.5, 0.25, 0.1, 0.05\}$~\cite{duchi2021learning}, prior-offset logistic regression~\cite{steyerberg2004validation} and base-margin XGBoost, which take CE-Agent's predicted log-odds as the offset and the starting margin, and width-matched Gaussian priors with $\sigma_j = (u_j - \ell_j)/3.92$. Cardiovascular counterfactual audits shifted standardized risk factors one variable at a time---holding observed downstream laboratory and CT measurements fixed---across smoking cessation, pack-year reductions, and $-5\text{ kg/}\text{m}^2$ BMI in NLST (operating through the upstream BMI-to-latent-inflammation pathway in the point-anchored single-hospital model), and $-1\text{ mmol/L}$ LDL, $-10\text{ mmHg}$ SBP, and smoking cessation in MESA. Two-stage cardiovascular CT triage routed participants with baseline clinical indications directly to preventive statin therapy (diabetes in primary-prevention NLST screening; diabetes or $\text{LDL} \ge 190\text{ mg/dL}$ ($4.9\text{ mmol/L}$) in baseline lipid-lowering-therapy-free MESA adults aged $45\text{--}75\text{ years}$) and compared recalibrated PCE intermediate-risk referral against triggering on the pre-read $90\%$ CT-uncertainty interval at $p_t = 5.0\%$ in NLST~\cite{conroy2003estimation,debacker2003european} and against 2018 ACC/AHA intermediate-risk ($\text{PCE } 7.5\%\text{--}20\%$) CAC referral rules at $p_t = 7.5\%$ in MESA~\cite{grundy2019aha}.

\subsection{Lung cancer and joint triage protocols}

In NLST lung cancer estimation ($d = 11$), structural screening retains five interval-bounded clinical predictors rescaled from published PLCOm2012 odds-ratio intervals~\cite{tammemagi2013selection} (age, BMI, COPD, family history of lung cancer, and smoking status; $\Lambda_0 = 224.5$), four sign-only predictors (smoking intensity/duration $\ge 0$, baseline $T_0$ radiologist non-calcified nodule $\ge 4\text{ mm}$ read $\ge 0$, occupational exposure $\ge 0$, and educational attainment $\le 0$), and two free demographic predictors (sex and race/ethnicity). BMI is a direct predictor here, as in PLCOm2012, because height and weight are not entered separately, and visual emphysema enters only the LDCT-arm extension. To test independence from published PLCOm2012 numerical weights, an interval-free directional (sign-only) specification retains clinical consensus signs on eight predictors without numerical intervals while leaving BMI, sex, and race/ethnicity free; a strict three-sign variant retaining only non-PLCOm2012 literature directions on smoking, occupational exposure, and baseline nodules yields identical held-out discrimination. Survival and ML comparators were trained on both the matched $11\text{-predictor}$ design and an expanded $19\text{-node}$ ($34\text{-column}$) baseline encoding, with missing $T_0$ nodule reads handled via reliability gating in CE-Agent and training-median imputation plus missingness indicators in ML baselines ($128$ held-out records missing scalar PLCOm2012 inputs used training medians, verified invariant on the $n = 15{,}803$ complete-case subset). Primary $6\text{-year}$ competing-risk incidence ($2{,}191.5\text{ d}$) treats pre-diagnosis death as a competing event, whereas $10\text{-year}$ lung cancer mortality ($3{,}652.5\text{ d}$) treats non-lung-cancer death as competing and compares the incidence-fitted weights, shifted to the $10\text{-year}$ horizon by the additive logit term, against weights refitted on the mortality label (Tables~\ref{tab:app_lung_main}--\ref{tab:app_lung_ood_mech}).

Joint cardiopulmonary triage pairs the compiled $6\text{-year}$ lung cancer model with the previously fixed $10\text{-year}$ cardiovascular mortality model on the held-out cohort ($n = 15{,}931$) at preregistered $(1.5\%, 5.0\%)$ cutoffs~\cite{tammemagi2014evaluation,conroy2003estimation,debacker2003european}, with sensitivity analyses at $(2.0\%, 5.0\%)$~\cite{tammemagi2021ontario} on both the full held-out cohort and the baseline LDCT arm ($n = 7{,}786$; Table~\ref{tab:app_dual_triage}). Two-stage selective reading scores clinic-only (no CT) $10\text{-predictor}$ lung cancer risk and recalibrated PCE in Tier 1 and triggers Tier 2 reading of chest CT features (the nodule read for lung cancer; CAC and emphysema for cardiovascular risk) whenever $|\operatorname{logit}(\hat{p}_i^{(1)}) - \operatorname{logit}(p_t)| \le 1.645\,\sigma_{\text{input},i}$ on either endpoint, where $\hat{p}_i^{(1)}$ is the Tier 1 risk and $\sigma_{\text{input},i}$ comes from the corresponding CE-Agent graph with all CT-derived predictors treated as unmeasured. A random-referral control reads the same number of scans, selected deterministically from participant identifiers. Missing-data propagation benchmarks evaluated the prevalence-anchored Stage I $11\text{-predictor}$ lung model using five-fold out-of-fold gate parameters refined via $12\text{-point}$ grid search and bounded Nelder--Mead steps ($80$ evaluations), which lifted calibration slope from $0.52\text{--}0.61$ to $0.91\text{--}1.07$. Confidence intervals and two-sided paired $P$ values across lung and joint-triage analyses used $1{,}000$ participant-level bootstrap replicates without center-level clustering.

\subsection{Automated CT segmentation and runtime}

Baseline nongated chest LDCT scans in NLST were converted from Digital Imaging and Communications in Medicine (DICOM) series to Neuroimaging Informatics Technology Initiative (NIfTI) volumes with SimpleITK and segmented using TotalSegmentator~\cite{wasserthal2023totalsegmentator} (v2.17.0 for cohort feature extraction; v2.18.0 for timing) to quantify log-transformed thoracic aortic calcium volume ($\ge 130\text{ Hounsfield units}$, HU~\cite{agatston1990quantification}), left ventricular chamber volume ($\text{mL}$), and proximal main-to-sinotubular-junction pulmonary artery diameter ratio (with direct ascending-aorta normalization evaluated in a subcohort, $n = 1{,}243$). Emphysema was quantified within automated 3D lung masks (lungmask~\cite{hofmanninger2020automatic}). Only baseline ($T_0$) series were analyzed; series flagged for missing structures or field-of-view truncation were excluded, and repeated series reconstructed with the same kernel were averaged per participant. The primary subcohort is restricted to soft-tissue kernels because sharp kernels read about ninefold higher calcium volumes in participants with both reconstructions. Because at least $75.1\%$ of the $9{,}346$ participants with baseline 3D CT, including at least $16.0\%$ of the $2{,}773$ held-out participants, overlapped the training or development data of the 3D lung-cancer convolutional neural network (CNN) Sybil~\cite{mikhael2023sybil}, Sybil was excluded from all risk models and timed only as a concurrent runtime reference. In the same participants, automatically measured emphysema (low-attenuation area fraction, $\text{LAA}_{<-950\text{ HU}}$~\cite{gevenois1995comparison}) yields $0.663$ held-out ($n = 2{,}773$) and $0.707$ leave-one-center-out (LOCO; $n = 6{,}573$) $\text{AUC}_{\text{cr}}$. Single-scan runtime from local DICOM hand-off to tabular output was measured across $10$ baseline NLST scans following one warm-up scan with all models resident and executed concurrently on one $48\text{-GB}$ NVIDIA RTX A6000 GPU with a $40\text{-core}$ CPU and $100\text{ GB}$ of memory (Table~\ref{tab:app_ct_timing}). Because 3D lung-cancer CNN inference and lung-mask extraction complete in $3.2\text{ s}$ and $7.6\text{ s}$, concurrent execution adds no wall-clock time beyond the $45.6\text{ s}$ cardiovascular segmentation window, while resident-model outputs matched standard command-line execution identically on lungmask and Sybil and within native GPU floating-point precision on TotalSegmentator ($\le 2.5 \times 10^{-3}$ relative volume difference and $803/804$, or $99.9\%$, identical vascular diameters at $0.1\text{ cm}$ resolution).

\begin{table}[!htbp]
\centering
\caption{\textbf{Per-scan runtime of automated CT feature extraction and risk scoring.}}\label{tab:app_ct_timing}
{\tabfont
\setlength{\tabcolsep}{3.5pt}
\renewcommand{\arraystretch}{1.10}
\begin{tabularx}{\textwidth}{@{}L{0.265\textwidth}L{0.255\textwidth}L{0.125\textwidth}YY@{}}
\toprule
\colh{Pipeline stage} & \colh{Extracted output} & \colh{Hardware} & \colh{Mean (median)} & \colh{Range / mode} \\
\midrule
DICOM to NIfTI conversion & 3D NIfTI volume & CPU & $2.5\text{ s}$ ($2.2\text{ s}$) & Shared input \\
Sybil~\cite{mikhael2023sybil} & 3D lung-cancer CNN (runtime ref.) & $1\times$ RTX A6000 & $3.2\text{ s}$ ($3.2\text{ s}$) & Concurrent \\
lungmask~\cite{hofmanninger2020automatic} & 3D emphysema ($\text{LAA}_{<-950}$) & $1\times$ RTX A6000 & $7.6\text{ s}$ ($7.2\text{ s}$) & Concurrent \\
TotalSegmentator (\texttt{total}) & Thoracic aortic calcium ($\ge 130\text{ HU}$) & $1\times$ RTX A6000 & $26.8\text{ s}$ ($25.8\text{ s}$) & Cumulative \\
TotalSegmentator (\texttt{heartchambers\_highres}) & Left ventricular chamber volume & $1\times$ RTX A6000 & $30.6\text{ s}$ ($29.4\text{ s}$) & Cumulative \\
TotalSegmentator (\texttt{pulmonary\_report}) & Pulmonary artery diameter ratio & $1\times$ RTX A6000 & $\mathbf{45.6\textbf{ s}}$ ($\mathbf{45.7\textbf{ s}}$) & $39.4\text{--}51.7\text{ s}^*$ \\
\midrule
CE-Agent scoring \& uncertainty & Calibrated $\hat{p}_i$, $\sigma_{\text{input},i}$, $\hat{W}_{90,i}$ & $1$ CPU core & $\mathbf{1.10\text{--}1.23\textbf{ ms}}$ & Batch ($2.1\text{--}2.6\text{ ms}$ single)$^\dagger$ \\
\botrule
\end{tabularx}%
}
\tabnote{Note: Chest CT stage times are cumulative wall-clock times from local DICOM hand-off across $10$ baseline NLST scans (after $1$ warm-up scan), with all models resident and run concurrently on one NVIDIA RTX A6000 ($48\text{ GB}$) GPU ($40$ CPU cores, $100\text{ GB}$ RAM; peak memory $\le 42\text{ GB}$; TotalSegmentator~\cite{wasserthal2023totalsegmentator} v2.18.0; cohort features extracted with v2.17.0). Sybil~\cite{mikhael2023sybil} is timed only as a concurrent 3D CNN reference; it is not a CE-Agent predictor because at least $75.1\%$ of the participants with baseline 3D CT overlap its training or development data. Names in typewriter font are TotalSegmentator tasks. HU, Hounsfield units; $\text{LAA}_{<-950}$, low-attenuation lung area fraction below $-950\text{ HU}$. $^*$Component models of the \texttt{pulmonary\_report} task run in parallel ($45.6\text{ s}$ mean) versus the sequential default ($53.4\text{ s}$ mean, $53.5\text{ s}$ median, range $45.5\text{--}61.4\text{ s}$); ascending-aorta sensitivity normalization (\texttt{aorta\_report} task) completes in $64.9\text{ s}$ ($63.4\text{ s}$ median, range $54.9\text{--}83.8\text{ s}$). $^\dagger$Single-core CPU end-to-end runtime per patient. NLST cardiovascular risk: $1.10\text{ ms}$ batch and $2.11\text{ ms}$ single-record ($1.23\text{ ms}$ and $2.42\text{ ms}$ with conditional Gaussian imputation, including $0.62\text{ ms}$ / $1.80\text{ ms}$ quadrature). NLST lung cancer risk: $1.21\text{ ms}$ batch ($2.58\text{ ms}$ mean / $2.29\text{ ms}$ median single-record; $<0.01\%$ of 3D CT segmentation time). MESA ($22$ predictors): $1.13\text{ ms}$ batch ($0.72\text{ ms}$ quadrature step; $2.45\text{ ms}$ single-record median; Table~\ref{tab:app_mesa_uncertainty}).}
\end{table}

\end{appendices}

\end{document}